\PassOptionsToPackage{dvipsnames,table,xcdraw}{xcolor}
\documentclass[acmtog]{acmart}
\acmSubmissionID{753}
\usepackage{booktabs}
\usepackage{subcaption}
\usepackage{makecell}
\usepackage{enumitem}
\usepackage{adjustbox}
\usepackage{capt-of}
\usepackage{multirow}
\usepackage{tabularx}
\usepackage{float}
\usepackage{dashrule}

\definecolor{CRow}{RGB}{30,90,200}
\definecolor{BRow}{RGB}{0,150,90}
\definecolor{PRow}{RGB}{170,0,170}

\usepackage{xspace}

\def\pass{$\color{Green}{\checkmark}$}
\def\fail{$\color{red}{\pmb{\mathsf{X}}}$}

\definecolor{count_color}{rgb}{1, 0.5, 0.0}
\definecolor{attribute_color}{rgb}{0.85, 0.0, 0.95}
\definecolor{spatial_color}{rgb}{0.3, 0.7, 1.0}

\newcommand{\counttext}[1]{\textbf{\textcolor{count_color}{#1}}}
\newcommand{\attrtext}[1]{\textbf{\textcolor{attribute_color}{#1}}}
\newcommand{\spatialtext}[1]{\textbf{\textcolor{spatial_color}{#1}}}

\newcommand{\evaldataset}{\emph{DrawWaldoWorlds}\xspace}
\newcommand{\evaldatasetsupp}{\emph{MultiHuman-Testbench}\xspace}
\newcommand{\pose}[0]{pose}
\newcommand{\Pose}[0]{Pose}

\definecolor{promptOrange}{RGB}{233, 113, 50}
\definecolor{promptPink}{RGB}{160, 42, 147}

\definecolor{tZero}{RGB}{42,94,155}
\definecolor{tOne}{RGB}{251, 255, 9}
\definecolor{tTwo}{RGB}{141, 217, 116}
\definecolor{tThree}{RGB}{246, 193, 0}

\newcommand{\orangephrase}[1]{\textcolor{promptOrange}{#1}}
\newcommand{\pinkphrase}[1]{\textcolor{promptPink}{#1}}

\newcommand{\boxsize}{0.2em}
\newcommand{\filledorange}{%
  \colorbox{promptOrange}{\textcolor{promptOrange}{\rule{\boxsize}{\boxsize}}}%
}
\newcommand{\filledpink}{%
  \colorbox{promptPink}{\textcolor{promptPink}{\rule{\boxsize}{\boxsize}}}%
}
\newcommand{\emptyorange}{%
  \fcolorbox{promptOrange}{white}{\textcolor{white}{\rule{\boxsize}{\boxsize}}}%
}
\newcommand{\emptypink}{%
  \fcolorbox{promptPink}{white}{\textcolor{white}{\rule{\boxsize}{\boxsize}}}%
}

\definecolor{revisionblue}{RGB}{0,92,160}
\newcommand{\revadded}[1]{#1}
\newcommand{\revtable}[1]{#1}
\newcommand{\projectpage}{\url{https://cornell-vailab.github.io/PeopleComposer/}}

\copyrightyear{2026}
\acmYear{2026}
\setcopyright{cc}
\setcctype{by}
\acmConference[SIGGRAPH Conference Papers '26]{Special Interest Group on Computer Graphics and Interactive Techniques Conference Conference Papers}{July 19--23, 2026}{Los Angeles, CA, USA}
\acmBooktitle{Special Interest Group on Computer Graphics and Interactive Techniques Conference Conference Papers (SIGGRAPH Conference Papers '26), July 19--23, 2026, Los Angeles, CA, USA}
\acmDOI{10.1145/3799902.3811129}
\acmISBN{979-8-4007-2554-8/2026/07}

\begin{document}

\author{Wenxuan Peng}
\affiliation{%
  \institution{Cornell University}
  \city{Ithaca}
  \country{USA}
}
\email{wp267@cornell.edu}

\author{Bharath Hariharan}
\affiliation{%
  \institution{Cornell University}
  \city{Ithaca}
  \country{USA}
}
\email{bh497@cornell.edu}

\author{Hadar Averbuch-Elor}
\affiliation{%
  \institution{Cornell University}
  \city{New York}
  \state{New York}
  \country{USA}
}
\email{hadarelor@cornell.edu}
\orcid{0000-0003-3476-0940}

\title{Composing People Together: Iterative Pose-Image Generation for Multi-Person Interaction Scenes}

\begin{abstract}
Despite recent progress, text-to-image models still struggle to generate semantically diverse and compositionally accurate multi-person interaction scenes, often collapsing to repetitive layouts, stereotypical poses, and poorly grounded interactions.
In this work, we bridge this gap by introducing a dual pose--image representation that brings person-centric structural priors into pretrained diffusion transformers.
Our model jointly predicts a 2D pose visualization image and its corresponding RGB image, enabling structure and appearance to co-evolve during learning.
At its core, a cross-modal alignment scheme binds text, pose, and image representations, ensuring consistent grounding across modalities. Furthermore, we design an iterative scene construction scheme, progressively generating complex multi-human interactions while effectively decomposing the overall generation complexity.
Extensive experiments demonstrate that our method substantially improves prompt alignment and scene diversity in multi-person image generation.
Project page: \projectpage.

\end{abstract}



\begin{teaserfigure}
  \centering
  \includegraphics[width=\textwidth]{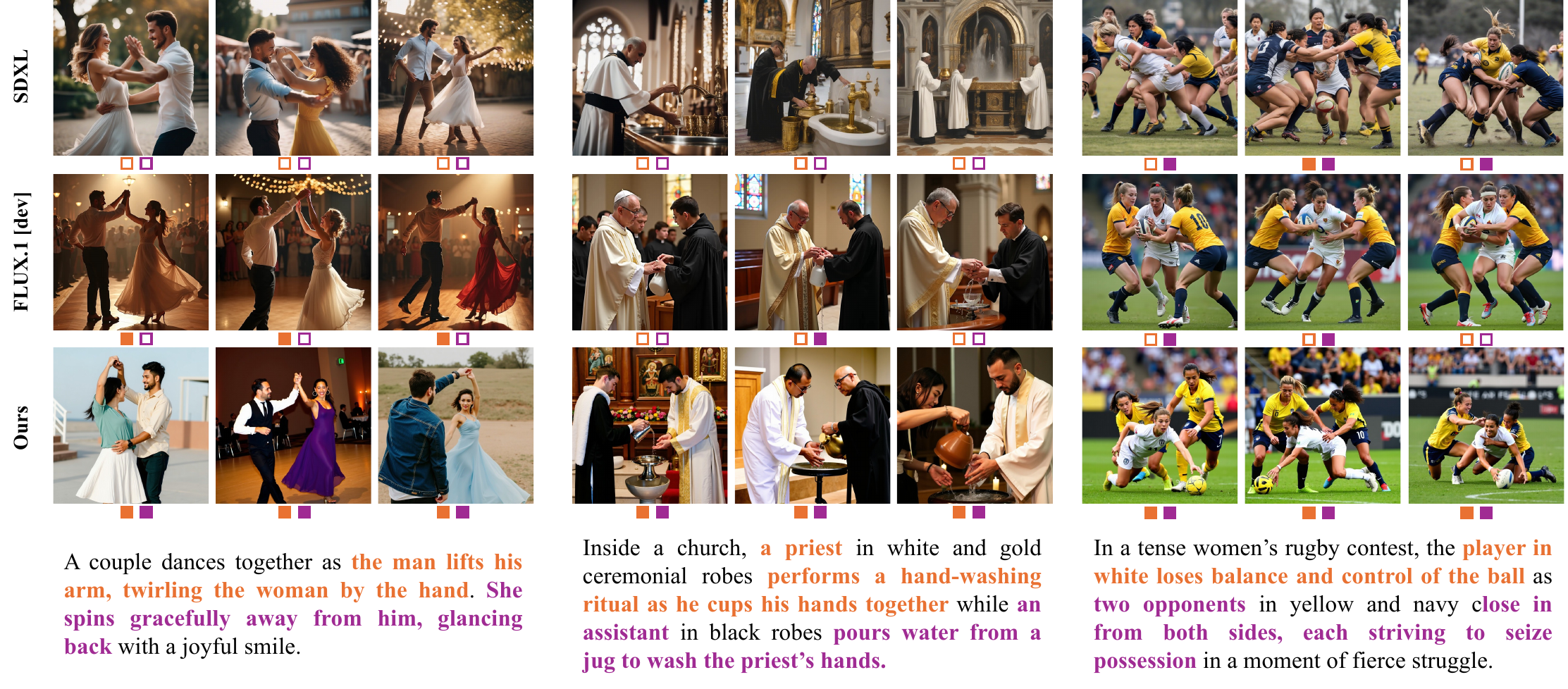}
  \caption{
  \textbf{Multi-Person Interaction Generation.}
  We show three random generations from SDXL, FLUX.1~[dev], and our method over three different multi-person interaction prompts (columns).
  \orangephrase{Orange} and \pinkphrase{purple} spans highlight key semantic segments of each interaction description. For every generated image we overlay a pair of markers, \filledorange{} / \emptyorange{} and \filledpink{} / \emptypink{}, indicating whether the image correctly realizes the corresponding orange or purple text segment (filled = aligned, empty = misaligned). Our method consistently realizes all text segments, reflecting stronger grounding of fine-grained interaction semantics while maintaining diverse scene structure.
  }
  \label{fig:teaser}
  \Description[Multi-person interaction generation comparison]{Grid comparing SDXL, FLUX, and our method across three multi-person interaction prompts. Markers indicate whether semantic prompt segments are satisfied for each generated image.}
\end{teaserfigure}

\maketitle

\section{Introduction}
Generating realistic humans has long been a central yet notoriously difficult problem in visual synthesis. Even subtle imperfections in facial geometry, body articulation, or contextual placement can disrupt perceptual realism, evoking the well-known \emph{uncanny valley} effect. These challenges are further amplified in multi-person generation, where not only must each individual appear anatomically plausible, but their spatial relationships, role-specific behaviors, and mutual affordances must also align coherently. Unlike multi-object compositions, which follow relatively rigid physical or functional constraints, multi--person interactions span an immense combinatorial space of poses, expressions, and social contexts. Capturing this variability while maintaining semantic alignment remains a core unsolved challenge for current text-to-image diffusion models.

As illustrated in Figure~\ref{fig:teaser}, earlier text-to-image models such as SDXL~\cite{podell2023sdxl} often fail to capture all entities and their distinct visual attributes within a coherent scene---for example, omitting either the priest or the assistant, or producing an incorrect number of rugby players. Newer models such as FLUX~\cite{flux2024} better capture person-level appearance and characteristics. However, correctly depicting their intended roles and interactions remains a significant challenge (\emph{e.g.}, none of the examples above depict the assistant actively pouring water to wash the priest's hands). Moreover, the generated images tend to collapse into a narrow range of poses, camera viewpoints, and spatial arrangements. This stands in stark contrast to the vast space of plausible human interactions, relationships, and visual configurations that such scenes naturally entail.

To better guide the generation process in complex, multi-entity prompts, prior work often relies on layout-based generation, where large language models provide coarse structural cues such as bounding boxes or spatial arrangements~\cite{phung2024grounded,dahary2024yourselfboundedattentionmultisubject}. While effective for organizing scene-level composition, such approaches offer only macro-level guidance and lack the granularity required to capture nuanced human structure and interaction. In this work, we move beyond coarse spatial guidance by introducing person-centric structural priors into diffusion transformers. Rather than imposing these priors as external constraints, we propose a dual pose--image representation that enables the model to internalize structure as part of the generative process. At the core of this design, a modified rotary positional encoding (RoPE) enforces spatial and semantic alignment across text, pose, and image modalities, allowing the model to reason jointly about geometry, appearance, and interaction.

Beyond providing a strong structural prior, pose estimation also enables iterative generation for composing complex multi-person scenes. This is because pose representations are inherently decomposable: unlike RGB images, they can be defined over arbitrary subsets of the people present in a scene. This property enables a person-by-person generation process in which the model incrementally constructs a scene by introducing one person at a time, conditioning each step on previously generated poses. Such iterative generation transforms the challenge of multi-person reasoning into a sequence of simpler sub-tasks while maintaining a coherent shared context.

To evaluate our approach and others on the challenge of generating complex multi-human interactions, we also propose a novel benchmark.
Our benchmark, \evaldataset{}, derives prompts of three different levels of detail from a prior dataset of complex and diverse human interactions~\cite{alper2023learning}.
We then evaluate ours and prior work on both the diversity of the generations as well as their adherence to both coarse-grained and detailed prompts.
We demonstrate that our approach produces diverse generations with significant improvements in alignment, capturing complex scenes with multiple people.

Explicitly stated, our contributions are:
\begin{enumerate}[itemsep=0pt, topsep=0pt,leftmargin=1.2em]
\item A dual pose--image representation that injects person-centric structural priors into pretrained diffusion transformers and enables iterative multi-person generation.
\item A new benchmark for evaluating diversity and alignment in multi-person interaction synthesis.
\item Significant improvements in both diversity and fidelity of multi-person generations compared to various existing techniques.
\end{enumerate}

\section{Related Work}

\subsection{Multi-Subject Text-to-Image Generation}
Pretrained text-to-image diffusion models~\cite{rombach2022highresolutionimagesynthesislatent,xue2024raphael,esser2024scaling} have achieved remarkable progress in generating diverse, high-quality images from textual descriptions.
However, these models still struggle to capture the fine-grained semantics of prompts involving multiple subjects.

To improve semantic alignment in multi-subject settings, prior work enhances the correspondence between textual concepts and generated visual entities.
Some methods refine text embeddings~\cite{feng2022training,tunanyan2023multi}, while others adjust attention mechanisms for better cross-modal alignment~\cite{chefer2023attendandexciteattentionbasedsemanticguidance,NEURIPS2023_0b08d733,shomer2026color}.
Large vision--language models have also been used as high-level controllers to iteratively steer diffusion models~\cite{wu2024self} or provide dual-process distillation~\cite{luo2025dual}.
Although effective, these approaches still struggle with multi-human interaction scenes.

Several methods introduce additional structural cues, such as structured text conditions~\cite{liu2023compositionalvisualgenerationcomposable} or coarse spatial layouts~\cite{phung2024grounded,dahary2024yourselfboundedattentionmultisubject}, including bounding-box approaches~\cite{xie2023boxdiff,chen2024training,xiao2023r,lee2024groundit} and methods that leverage richer geometric inputs such as depth or segmentation maps~\cite{zhou2025dreamrenderer,li2025seg2any}.
Recent approaches further leverage LLMs to infer coarse scene layouts directly from text~\cite{zhang2024realcompo,zhang2025creatilayout}.
We adopt a similar paradigm but show that coarse conditioning alone is insufficient for multi-person generation, which requires modeling the rich space of human interactions.
To address this, we learn fine-grained, pose-based structural priors directly within a 2D generative model, without requiring additional external supervision.

\subsection{Human-Centric Image Generation}
\label{sec:human_centric_generation}
A large body of work focuses on generating realistic images of people, including domains such as fashion~\cite{roy2022tips,wang2024stablegarment,chong2024catvton,girella2025lots} and portrait synthesis~\cite{chen2023s,xu2025hifi}; see Jia \emph{et al.}~\cite{jia2024human} for a survey.
Person-centric generation has also extended to text-guided 3D pose~\cite{feng2024chatpose,li2025unipose} and motion generation~\cite{tevet2022human,jiang2023motiongpt,guo2024momask}.

Human pose serves as a compact structural prior and is widely used in controllable person-centric image generation.
Pose-guided frameworks~\cite{zhang2023adding,shen2023advancing,wang2024stable} address reposing, reenactment, and motion-guided synthesis.
Examples include DisCo~\cite{wang2023disco}, HumanSD~\cite{ju2023humansd}, and Stable-Pose~\cite{wang2024stable}, which combine pose conditioning with text guidance.
While these methods achieve high structural fidelity, they rely on accurate pose annotations and focus mainly on single-person generation.
Other works~\cite{wang2024towards,li2025hrhuman} leverage pose or human vision foundation models~\cite{khirodkar2024sapiens} as auxiliary supervision, but similarly target individual-person scenarios.

Several recent methods similarly target multi-human or interaction-aware synthesis.
InteractDiffusion~\cite{hoe2024interactdiffusion} and VerbDiff~\cite{cha2025verbdiff} share our motivation of interaction-aware generation but target human--object rather than human--human interactions; PersonaCraft~\cite{kim2025personacraft} addresses identity-preserving multi-human generation from reference photos and 3D body models.
Most closely related, Chains~\cite{wei2024chains} also employs skeletons as structural intermediates for multi-person generation through a chained conditional pipeline.
By contrast, we learn an intermediate dual pose--image representation without requiring external pose input, and jointly predict pose and image within a single diffusion model to specifically ground fine-grained human--human interactions from text.

\subsection{Multimodal Expansion in Text-guided Models}
Following the success of text-to-image generation, recent work extends diffusion models to jointly generate additional modalities alongside RGB.
Some approaches adapt a color-pretrained VAE to encode depth or geometry~\cite{krishnan2025orchid,wang2024diffx}, or jointly predict outputs such as alpha mattes~\cite{li2024drip}.
JointNet~\cite{zhang2023jointnet} follows a ControlNet-like architecture~\cite{zhang2023adding} by duplicating diffusion pathways to jointly model multiple modalities.
Other recent methods~\cite{byung2025jointdit,kouzelis2025boosting,chefer2025videojam} introduce parallel diffusion paths within transformer architectures to jointly model RGB with depth or high-level features.

Our method similarly expands text-to-image generation through a dual-path design in a DiT architecture but focuses on integrating pose as an additional latent modality, enabling joint reasoning over structural and appearance features for multi-person synthesis.

\section{Method}
\label{sec:method}

Given a text prompt describing a multi-person interaction, our goal is to generate diverse, realistic images that faithfully ground the specified interactions. We begin by providing background on the architecture and the positional encoding scheme of the FLUX model (Section \ref{sec:prelim}).
Motivated by the premise that human pose encodes essential structural information for multi-person interaction, we introduce a dual image--pose representation that incorporates pose prediction as an auxiliary objective (Section \ref{sec:joint-rep}).
We then present our iterative generation scheme, which leverages the decomposable nature of pose to progressively construct multi-person scenes by adding one person at a time, effectively decomposing complex multi-person reasoning into a sequence of simpler sub-tasks (Section~\ref{sec:pose_autogressive_gen}).
Finally, we present the data used for training (Section \ref{sec:data}).

\subsection{Preliminaries}
\label{sec:prelim}
FLUX~\cite{flux2024}, which our method builds upon, is a Diffusion Transformer (DiT)~\cite{peebles2023scalable} model. FLUX is trained to denoise single-image tokens conditioned on text prompts. It concatenates text embeddings (from a frozen T5 encoder) and latent image tokens (from the FLUX VAE) into a single sequence processed by stacked transformer blocks. Each token receives a 3D rotary positional encoding (RoPE)~\cite{su2024roformer} based on its space--time coordinates $(\tau,x,y)$. For single-image inputs, $\tau$ is $0$ and $(x,y)$ indexes the latent grid. Text tokens share $(\tau,x,y)=(0,0,0)$.
The FLUX model also computes a pooled CLIP text embedding, which modulates the hidden activations~\cite{labs2025flux1kontextflowmatching}.

\begin{figure}[t]
  \centering
  \includegraphics[width=\columnwidth]{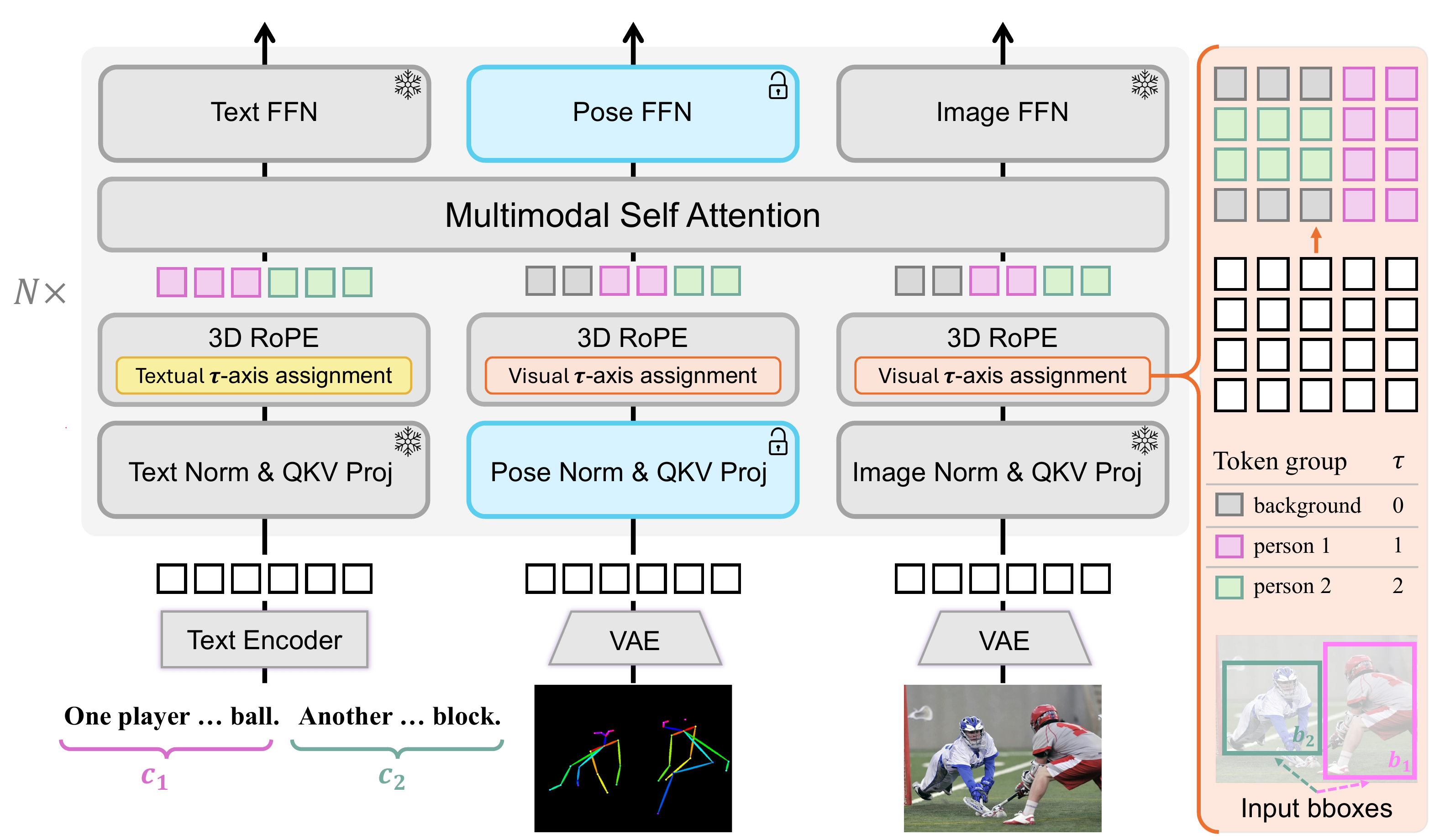}
\caption{\textbf{An overview of our dual pose--image diffusion transformer.} At each training iteration, random noise is added to the encoded tokens of the input images and their corresponding pose images. As illustrated above over the image branch (right), we enforce role-aware semantic binding via our proposed $\tau$-axis assignment. Text, bounding boxes and tokens associated with specific interacting entities are shown in unique colors (person 1 in \textcolor[rgb]{0.855,0.459,0.804}{pink} and person 2 in \textcolor[rgb]{0.455,0.675,0.620}{green}
).
}
\label{fig:method_refined}
\Description[Dual pose-image diffusion transformer overview]{Diagram of the dual pose-image diffusion transformer, showing image and pose branches, text and bounding-box conditioning, role-aware tau-axis assignments, and the trainable pose-related components.}
\end{figure}

\subsection{Dual \Pose{}--Image Generation}
\label{sec:joint-rep}
\medskip \noindent \textbf{Prompt Parsing and Coarse Layout Extraction.}
As a precursor to our generation process, we convert the user-provided interaction prompt into a set of structured elements that our method conditions on.
Specifically, we use an LLM to transform the prompt into a global interaction description $c_g$ together with an ordered set of per-person descriptions and associated bounding boxes $\{(c_i, b_i)\}$.
Beyond providing a structured, person-centric form, this parsing step promotes greater diversity in the generated outputs, particularly for concise or high-level input prompts. In such cases, the LLM expands the description by instantiating multiple plausible person-specific roles and interaction details, enabling multiple faithful realizations of the same interaction.
The parser is instructed to follow a coherent person order, typically starting from the interaction core and then expanding to nearby and more peripheral people.

\medskip \noindent \textbf{Generating and Encoding Pose with a Parallel Stream.}
While structured prompts and bounding boxes specify who appears in the scene and roughly where,
they provide only coarse layout cues and are insufficient for modeling human interactions, which depend on body configuration and relative orientation in addition to spatial location.
To capture this fine-grained person-centric structure, we introduce human pose as a compact, appearance-independent representation of how interactions are realized through human body configuration.
Rather than treating pose as an external condition, we incorporate it directly into the generation process by predicting pose jointly with the RGB image, requiring the model to explicitly reason about interaction-relevant structure during generation instead of leaving it implicit in appearance alone.

We represent poses as rendered 2D pose images in the standard OpenPose visualization format~\cite{cao2019openpose}. This pixel-based representation---rather than coordinates or per-joint heatmaps~\cite{fang2022alphapose,yu2021lite,toshev2014deeppose}---allows us to reuse the FLUX VAE encoder directly, maximizing compatibility with pretrained weights. We add noise to the encoded pose and image inputs,
and concatenate them with T5 embeddings (from the concatenated per-person descriptions $c_1\oplus \cdots \oplus c_N$) to form a multimodal token sequence (see Fig.~\ref{fig:method_refined}).

To process this sequence, we introduce a parallel pose stream initialized from the image-branch DiT parameters, forming a dual-branch diffusion transformer.
The pose stream provides a separate stream for pose tokens, while the original text and image streams of the pretrained FLUX backbone are preserved. All modalities interact through shared self-attention layers, enabling information exchange across text, pose, and image tokens; then separate output heads are applied to denoise the pose and image latents. We also retain FLUX's block-wise modulation and compute the pooled CLIP embedding from the global description $c_g$ (together with the diffusion timestep) to modulate the hidden activations.

To preserve the rich priors in the pretrained model, we freeze the entire text and image streams of the pretrained backbone.
The only newly trainable modules are LoRA parameters for the transformer weights in the pose stream (blue boxes in Fig.~\ref{fig:method_refined}) and a pair of input and output projection layers for the pose stream, copied from the original image projections and unfrozen during training (not shown). The original backbone structure remains completely intact.
This selective adaptation preserves the backbone's visual quality while injecting interaction-aware structural priors through the pose pathway and shared attention.

\medskip \noindent \textbf{Enforcing Cross-Modal Alignment via RoPE.}
We leverage the 3D rotary positional encoding already present in diffusion transformers to achieve person-level semantic alignment across modalities.
In the original FLUX model, while the spatial coordinates $(x,y)$ encode the layout of visual tokens, the temporal coordinate $\tau$ is fixed to 0 for image generation.
We repurpose this otherwise unused dimension to enable person-level semantic binding across modalities.

Concretely, all tokens associated with the same person, whether text, pose, or image tokens, are assigned a shared $\tau$ index, establishing a unified role identity across text, pose, and appearance.
To this end, we use the bounding boxes $b_i$ to identify the $i$-th person's visual tokens.
Image and pose tokens inside $b_i$ are then assigned $\tau=i$.
Similarly, textual tokens in the corresponding description $c_i$ are assigned $\tau=i$ (see Fig.~\ref{fig:method_refined}, right). Tokens outside all bounding boxes use $\tau=0$.
This role-aware indexing binds each person's textual description to the corresponding regions in both pose and image.
It also allows bounding boxes to control the spatial layout of each generated person, without modifying the backbone architecture or introducing additional parameters.
Finally, we assign the same spatial coordinates $(x,y)$ to pose and image tokens at corresponding locations, thereby ensuring that the structural and visual representations remain spatially aligned throughout the generation process.

\begin{figure}[t]
  \centering
  \setlength{\tabcolsep}{4pt}
  \renewcommand{\arraystretch}{1.1}

  \newlength{\cellw}
  \setlength{\cellw}{0.31\columnwidth}

  \begin{tabular}{@{}>{\centering\arraybackslash}m{\cellw}
                  >{\centering\arraybackslash}m{\cellw}
                  >{\centering\arraybackslash}m{\cellw}@{}}

  {\scriptsize$
    \textcolor{CRow}{c_g, c_1}\;
    \textcolor{black!45}{\vert}\;
    \textcolor{BRow}{b_1}\;
    \textcolor{black!45}{\vert}\;
    \textcolor{PRow}{\varnothing}
  $} &
  {\scriptsize$
    \textcolor{CRow}{c_g, c_1, c_2}\;
    \textcolor{black!45}{\vert}\;
    \textcolor{BRow}{b_1, b_2}\;
    \textcolor{black!45}{\vert}\;
    \textcolor{PRow}{P_1}
  $} &
  {\scriptsize$
    \textcolor{CRow}{c_g, c_1, c_2, c_3}\;
    \textcolor{black!45}{\vert}\;
    \textcolor{BRow}{b_1, b_2, b_3}\;
    \textcolor{black!45}{\vert}\;
    \textcolor{PRow}{P_2}
  $} \\

    \includegraphics[width=\linewidth]{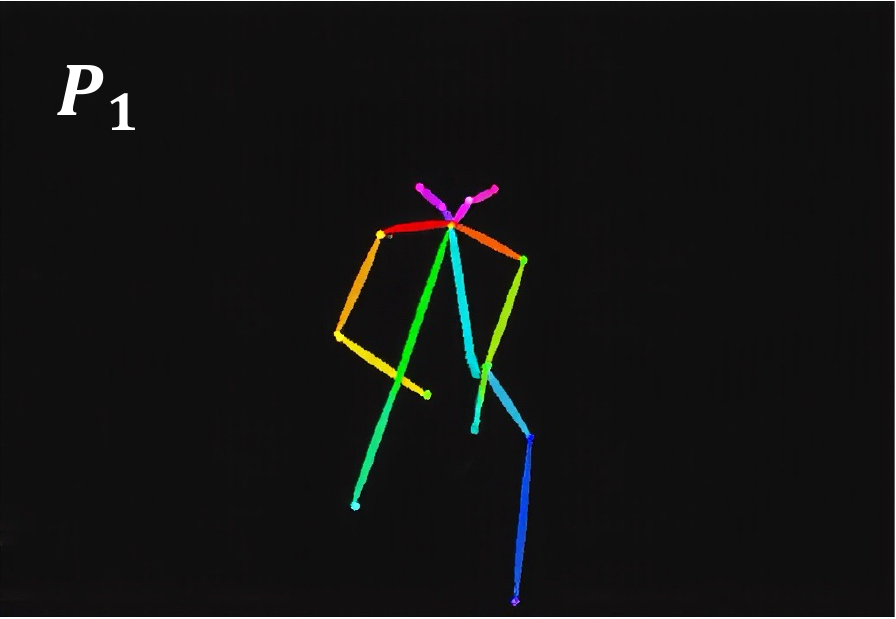} &
    
    \includegraphics[width=\linewidth]{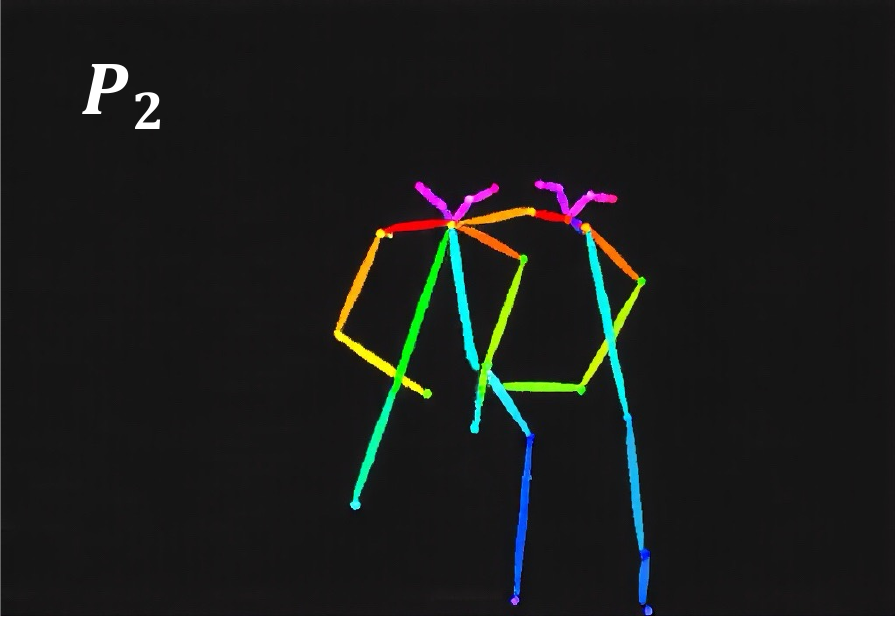} &
    \includegraphics[width=\linewidth]{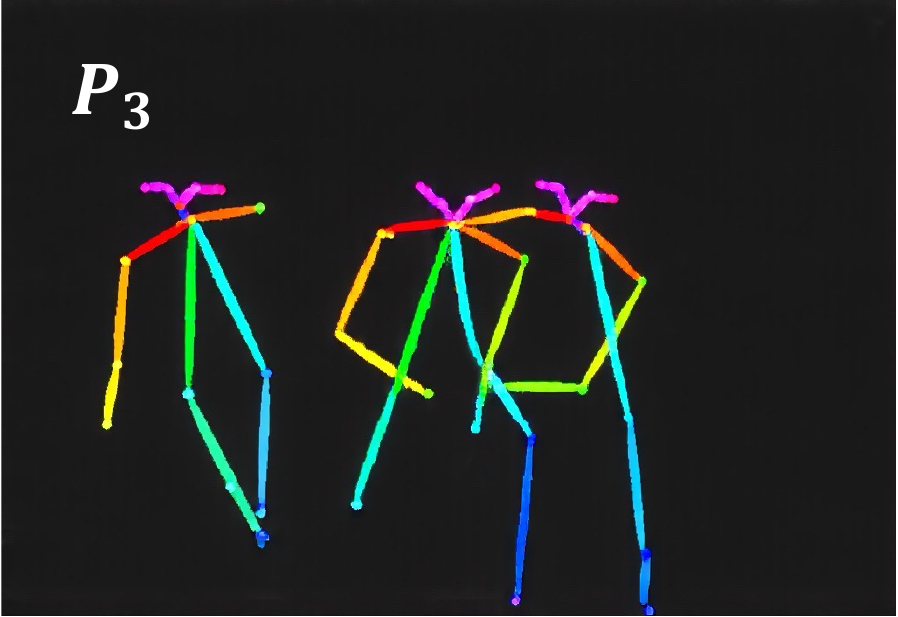} \\[-2pt]

    \includegraphics[width=\linewidth]{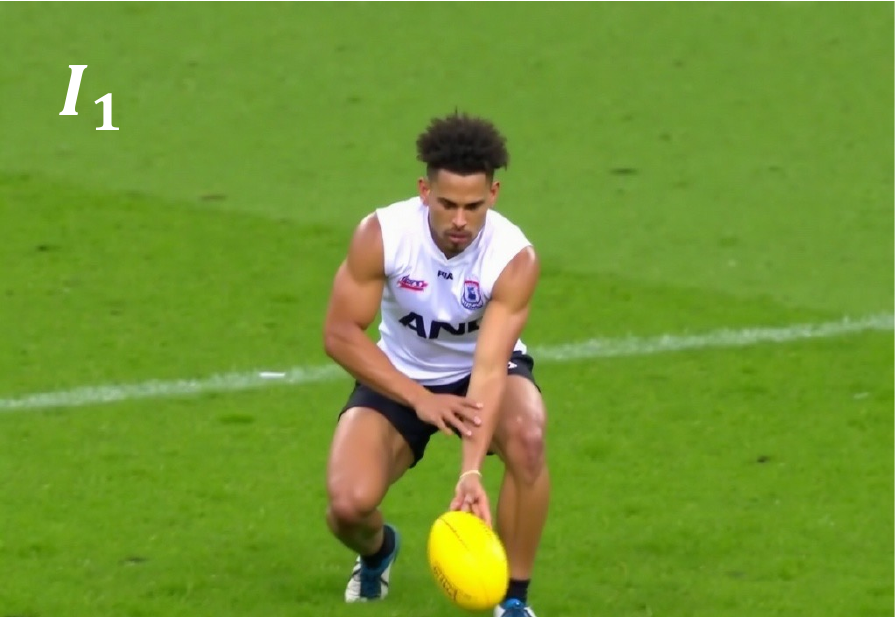} &
    \includegraphics[width=\linewidth]{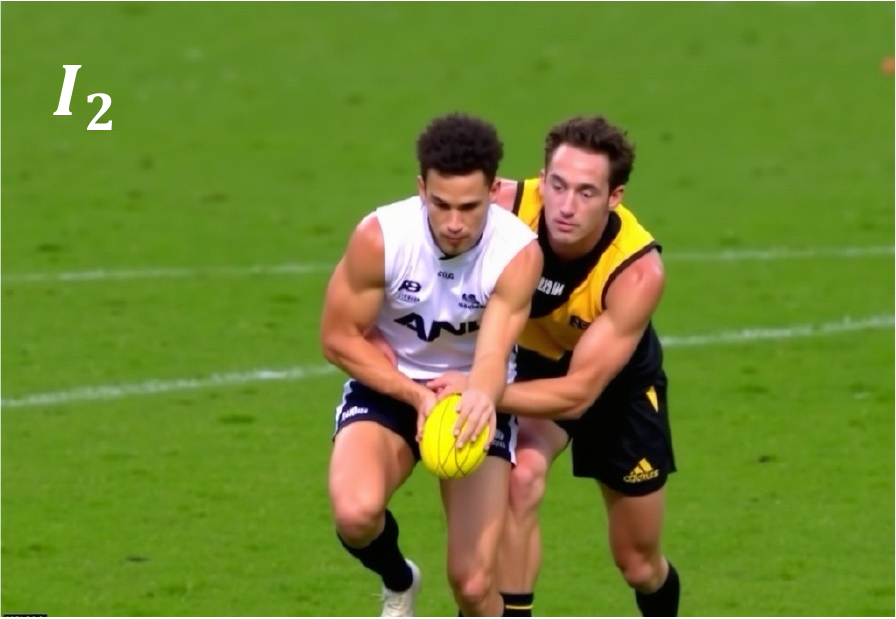} &
    \includegraphics[width=\linewidth]{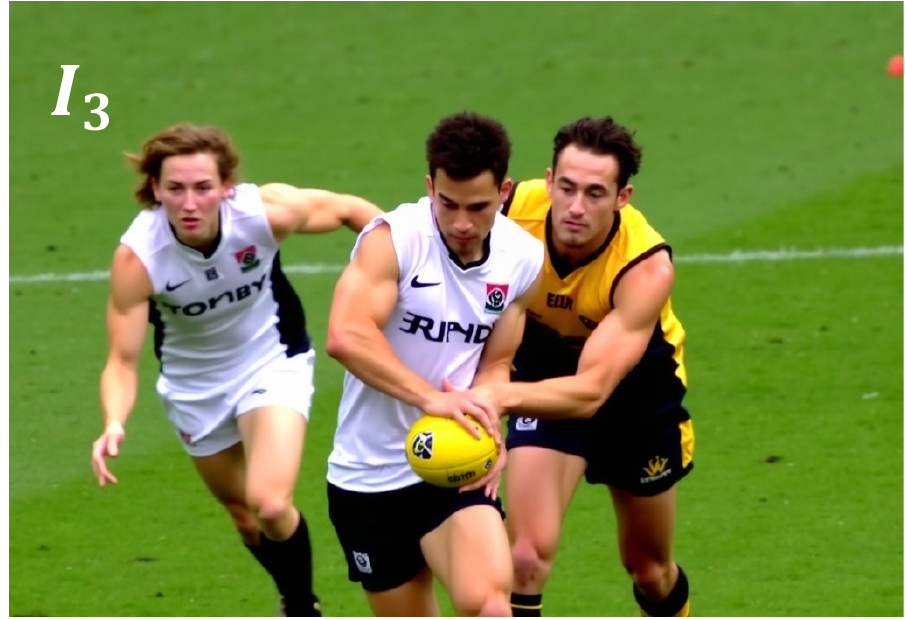} \\

  \end{tabular}
  Condition (\textcolor{CRow}{text} | \textcolor{BRow}{bounding boxes} | \textcolor{PRow}{pose}) and Pose--Image Output
  \caption{ Our iterative pose--image generation scheme progressively adds interacting people by conditioning on the previously generated pose image (i.e., the \emph{scene state}), along with partial text and bounding boxes, as illustrated on top. For the example above, the input text prompt is: \emph{``In an AFL match, a player drives forward with the ball as an opponent battles to strip it, while a teammate races in for support.''} The full parsed text descriptions are provided below\setcounter{footnote}{0}\protect\footnotemark{}.}  
  \label{fig:ar-pipeline-refined}
  \Description[Iterative pose-image generation example]{Three-step generation example showing text, bounding-box, and pose conditioning above pose and RGB outputs as additional people are composed into the scene.}
\end{figure}

\footnotetext{
$c_g$: In an AFL match, a player drives forward with the ball as a defender wraps him from behind, trying to strip it away, while a teammate rushes in to support.
$c_1$: On a green sports field, in the center, a player in a white-and-black jersey bends low toward the yellow AFL ball, moving in to secure it with his body lowered and angled forward in a strong, athletic stance.
$c_2$: From just behind, an opponent in a black-and-yellow jersey wraps in against him, both arms locking around his upper body as he reaches in to try to knock the ball loose.
$c_3$: To the left of the contest, a teammate in a matching white-and-black jersey sprints in, body angled forward and arms poised as he closes the gap to support the pickup.
}

\subsection{Pose-Guided Iterative Generation}
\label{sec:pose_autogressive_gen}
\Pose{} images have an additional advantage beyond being structural priors for image generation:
unlike the full RGB images depicting multi-person interactions, \pose{} images are inherently \emph{decomposable}.
We exploit this property by treating pose as the \emph{scene state} and adopting an iterative formulation for multi-human generation, where a scene is constructed incrementally by adding one person at a time on top of previously generated poses.
As we show in our experiments, this iterative generation simplifies the problem and is particularly important for input prompts that contain a relatively high number of interacting people.
Figure~\ref{fig:ar-pipeline-refined} shows an example of this generation process.

\medskip \noindent \textbf{Iterative Sequence Construction.}
In our iterative formulation, generation proceeds in stages, where each stage~$i$ adds the $i$-th person conditioned on the previously synthesized scene containing people~$1,\ldots,i\!-\!1$.
We define the \emph{scene state} as a pose image $P_i$ depicting all people generated up to stage $i$, with $P_0 = \varnothing$ denoting an empty canvas.
At stage $i$, the next person's description $c_i$ is appended to the text prompt encoded by T5, forming the extended text input $\mathcal{C}_i = c_1 \oplus \cdots \oplus c_i$,
and the previously generated pose visualization $P_{i-1}$ is inserted into the multimodal token sequence to serve as structural context.
Conditioned on $c_g$, $\mathcal{C}_i$, and $P_{i-1}$, the model jointly predicts the updated pose state $P_i$ and a corresponding image $I_i$.
Only the pose state is carried forward to the next stage; the final-stage image $I_N$ is the output.

During training, each $N$-person image yields $N$ supervision stages by decomposing the ground-truth pose rendering into intermediate states $P_1,\ldots,P_N$, where each $P_i$ contains the skeletons of the first $i$ people in the parsed order.
For intermediate stages ($i < N$), the model is supervised only on $P_i$; at the final stage it is supervised on both $P_N$ and the image $I_N = I$, following the standard denoising objective.
Additional discussion of occlusion handling, bounding-box overlap, and identity consistency across stages is provided in the Appendix.

\subsection{Training data}
\label{sec:data}
We build our training set from the \emph{Who's Waldo} dataset~\cite{cui2021whoswaldo},
a large-scale, person-centric vision--language dataset originally introduced for human--human interaction reasoning and visual grounding.
The dataset contains a diverse collection of real-world images depicting rich interactions between multiple people.

We first apply a filtering step to remove low-quality images as well as images that do not contain meaningful human--human interactions (e.g., people merely co-present without interaction), resulting in approximately 30k high-quality interaction images. We then augment each image with per-person pose detections~\cite{baidu_body_pose_api} and use a multimodal LLM to construct a structured interaction specification consisting of a global interaction description and an ordered list of per-person descriptions. Each description is matched to its corresponding individual using detected bounding boxes, yielding aligned text, pose, and spatial layout for every person. Further details of the data processing and filtering pipeline are provided in the Appendix.

\section{Experiments}

\label{sec:exp}
We present qualitative and quantitative experiments to evaluate our approach,
along with a human user study providing complementary evidence of improved quality in generated multi-person interactions.
Additional implementation details, evaluations, and a discussion of failure cases are provided in the Appendix.

\subsection{Benchmarks}

While several benchmarks, including DrawBench~\cite{saharia2022photorealistic}, GenEval~\cite{ghosh2024geneval}, and CompoundPrompts~\cite{sella2025instancegen}, evaluate text-to-image models, they do not explicitly target multi-person interactions.
A recent effort, \evaldatasetsupp{}~\cite{borse2025multihuman}, introduces a benchmark for multi-human image generation, originally designed for identity-preserving generation from reference photos.
The benchmark provides prompts organized into three categories---Single-Person, Multi-Person Simple, and Multi-Person Complex---along with a question--answer evaluation protocol.
We conduct an evaluation in a text-to-image setting, using the prompts and evaluation protocol without any reference images or identity conditioning.
However, the prompts primarily emphasize individual depiction rather than inter-person relations, so they do not explicitly test interaction-centric generation.

\paragraph{DrawWaldoWorlds}
We therefore introduce \evaldataset{}, a benchmark designed for \textbf{interaction-centric} multi-human text-to-image generation.
Rather than only verifying the presence of multiple people, \evaldataset{} evaluates whether a model can correctly ground \emph{who does what to whom} by jointly specifying role-specific actions and inter-person relations.
It also enables controlled analysis across prompt specificity, from under-specified interaction scenes to highly constrained descriptions requiring fine-grained compositional correctness.
We build \evaldataset{} based on the Waldo and Wenda test set~\cite{alper2023learning}, which provides images annotated with human--human interactions for vision--language reasoning.
To adapt this dataset for generative evaluation, we use a multimodal LLM to generate captions from each image and its interaction annotation. For each image, the LLM produces three prompts at increasing detail: a short interaction-focused description (Tier~A), a moderately detailed description (Tier~B), and a fine-grained image-specific description (Tier~C). This tiered design evaluates user-like prompts of varying specificity, testing alignment under progressively stronger compositional constraints while measuring diversity when prompts remain under-specified.
Representative three-tier prompts are included in Figure~\ref{fig:comparison}, with more examples in the Appendix.

\subsection{Metrics}
We evaluate model performance along two complementary dimensions: semantic alignment to the input text and diversity of the generated outputs. 
We also include an image quality evaluation in the Appendix.

\paragraph{Semantic alignment}
Semantic alignment measures whether a generated image correctly realizes the interactions specified in the prompt.
We use a VQA-based protocol in which a vision--language model (VLM) answers yes/no questions about prompt satisfaction.
We report two complementary metrics: a probabilistic score and a strict accuracy score.
\textbf{VQAScore} (VQA Sim)~\cite{lin2024evaluating} is the probability that the VLM answers ``Yes,'' providing a soft alignment signal.
\textbf{VQA Accuracy} (VQA Acc) records whether the VLM's discrete answer is ``Yes,'' yielding a stricter correctness criterion.

\paragraph{Diversity}
Diversity measures whether a model can produce varied yet plausible realizations of the same interaction prompt.
For each prompt, we evaluate a set of five generated images and compute diversity over this set; the sampling protocol for each model family is detailed in the Appendix.

We report three complementary metrics.
\textbf{DINO Feature Diversity} (DINO Diff) captures semantic and structural variation by computing pairwise distances between DINOv2 feature embeddings~\cite{parmar2025scaling}.
\textbf{LPIPS}~\cite{zhang2018unreasonable} captures perceptual variation via deep-feature distances and is widely used in generative modeling.
\textbf{GRADE}~\cite{rassin2024grade} captures semantic diversity by identifying prompt-relevant aspects with an LLM, querying them via VLM-based visual question answering, and measuring dispersion in the resulting answer distributions.

For \evaldataset{}, we report diversity on Tiers A and B only, as Tier C prompts fully specify the scene in high detail and leave limited room for meaningful variation. In \evaldatasetsupp{}, which does not define tiered prompt granularity, diversity is reported once over the full prompt set.

\subsection{Baselines}
We compare our method with a broad set of baselines representative of three major paradigms in text-to-image generation. The models reported in Tables~\ref{tab:main_results} and~\ref{tab:mmtest_main} fall into the following categories.

\paragraph{Text-to-image models}
This category includes standard T2I systems that synthesize images directly from a text prompt without additional structural guidance:
\emph{FLUX~[dev]} \cite{flux2024},
\emph{SD3.5-Large} \cite{esser2024scaling},
\emph{SDXL} \cite{podell2023sdxl}.
In addition, we include \emph{FLUX~[dev] + SGI}~\cite{parmar2025scaling}, which applies inference-time group selection on the same FLUX model to improve set-level diversity across multiple generated samples.
For these models, we evaluate both alignment and diversity by feeding the \emph{original} Tier~A/B/C prompts from \evaldataset{} and computing the corresponding metrics on their generated images.

\paragraph{Editing and inpainting models}
Since our method generates scenes person by person, we also compare to editing and inpainting baselines that can similarly iteratively modify an existing image by progressively adding one person at a time:
\emph{FLUX Kontext [dev]}~\cite{labs2025flux1kontextflowmatching} performs text-based editing, and \emph{FLUX Fill [dev]}~\cite{flux2024} supports masked-region inpainting.
For both models, we convert our parsed prompts into stepwise editing instructions that add one person at a time. Initialization and masking details are provided in the Appendix.

\paragraph{Layout-controlled models}
We also compare to other baselines that use bounding-box layouts paired with per-object textual descriptions, including recent representative methods
\emph{CreatiLayout}~\cite{zhang2025creatilayout} and \emph{RealCompo}~\cite{zhang2024realcompo}. 
Each model receives per-person descriptions and bounding boxes from our parsing pipeline in a single pass and generates the corresponding image.
Additional layout- and interaction-conditioned methods are provided in the Appendix.

\subsection{Comparisons}

We report quantitative results on both benchmarks, followed by qualitative comparisons, a user study, and ablations.

\paragraph{DrawWaldoWorlds}
Results on \evaldataset{} are reported in Table~\ref{tab:main_results}.
Our approach achieves the highest VQA Acc across all prompt tiers and the highest aggregate VQA Sim.
As prompt detail increases from Tier A to Tier C, VQA Accuracy drops for all models due to higher compositional complexity.
Tier A evaluates the core interaction, while Tiers B and C require increasingly fine-grained person-level actions and relations.
Our method maintains the highest alignment at every tier.
The aggregate VQA Sim shows the same overall ranking with higher absolute values because probabilistic scores penalize errors more softly than strict accuracy.

For diversity, our method outperforms most baselines on Tiers~A and~B.
Stronger alignment enables more reliable realization of prompt-level variability, yielding outputs that are both diverse and semantically coherent.
Editing / inpainting-based methods occasionally report higher diversity scores, but repeated multi-round editing can introduce visible artifacts that inflate feature-space diversity without improving perceptual or semantic quality.
Layout-controlled models show consistently low alignment across tiers.
Although bounding boxes constrain where people appear, this coarse layout signal is insufficient for interaction-specific structure, and these methods often fail to preserve text-to-entity correspondence in multi-human scenes.

\paragraph{MultiHuman-Testbench}
Table~\ref{tab:mmtest_main} reports results on \evaldatasetsupp{}.
Compared to \evaldataset{}, scores are more clustered, consistent with this benchmark's weaker emphasis on explicit human--human interactions.
Despite this reduced separation, our method maintains strong alignment on both interaction-focused and non-interaction-focused evaluations, while most baselines drop more noticeably on \evaldataset{}.

Even within this tighter performance range, the Multi Complex subset remains discriminative: we achieve the highest VQA Accuracy, outperforming the next best baseline by about 11 percentage points.
Although \evaldatasetsupp{} is not interaction-centric by design, this subset still requires compositional multi-person grounding, where prior methods---including layout-controlled models with coarse boxes---remain less reliable.
For the simpler subsets, alignment is generally high across models; a detailed per-subset analysis is provided in the Appendix.

\paragraph{Qualitative results}
Figures~\ref{fig:comparison} and~\ref{fig:gallery} provide qualitative evidence that complements our quantitative evaluation. Figure~\ref{fig:comparison} compares all baselines on \evaldataset{} across prompt tiers (A/B/C), while Figure~\ref{fig:gallery} showcases detailed Tier~C interaction scenes together with the iterative generation process of our method.
As prompt specificity increases, our method more reliably realizes the intended people and interactions, while baseline approaches often omit participants, misbind role-specific actions across individuals, or collapse to repetitive scene configurations.
The Appendix further includes a full qualitative comparison across all baselines and a discussion of failure cases on \evaldatasetsupp{}. 

\begin{table}[t]
\caption{\textbf{Evaluation on \evaldataset{}.} We evaluate all models using our benchmark's tiered prompt protocol. Our method achieves the best alignment across all tiers while maintaining strong diversity.}
\label{tab:main_results}
\Description[Main benchmark results]{Table comparing methods on DrawWaldoWorlds using DINO Diff, LPIPS, GRADE, VQA accuracy by prompt tier, and VQA similarity.}
\centering
\small
\renewcommand{\arraystretch}{1.2}
\setlength{\tabcolsep}{4.0pt}
\resizebox{\linewidth}{!}{%
\begin{tabular}{l cc cc cc ccc c}
\toprule
{\normalsize \textbf{Method}} &
\multicolumn{2}{c}{\textbf{DINO Diff} $\uparrow$} &
\multicolumn{2}{c}{\textbf{LPIPS} $\uparrow$} &
\multicolumn{2}{c}{\textbf{GRADE} $\uparrow$} &
\multicolumn{3}{c}{\textbf{VQA Acc} $\uparrow$} &
\textbf{VQA Sim} $\uparrow$ \\
\cmidrule(l{4pt}r{4pt}){2-3}
\cmidrule(l{4pt}r{4pt}){4-5}
\cmidrule(l{4pt}r{4pt}){6-7}
\cmidrule(l{6pt}r{6pt}){8-10}
& \textbf{A} & \textbf{B} &
  \textbf{A} & \textbf{B} &
  \textbf{A} & \textbf{B} &
  \textbf{A} & \textbf{B} & \textbf{C} & \\
\midrule
\multicolumn{11}{l}{{\small \textit{\textbf{T2I Models}}}} \\[-2pt]
FLUX [dev]               & 0.52 & 0.50 & 0.60 & 0.59 & 0.31 & 0.25 & 0.71 & 0.56 & 0.39 & 0.83 \\
FLUX [dev] + SGI         & 0.63 & 0.60 & 0.65 & 0.64 & 0.54 & 0.51 & 0.70 & 0.58 & 0.37 & 0.80 \\
SD3.5-Large              & 0.55 & 0.52 & 0.64 & 0.61 & 0.27 & 0.24 & 0.68 & 0.55 & 0.31 & 0.76 \\
SDXL                     & 0.62 & 0.60 & 0.64 & 0.63 & 0.39 & 0.31 & 0.55 & 0.20 & 0.12 & 0.57 \\
\midrule
\multicolumn{11}{l}{{\small \textit{\textbf{Editing/Inpainting Models}}}} \\[-2pt]
FLUX Kontext [dev]       & 0.61 & 0.59 & 0.66 & \textbf{0.65} & 0.55 & 0.52 & 0.50 & 0.34 & 0.28 & 0.61 \\
FLUX Fill [dev]          & 0.63 & 0.62 & 0.62 & 0.61 & \textbf{0.60} & \textbf{0.54} & 0.48 & 0.33 & 0.32 & 0.65 \\
\midrule
\multicolumn{11}{l}{{\small \textit{\textbf{Layout-Controlled Models}}}} \\[-2pt]
CreatiLayout             & 0.54 & 0.54 & 0.60 & 0.61 & 0.55 & 0.50 & 0.41 & 0.10 & 0.06 & 0.55 \\
RealCompo                & 0.59 & 0.58 & 0.63 & 0.60 & 0.57 & 0.51 & 0.38 & 0.10 & 0.04 & 0.56 \\
\midrule
Ours                     & \textbf{0.67} & \textbf{0.63} & \textbf{0.67} & \textbf{0.65} & 0.55 & \textbf{0.54} &
\textbf{0.84} & \textbf{0.72} & \textbf{0.56} & \textbf{0.87} \\
\bottomrule
\end{tabular}
}%

\end{table}

\begin{table}[t]
\caption{\textbf{Evaluation on \evaldatasetsupp{} (text-to-image setting).} We evaluate all models using the benchmark's original QA protocol. Our method achieves the best overall VQA Accuracy in the multi-person complex setting. A detailed per-subset analysis is provided in the Appendix.}
\label{tab:mmtest_main}
\Description[MultiHuman-Testbench results]{Table comparing text-to-image, editing, inpainting, and layout-controlled methods on MultiHuman-Testbench, with VQA accuracy reported for single-person, multi-person simple, and multi-person complex subsets.}
\revtable{\centering
\small
\renewcommand{\arraystretch}{1.2}
\setlength{\tabcolsep}{5pt}
\resizebox{\linewidth}{!}{%
\begin{tabular}{l c c c ccc c}
\toprule
{\normalsize \textbf{Method}} &
\textbf{DINO Diff} $\uparrow$ &
\textbf{LPIPS} $\uparrow$ &
\textbf{GRADE} $\uparrow$ &
\multicolumn{3}{c}{\textbf{VQA Acc} $\uparrow$} &
\textbf{VQA Sim} $\uparrow$ \\
\cmidrule(lr){5-7}
 &  &  &  & \makecell{\scriptsize Single} & \makecell{\scriptsize Multi\\\scriptsize Simple} & \makecell{\scriptsize Multi\\\scriptsize Complex} & \\
\midrule
\multicolumn{8}{l}{{\small \textit{\textbf{T2I Models}}}} \\[-2pt]
FLUX [dev]         & 0.57 & 0.61 & 0.42 & 0.77 & 0.78 & 0.41 & 0.85 \\
SD3.5-Large        & 0.56 & 0.63 & 0.41 & 0.78 & \textbf{0.86} & 0.39 & 0.88 \\
SDXL               & 0.62 & 0.62 & 0.49 & 0.78 & 0.77 & 0.37 & 0.83 \\
\midrule
\multicolumn{8}{l}{{\small \textit{\textbf{Editing/Inpainting Models}}}} \\[-2pt]
FLUX Kontext [dev] & 0.64 & 0.65 & 0.51 & 0.76 & 0.82 & 0.44 & 0.86 \\
FLUX Fill [dev]    & \textbf{0.65} & \textbf{0.66} & \textbf{0.58} & \textbf{0.80} & 0.79 & 0.36 & 0.85 \\
\midrule
\multicolumn{8}{l}{{\small \textit{\textbf{Layout-Controlled Models}}}} \\[-2pt]
CreatiLayout       & 0.63 & 0.64 & 0.48 & 0.75 & 0.78 & 0.38 & 0.84 \\
RealCompo          & 0.62 & 0.64 & 0.52 & 0.57 & 0.50 & 0.15 & 0.77 \\
\midrule
\textbf{Ours}      & 0.64 & \textbf{0.66} & 0.55 & \textbf{0.80} & 0.84 & \textbf{0.55} & \textbf{0.89} \\
\bottomrule
\end{tabular}
}%
}
\end{table}

\subsection{User Study}
\label{sec:user_study}

To further validate alignment, we conduct a user study comparing our model with two representative baselines: FLUX~[dev] (T2I) and FLUX Kontext~[dev] (iterative editing).
The study focuses on semantic alignment between prompt and generated image.
\begin{table}[t]
\centering
\caption{
User preference percentages comparing our model against two baselines.
Each row reports how often participants preferred our results, the baseline's results, or selected ``Cannot decide'' over 15 Tier~B prompt-image pairs. These results show that our model is preferred by human judges in both cases.
}
\label{tab:user_study_results}
\Description[User study preference results]{Table reporting human preference percentages for our method versus FLUX and FLUX Kontext, including cases where participants selected cannot decide.}
\normalsize
\renewcommand{\arraystretch}{1.1}
\begin{tabular}{@{}lccc@{}}
\toprule
Comparison & Ours & Baseline & Cannot \\
& better & better & decide \\
\midrule
Ours vs.\ FLUX [dev]         & \textbf{63.30\%} & 28.00\% & 8.70\% \\
Ours vs.\ FLUX Kontext [dev] & \textbf{79.00\%} &  11.30\% & 9.70\% \\
\bottomrule
\end{tabular}
\end{table}

\paragraph{Setup}
Each question presents one text prompt and two generated images---one from our model and one from a baseline---in randomized order.
Participants select the image that best matches the prompt. If both are equally aligned, they may choose the higher-quality image; a ``Cannot decide'' (Skip) option is also available.
We sample 30 prompt--image pairs from Tier~B: 15 comparisons with FLUX~[dev] and 15 with FLUX Kontext~[dev], mixed and randomly shuffled.
We choose Tier~B because Tier~C prompts are long and highly detailed, which makes it cognitively demanding for users to evaluate all fine-grained criteria across models. On the other hand, Tier~A prompts are short and under-specified, which often leads to ambiguous or subjective interpretations.
This makes Tier~B a practical middle ground for human evaluation, balancing semantic specificity with manageable judgment difficulty.
Each of the 20 participants completes all 30 comparisons.

\paragraph{Results}
Table~\ref{tab:user_study_results} reports user preference rates for both comparisons.
Our model is preferred in most cases: 63.3\% against FLUX~[dev] and 79.0\% against FLUX Kontext~[dev].
These results are consistent with our automatic evaluation and reinforce that our model follows the text more faithfully.

\subsection{Ablation Study}

To assess each component's contribution, we evaluate three ablated variants in Table~\ref{tab:ablation}.
All ablations are trained on the same data and evaluated under the same protocol as the full model, while using the same parsed person-level descriptions from our prompt parser as input.
\begin{table}[!ht]
\centering
\caption{
\textbf{Ablation study.}
We compare variants removing pose supervision (w/o Pose), iterative composition (w/o Iter.), and role-aware $\tau$-axis encoding (w/o RoPE).
Removing any component reduces semantic alignment and/or overall performance, with the full model achieving the strongest alignment.
}
\label{tab:ablation}
\Description[Ablation study results]{Table comparing the full model with variants that remove pose supervision, iterative composition, or role-aware tau-axis encoding, using diversity and semantic alignment metrics.}
\renewcommand{\arraystretch}{1.2}
\setlength{\tabcolsep}{4pt}
\resizebox{1\columnwidth}{!}{%
\begin{tabular}{l cc cc cc ccc c}
\toprule
\textbf{Method} &
\multicolumn{2}{c}{\textbf{DINO Diff} $\uparrow$} &
\multicolumn{2}{c}{\textbf{LPIPS} $\uparrow$} &
\multicolumn{2}{c}{\textbf{GRADE} $\uparrow$} &
\multicolumn{3}{c}{\textbf{VQA Acc} $\uparrow$} &
\textbf{VQA Sim} $\uparrow$ \\
\cmidrule(l{4pt}r{4pt}){2-3}
\cmidrule(l{4pt}r{4pt}){4-5}
\cmidrule(l{4pt}r{4pt}){6-7}
\cmidrule(l{6pt}r{6pt}){8-10}
& \textbf{A} & \textbf{B} & \textbf{A} & \textbf{B} & \textbf{A} & \textbf{B} &
\textbf{A} & \textbf{B} & \textbf{C} & \\

\midrule
w/o Pose & 0.62 & 0.58 & \textbf{0.68} & 0.63 & 0.51 & 0.46 & 0.66 & 0.49 & 0.31 & 0.74 \\
w/o RoPE & 0.65 & 0.61 & 0.65 & 0.59 & 0.52 & 0.49 & 0.78 & 0.63 & 0.44 & 0.82 \\
w/o Iter. & 0.61 & 0.58 & 0.66 & 0.64 & \textbf{0.58} & 0.51 & 0.73 & 0.57 & 0.37 & 0.83 \\
Ours & \textbf{0.67} & \textbf{0.63} & 0.67 & \textbf{0.65} & 0.55 & \textbf{0.54} &
\textbf{0.84} & \textbf{0.72} & \textbf{0.56} & \textbf{0.87} \\
\bottomrule
\end{tabular}
}%
\end{table}

\paragraph{w/o Pose}
This variant removes pose as an auxiliary target and directly fine-tunes the pretrained model on the same multi-person data.
Average VQA Acc drops by about 20 points across tiers, even below the pretrained baseline.
Since this variant uses the same training data with no architectural changes beyond standard fine-tuning, the drop confirms that our gains stem from the architectural design rather than from in-domain data alone.
Pose provides essential person-centric supervision, and training only the pose branch while freezing the pretrained image backbone yields significantly better alignment.

\paragraph{w/o RoPE}
This variant removes our role-aware $\tau$-axis assignment for text tokens while retaining pose prediction, iterative generation, and bounding-box layout.
All text tokens revert to default RoPE coordinates $(\tau,x,y)=(0,0,0)$, removing explicit person-level text--vision binding.
Alignment drops relative to the full model, though this variant still outperforms baseline models, reflecting the benefits of pose supervision and iterative composition.
The additional drop versus the full model highlights the importance of identity-aware cross-modal positional encoding for binding per-person descriptions to visual entities.

\paragraph{w/o Iter}
This variant removes iterative generation and produces all people in one pass, while still jointly predicting pose and image under the same structured prompt and role-aware settings.
Alignment degrades noticeably versus the full model, confirming that person-by-person iterative composition is critical for decomposing multi-person generation into simpler subproblems and reducing semantic entanglement.
Together, these ablations show that pose-based structural supervision, role-aware cross-modal binding, and iterative composition each contribute to performance, and all three are necessary for strong semantic alignment in multi-person interaction generation.

\section{Conclusion}
In this work, we presented a framework for multi-person text-to-image generation that explicitly models person-centric structure
through a dual pose--image representation.
By jointly generating pose and appearance, our model introduces a structured representation that captures interaction-relevant information
beyond coarse spatial layout.
We further incorporate role-aware cross-modal alignment by modifying the rotary positional encoding,
enabling consistent binding between text, pose, and image tokens at the level of individual people.
Building on this structured representation, we adopt an iterative, person-by-person generation scheme that leverages the decomposability of pose to progressively compose complex multi-person scenes.
Altogether, these components enable improved semantic alignment under fine-grained interaction prompts
while preserving strong diversity, as demonstrated by extensive quantitative, qualitative, and user study results.

More broadly, our work highlights how explicit, person-centric structural abstractions can serve as a powerful bridge between high-level semantic intent and low-level visual generation. By disentangling interaction structure from appearance and leveraging iterative composition, our approach offers a scalable way to manage the combinatorial complexity inherent in multi-entity scenes. Looking ahead, integrating structured, decomposable representations as first-class modalities within generative models opens new opportunities for more faithful, interpretable, and controllable synthesis, not only for human-centered interactions, but also for increasingly complex scenes involving richer dynamics, longer temporal horizons, and more diverse interacting entities.

\clearpage
\bibliographystyle{ACM-Reference-Format}
\bibliography{main,blended}
\clearpage
\appendix

\providecommand{\blacksquare}{\mathord{\hbox{\rule{0.9ex}{0.9ex}}}}
\providecommand{\square}{\mathord{\hbox{\fbox{\rule{0pt}{0.9ex}\hspace{0.9ex}}}}}

\newcommand{\minipageimage}[3][0.12]{%
  \begin{minipage}[t]{#1\linewidth}%
    \centering \small #3\\
    \includegraphics[width=\linewidth]{#2}%
  \end{minipage}%
}

\newcommand{\ExtOursA}{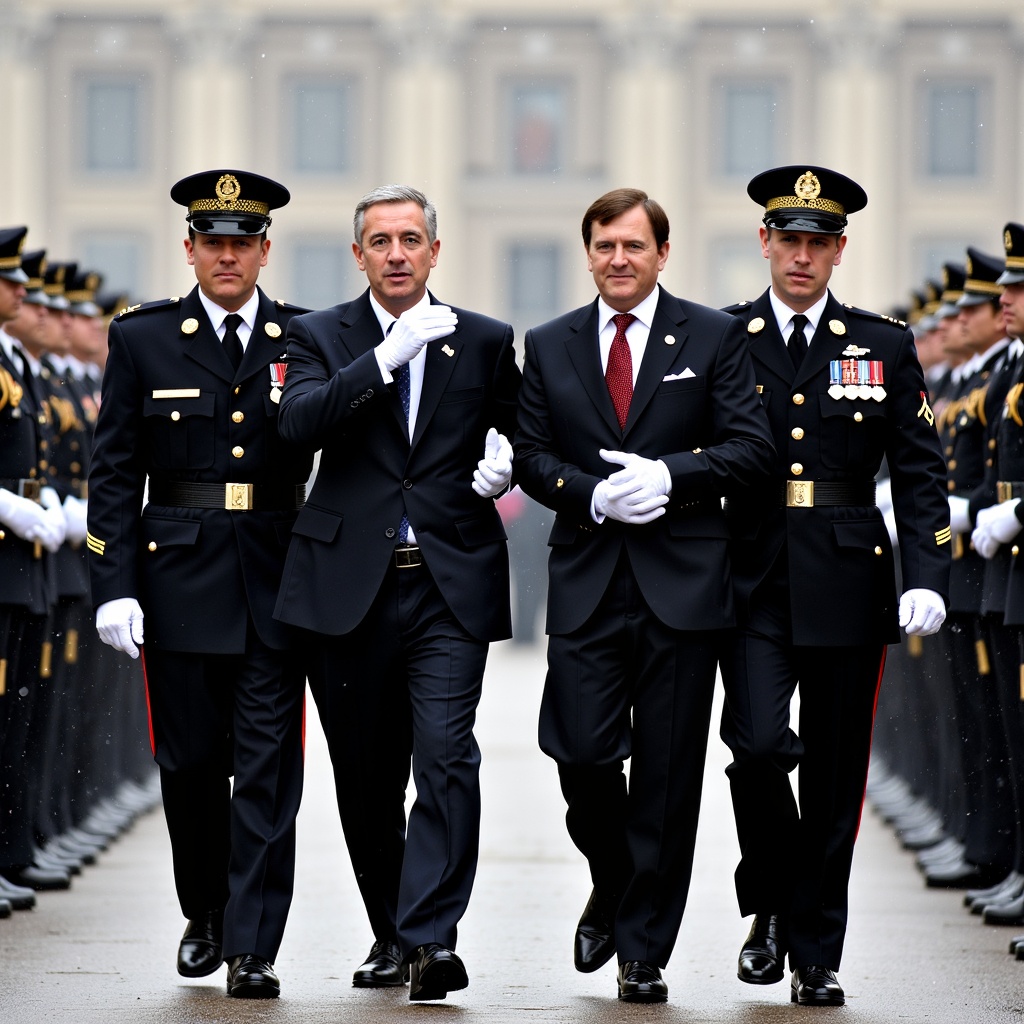}
\newcommand{\ExtOursB}{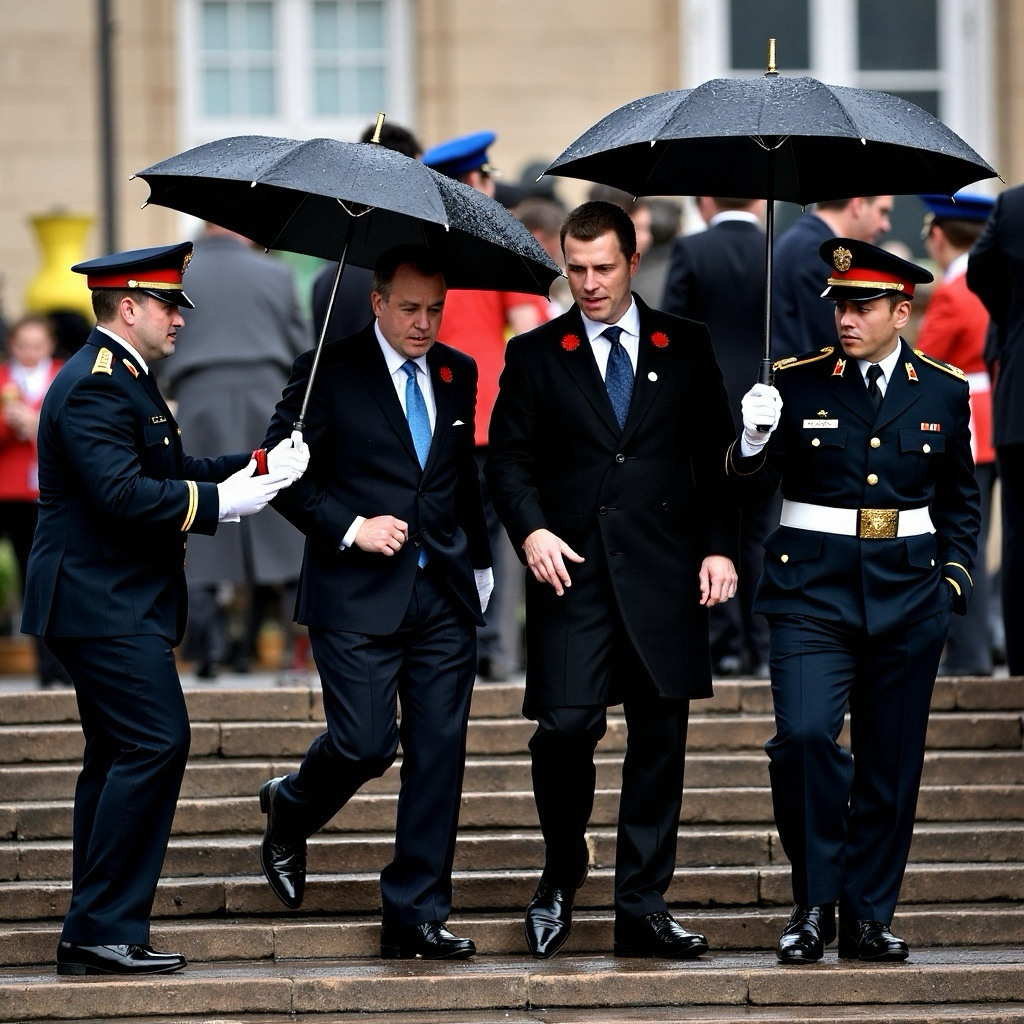}
\newcommand{\ExtOursC}{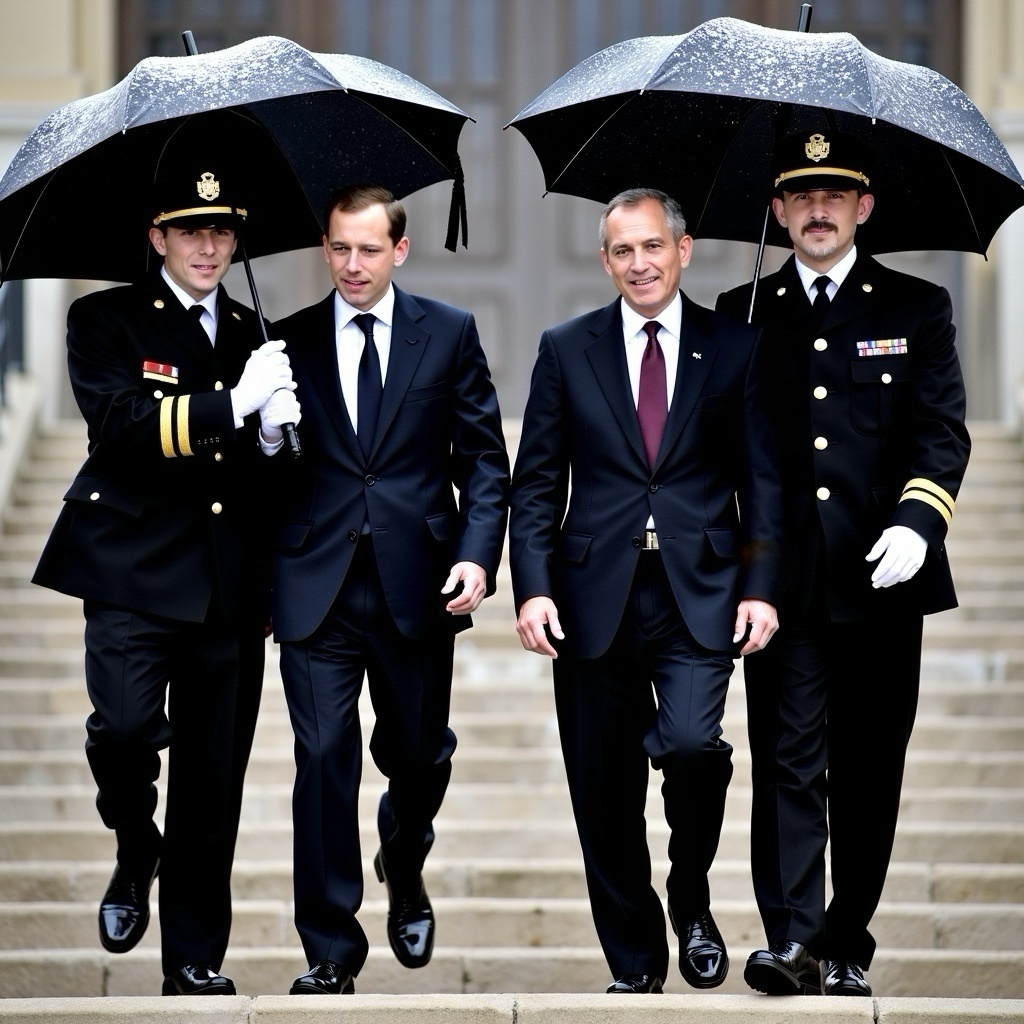}

\newcommand{\ExtBaseA}{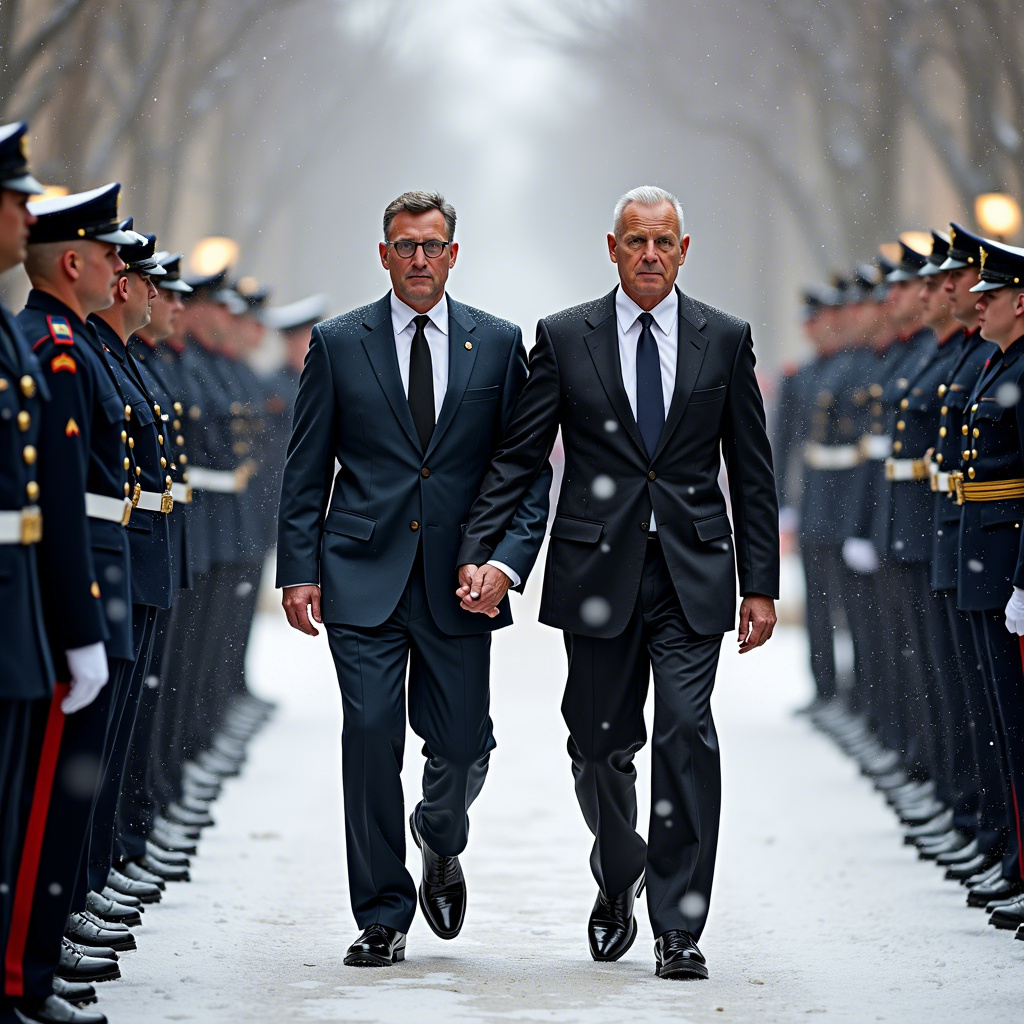}
\newcommand{\ExtBaseB}{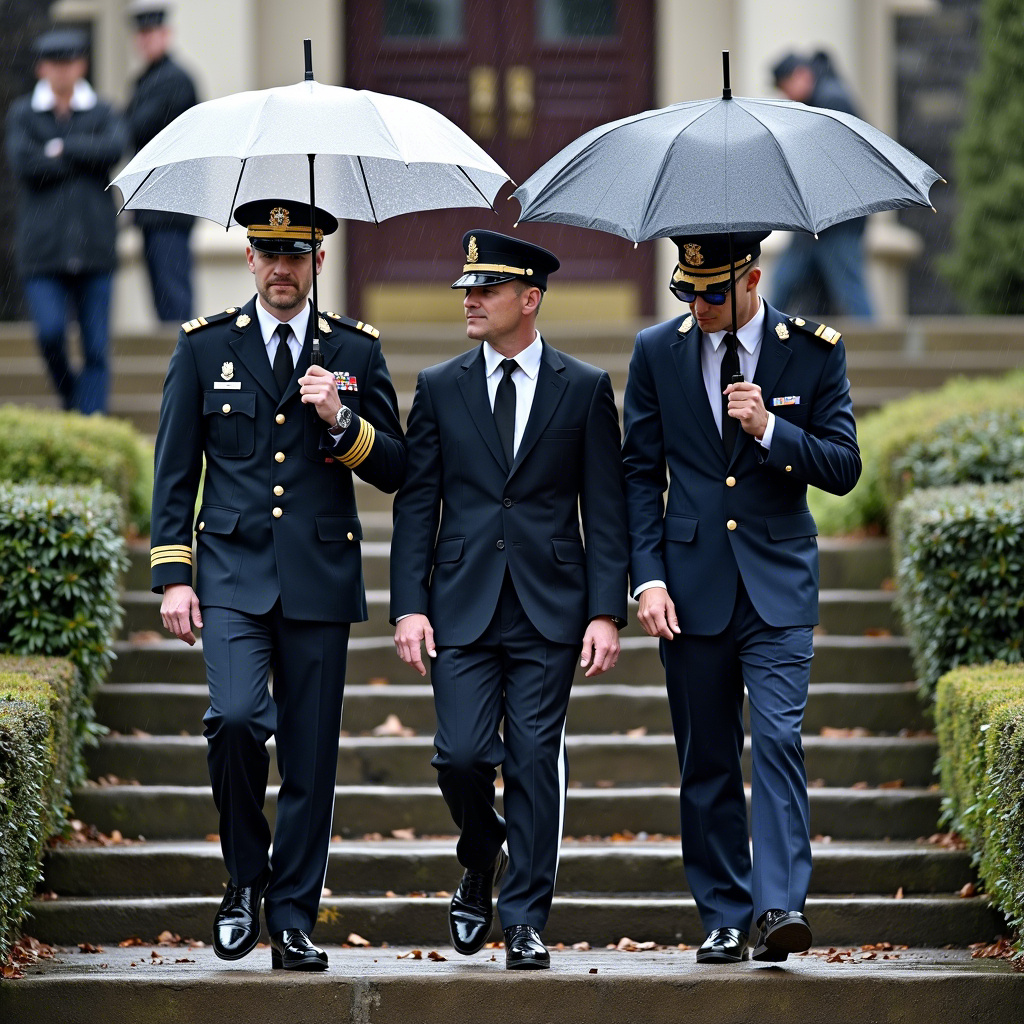}
\newcommand{\ExtBaseC}{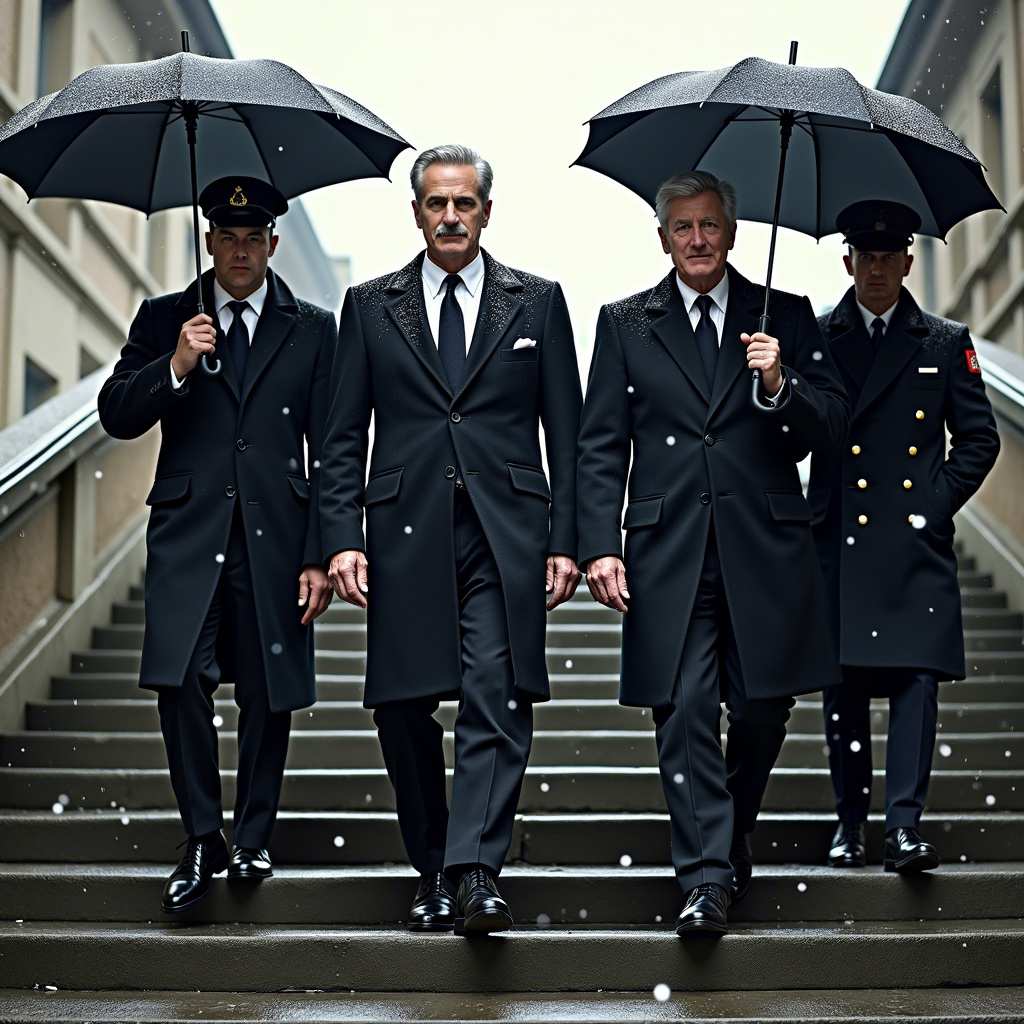}

\newcommand{\ExtSDLargeA}{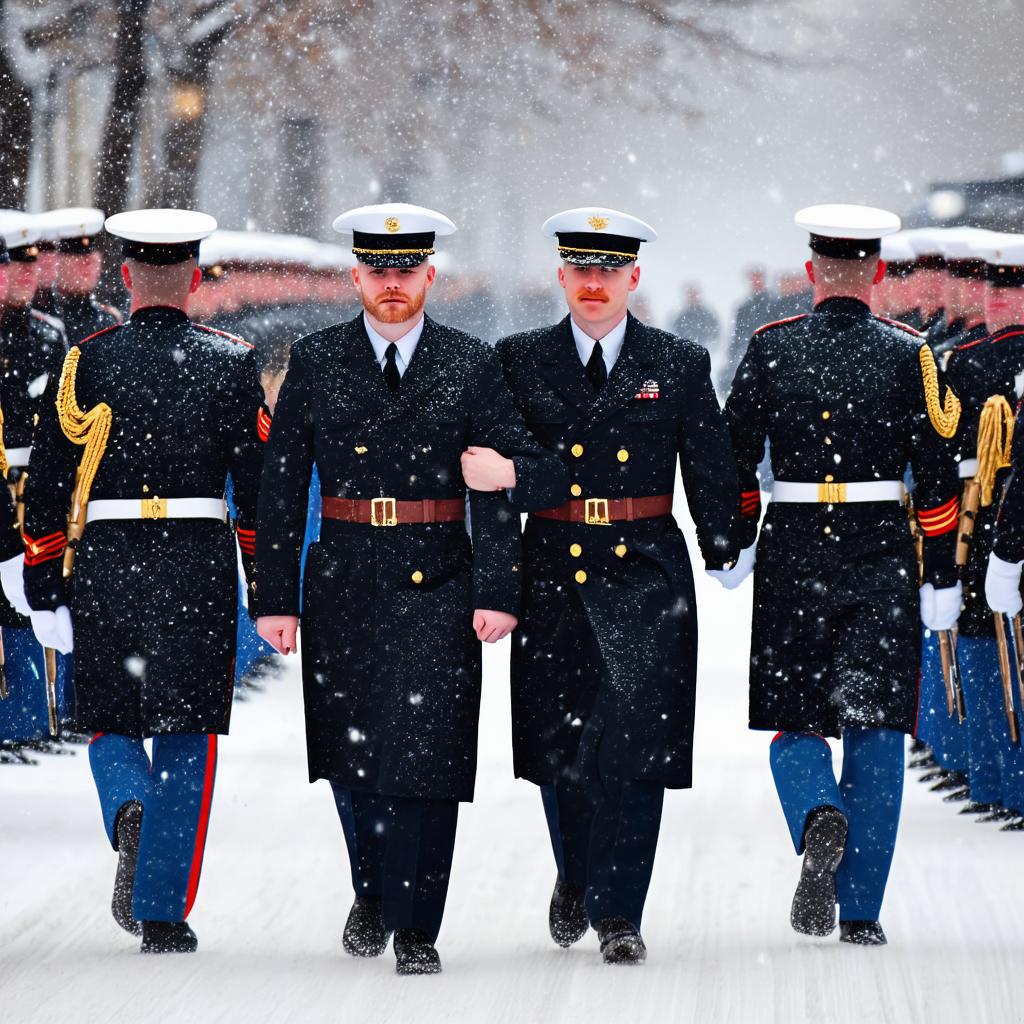}
\newcommand{\ExtSDLargeB}{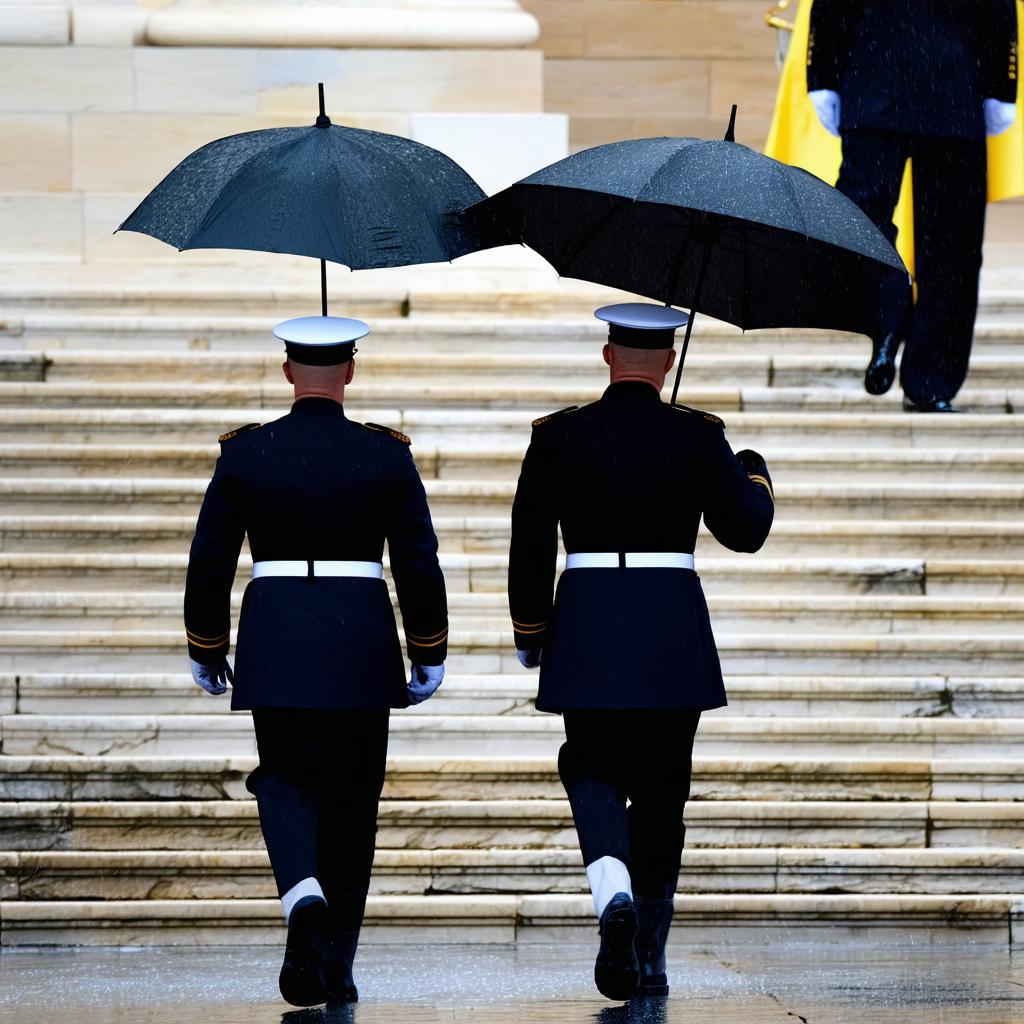}
\newcommand{\ExtSDLargeC}{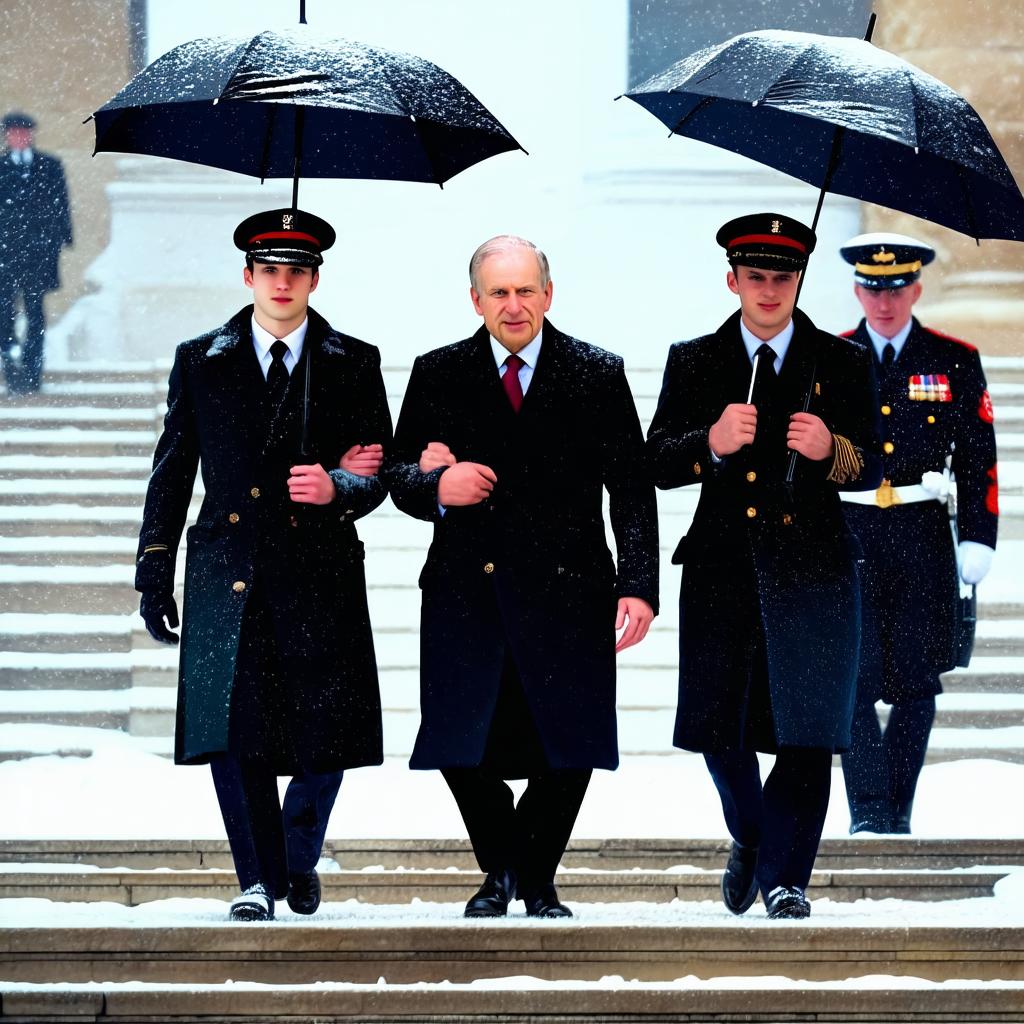}

\newcommand{\ExtSDXLA}{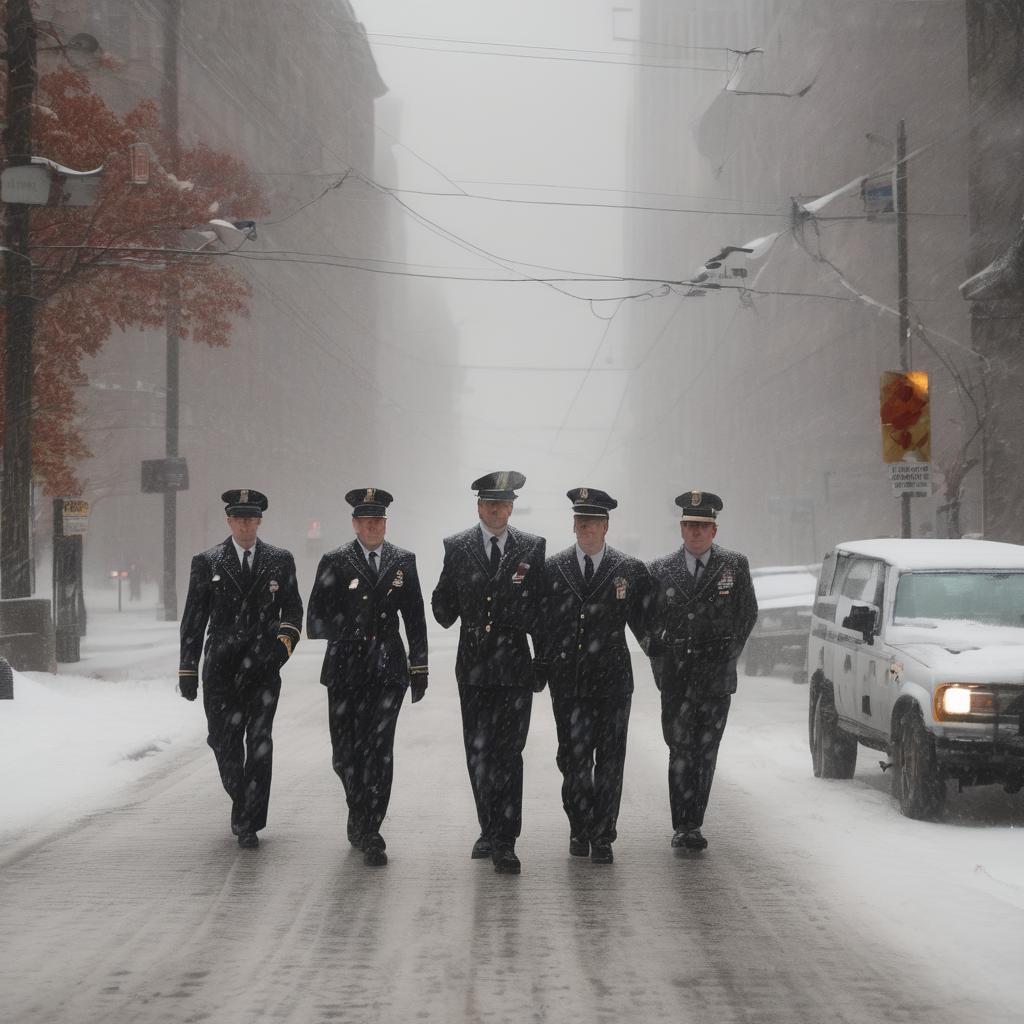}
\newcommand{\ExtSDXLB}{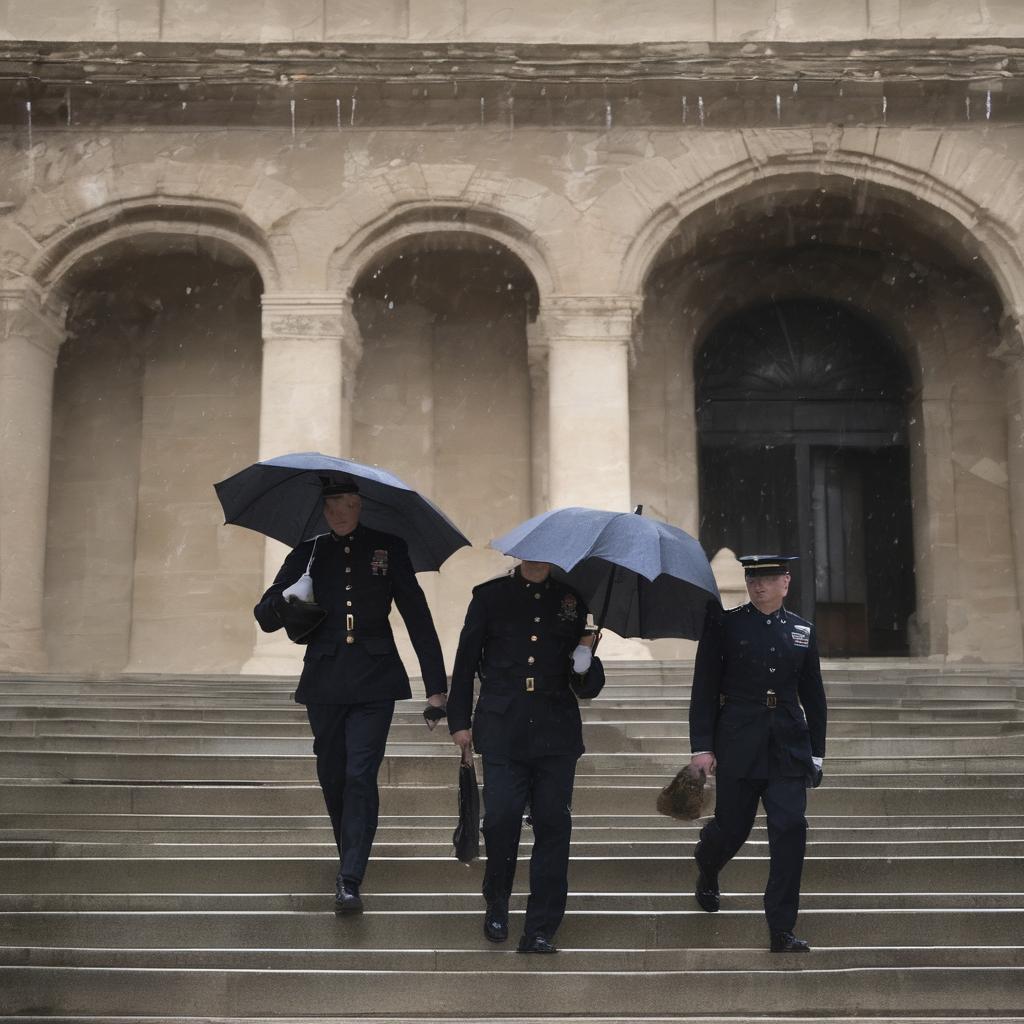}
\newcommand{\ExtSDXLC}{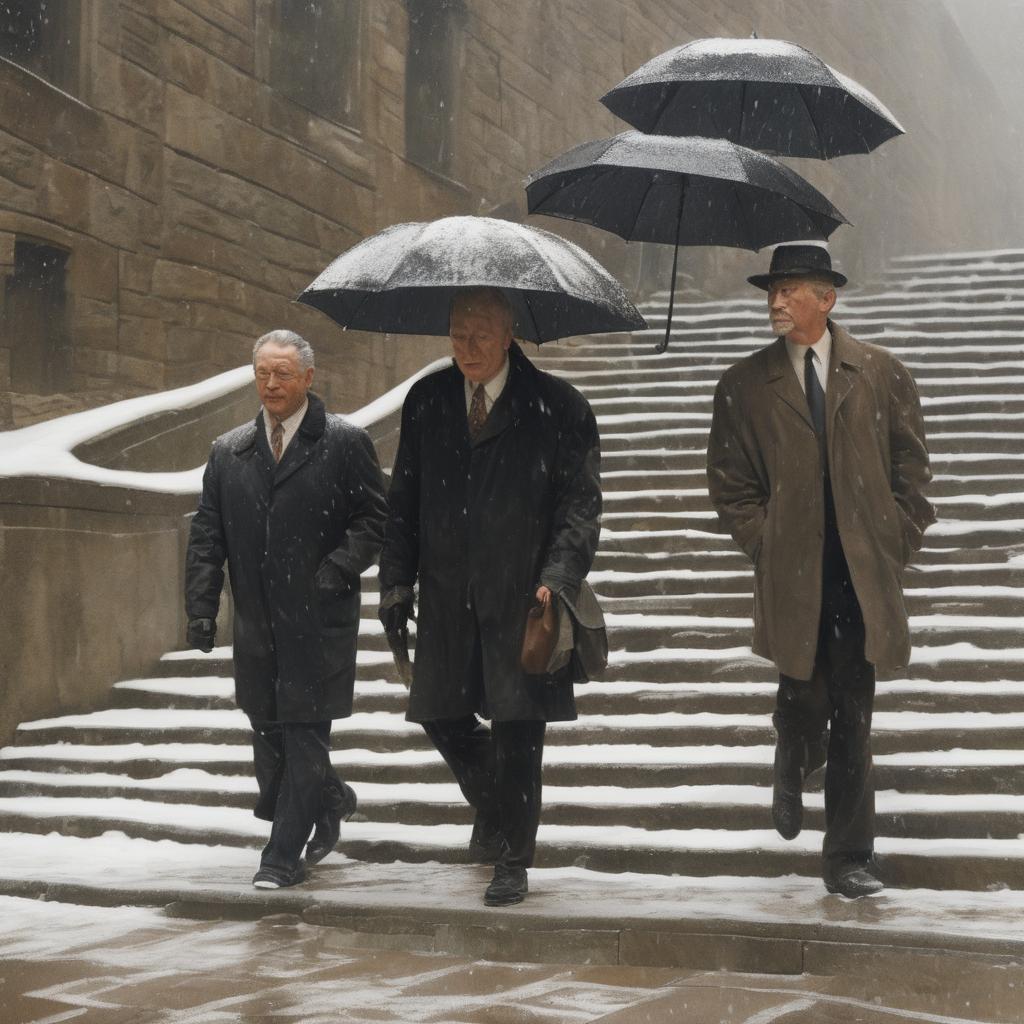}

\newcommand{\ExtFLUXKA}{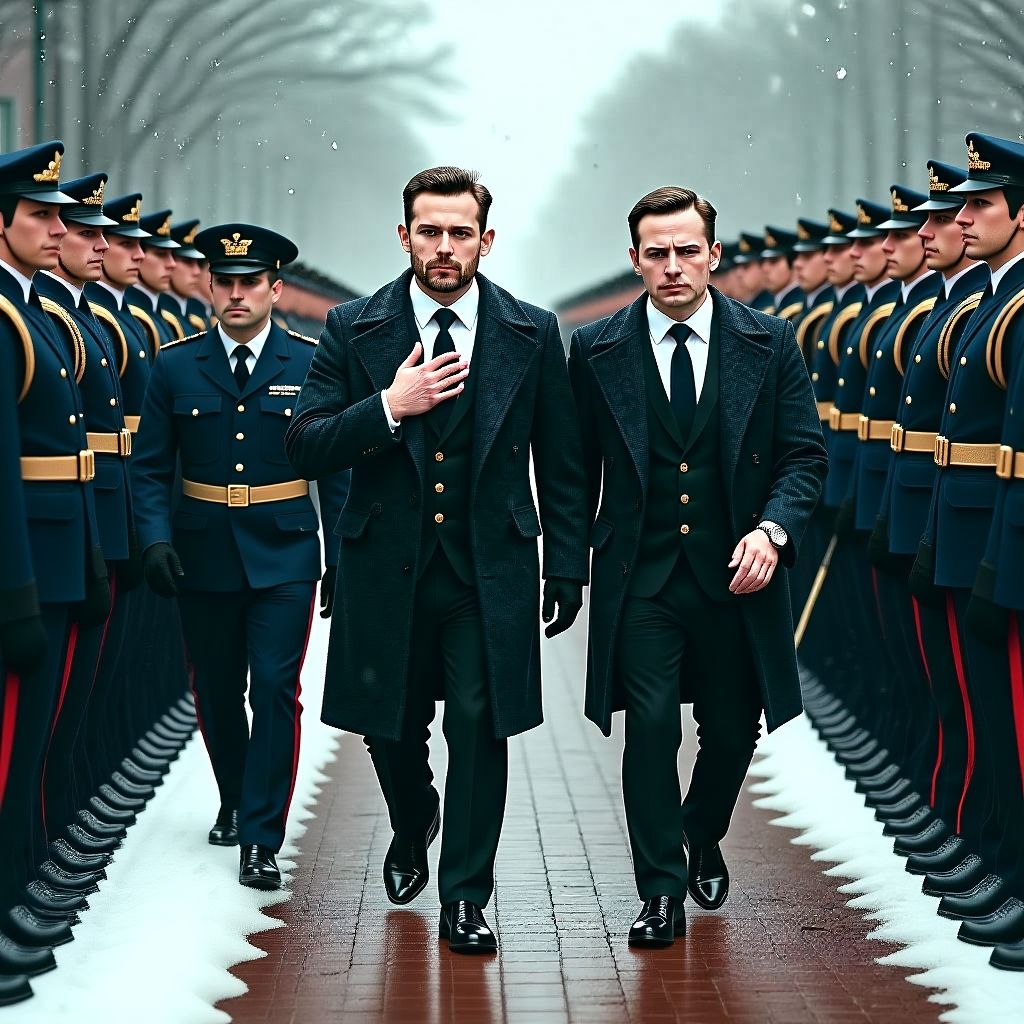}
\newcommand{\ExtFLUXKB}{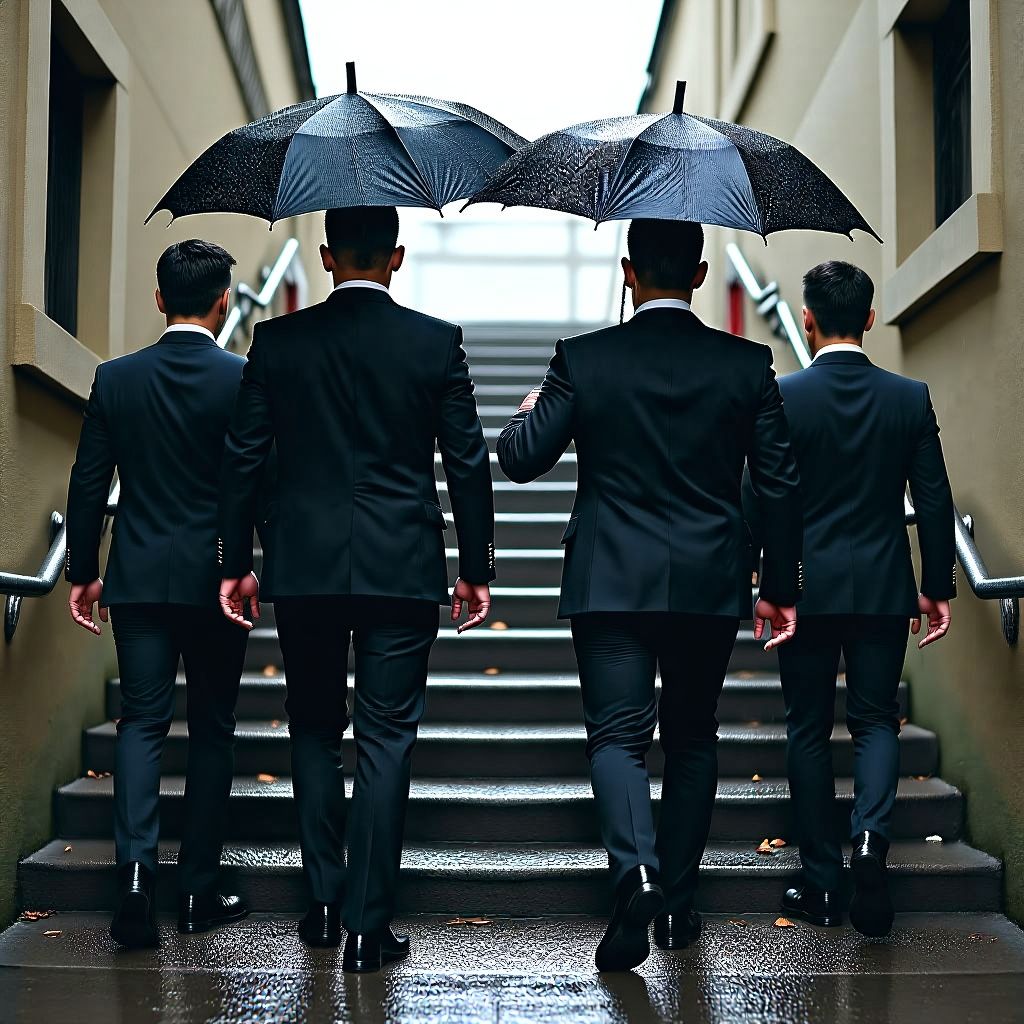}
\newcommand{\ExtFLUXKC}{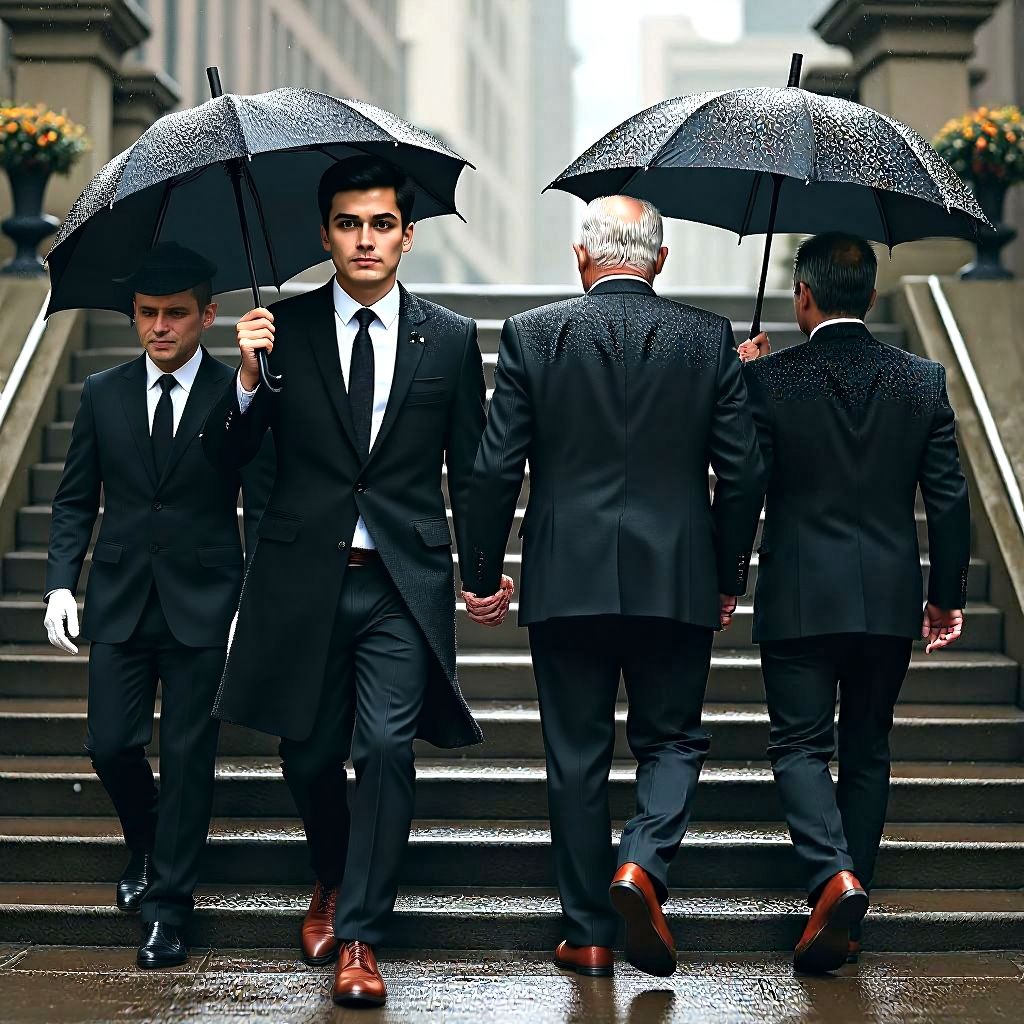}

\newcommand{\ExtFLUXFA}{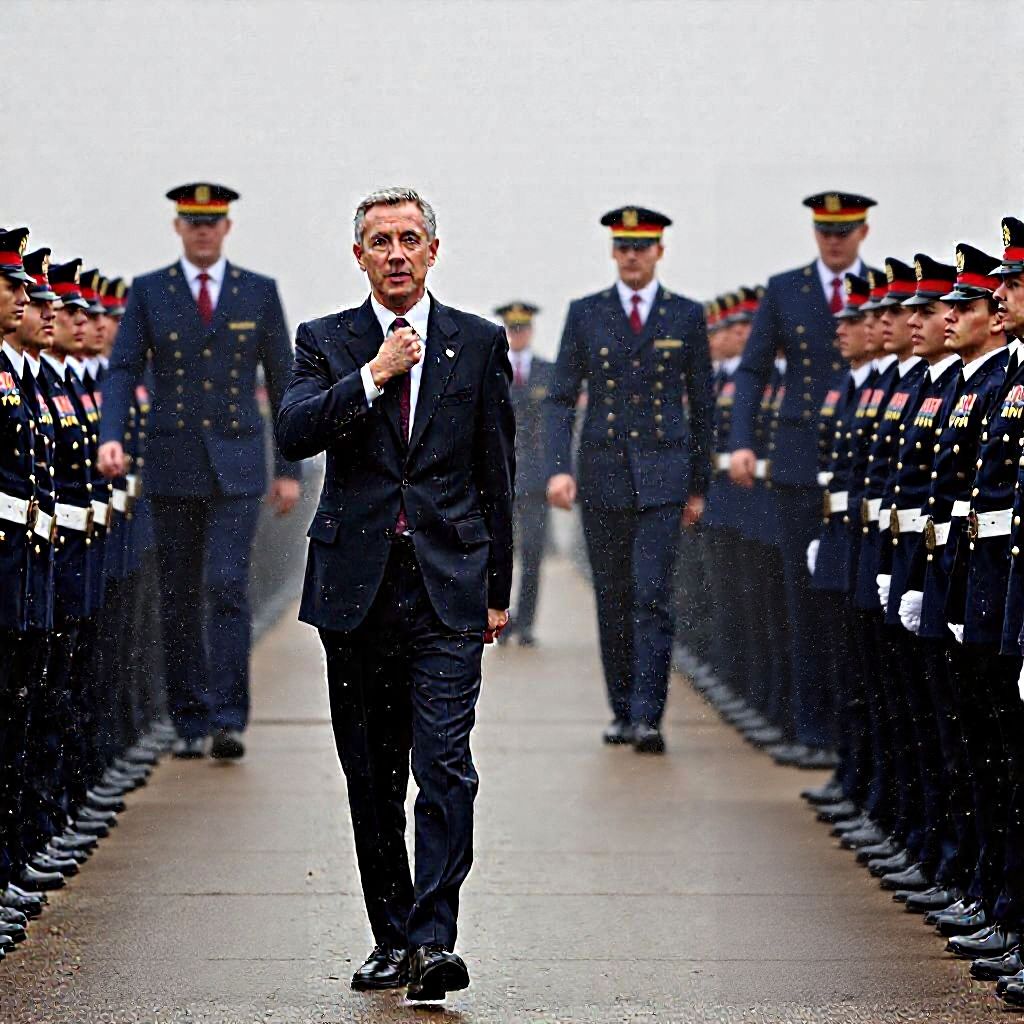}
\newcommand{\ExtFLUXFB}{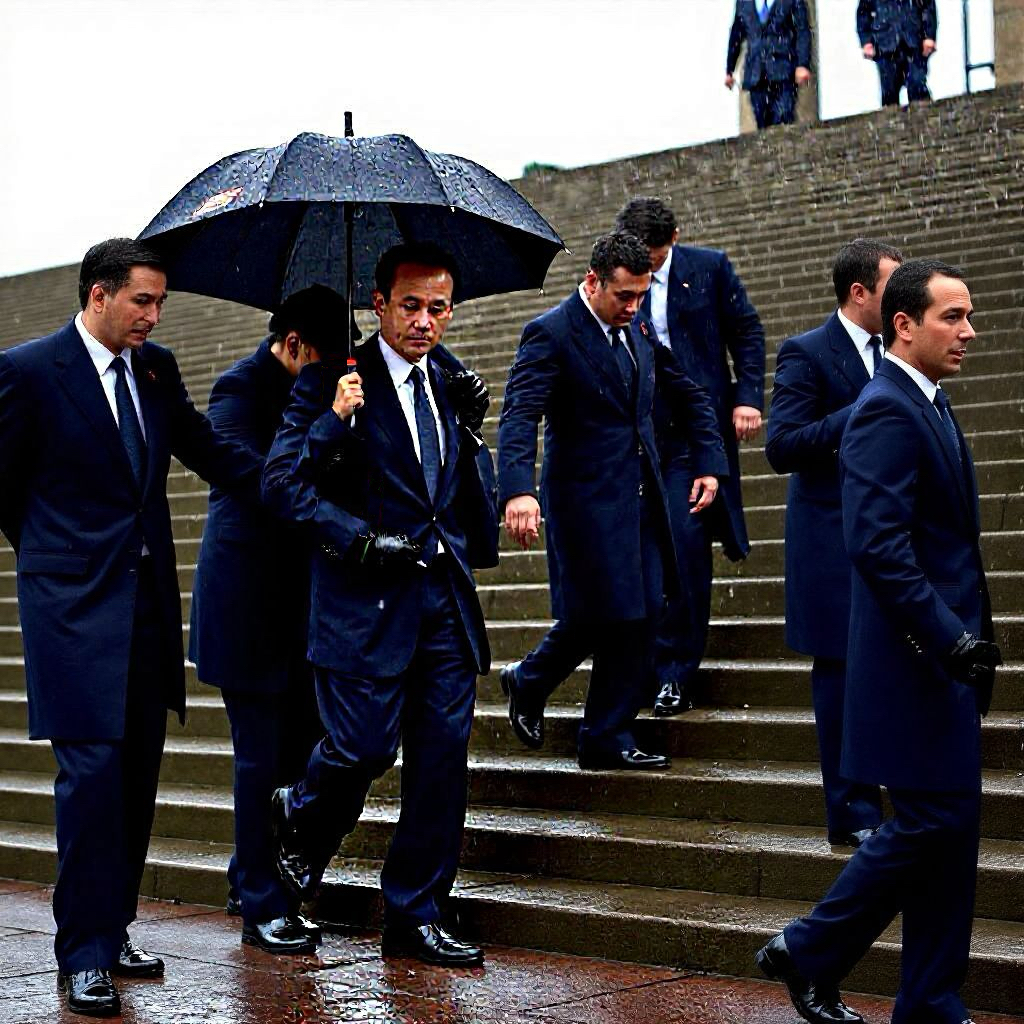}
\newcommand{\ExtFLUXFC}{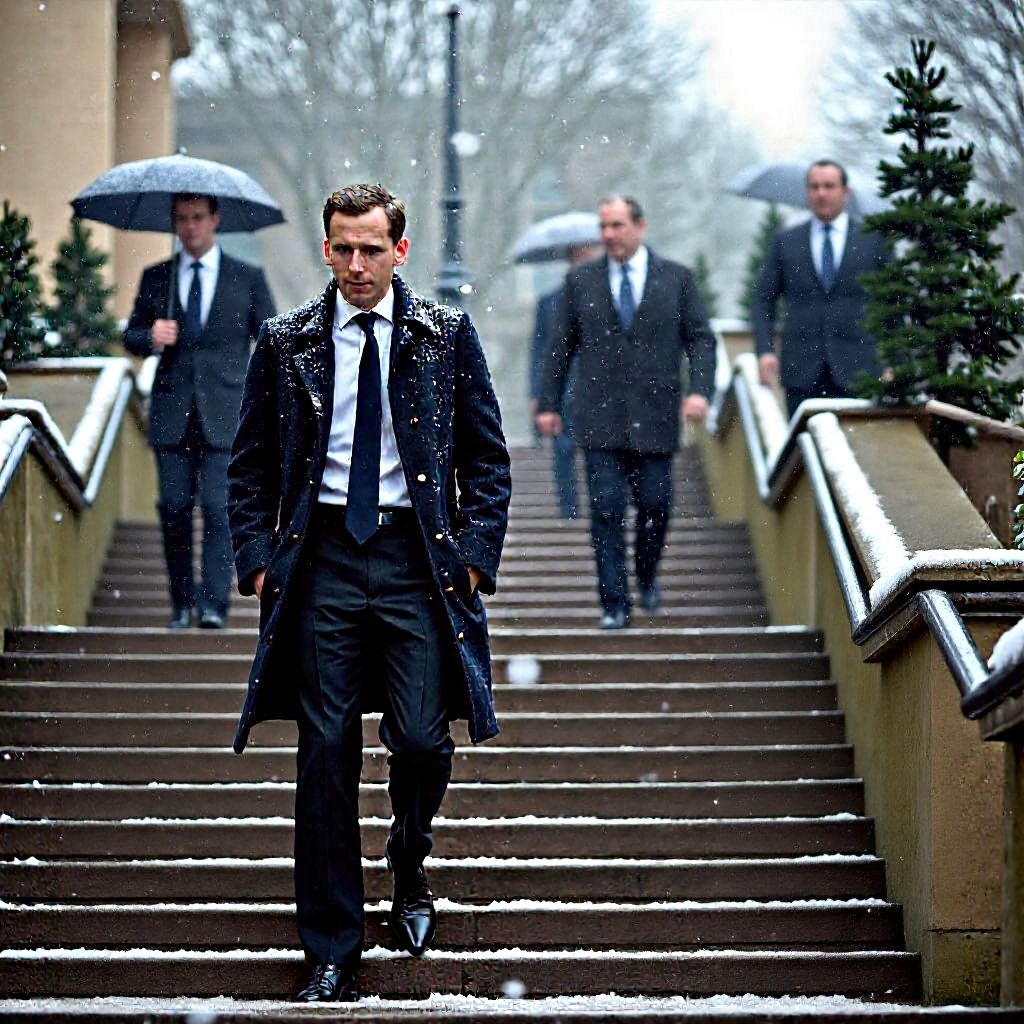}

\newcommand{\ExtCreatiA}{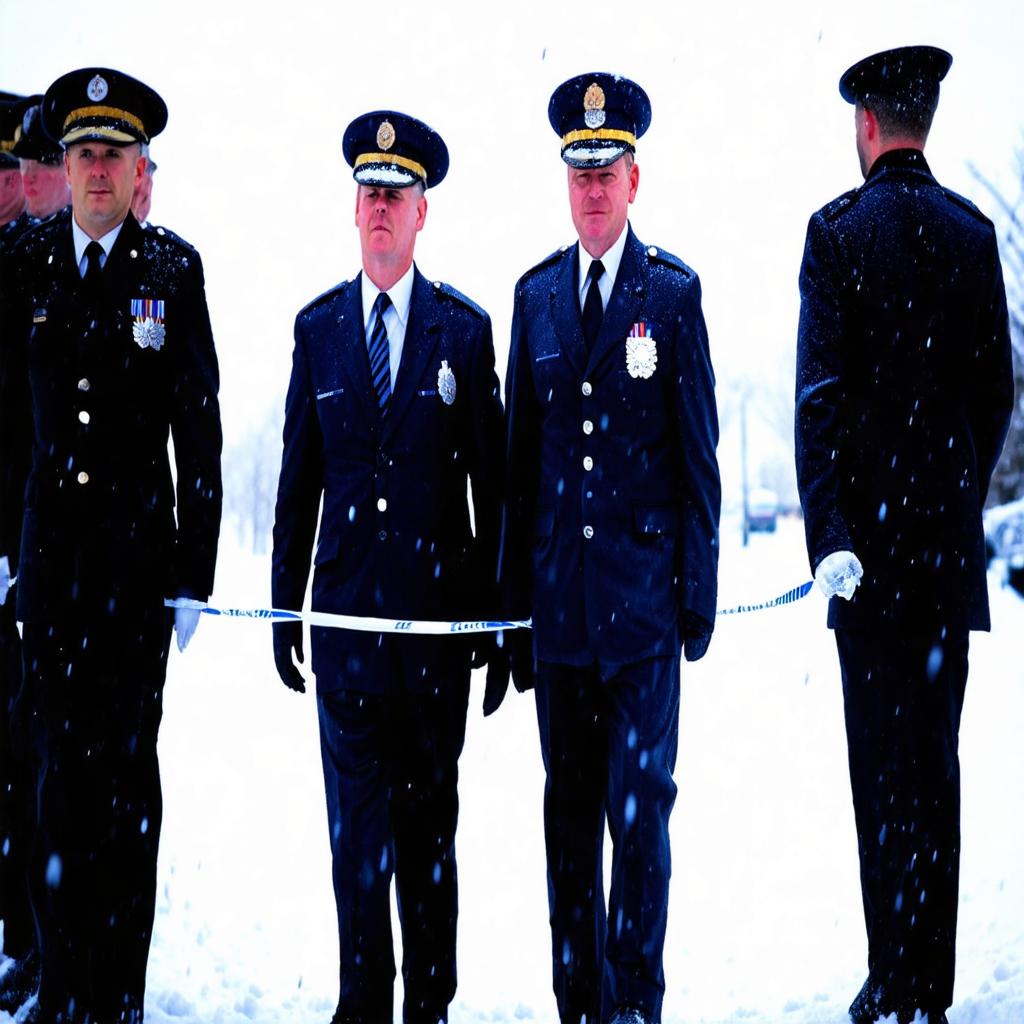}
\newcommand{\ExtCreatiB}{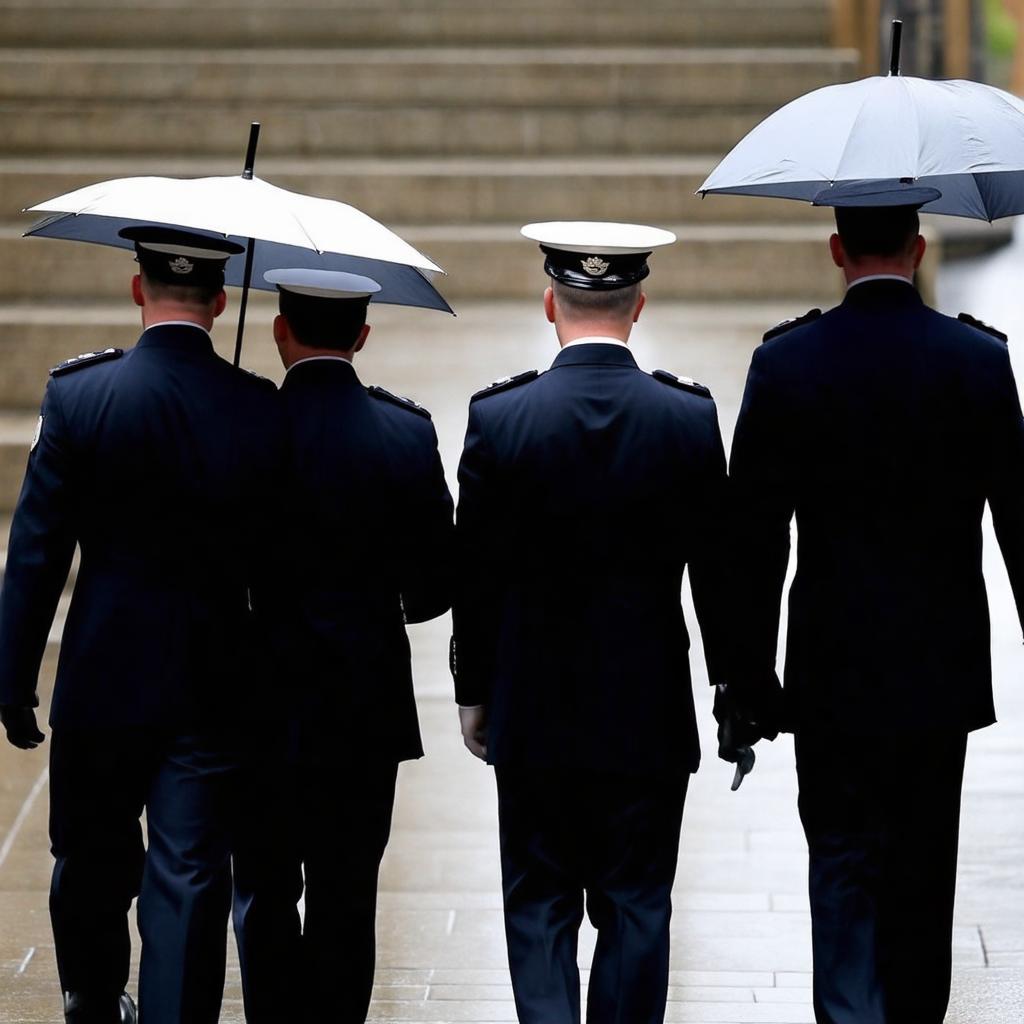}
\newcommand{\ExtCreatiC}{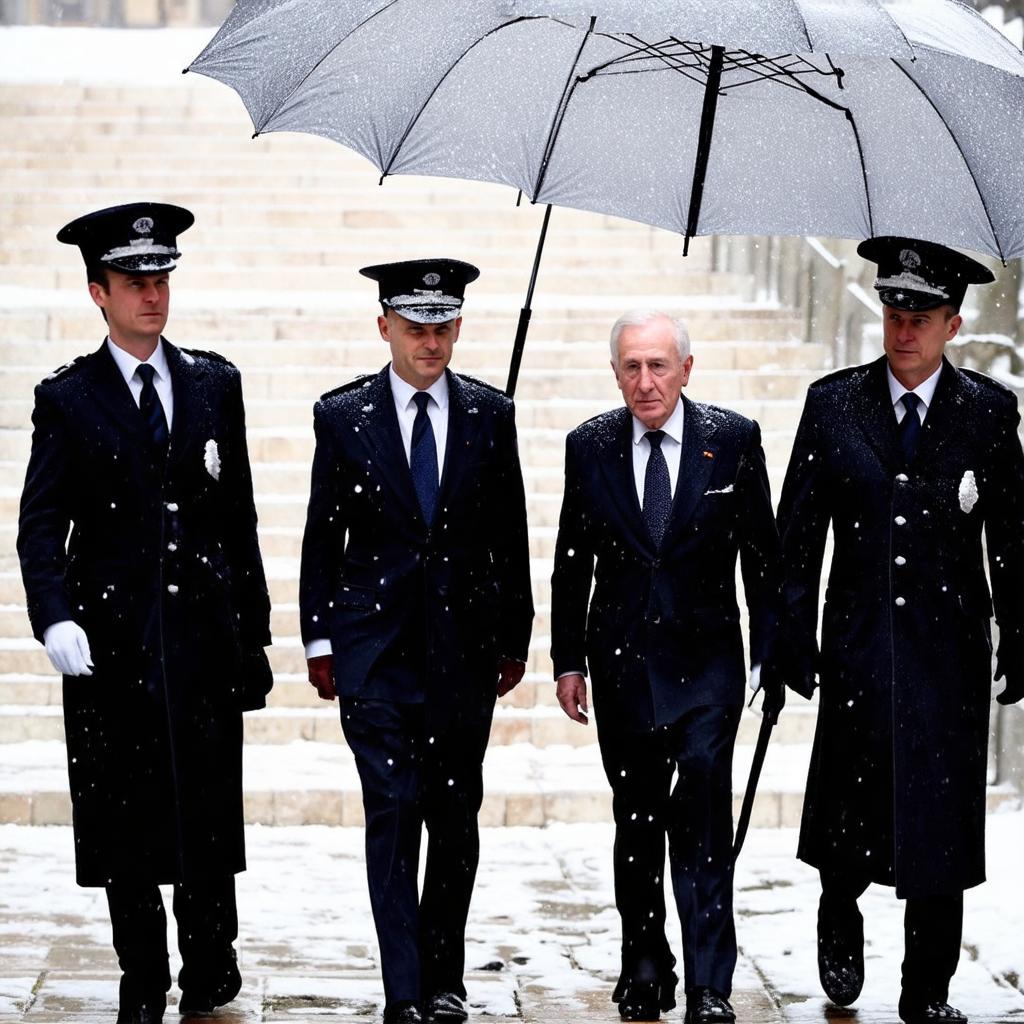}

\newcommand{\ExtRealCPA}{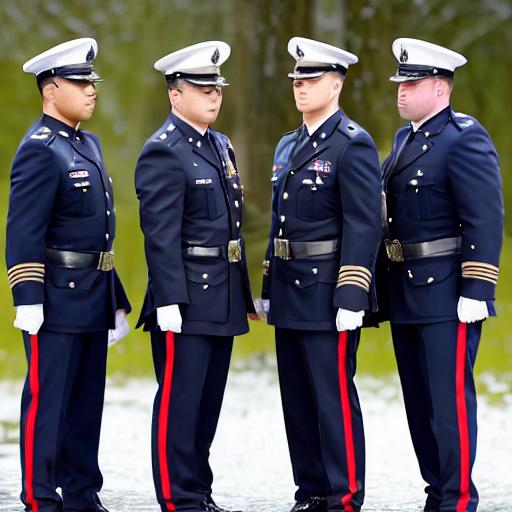}
\newcommand{\ExtRealCPB}{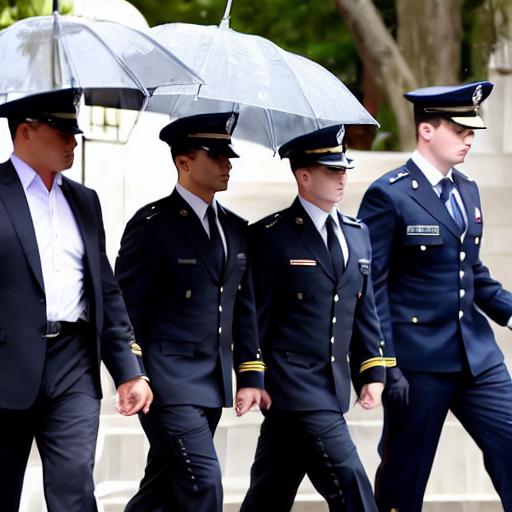}
\newcommand{\ExtRealCPC}{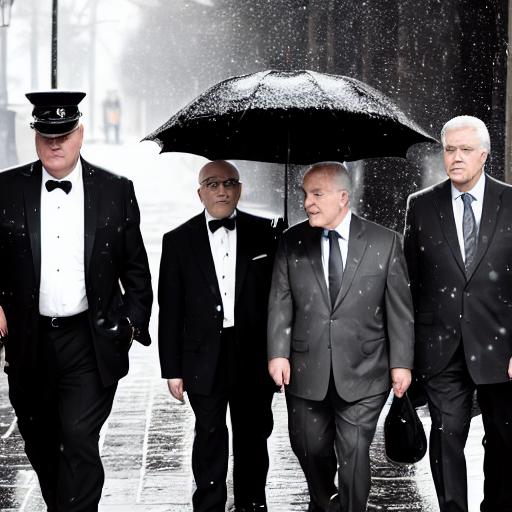}

\newcommand{\ExtTwoOursA}{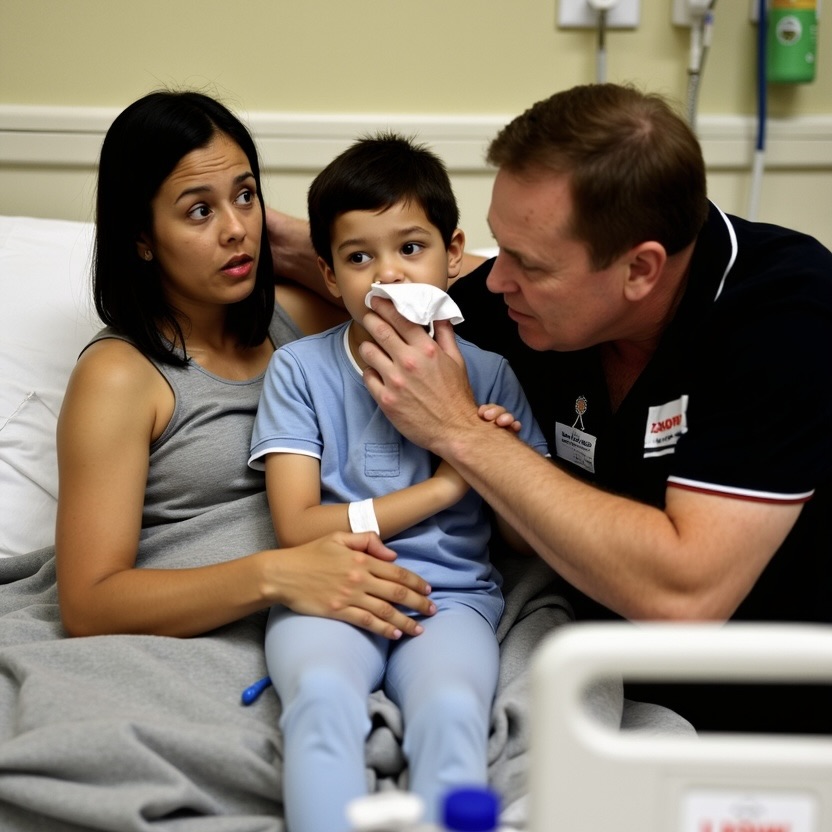}
\newcommand{\ExtTwoOursB}{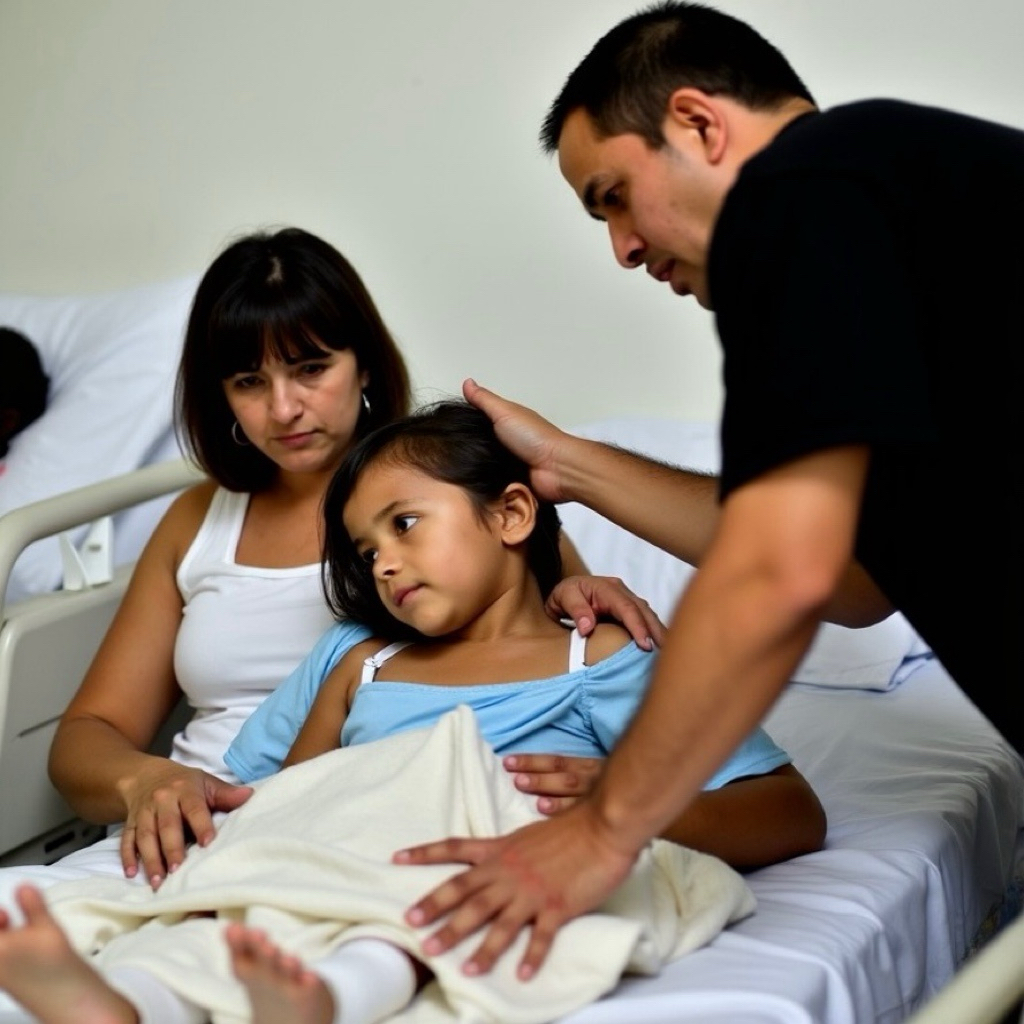}
\newcommand{\ExtTwoOursC}{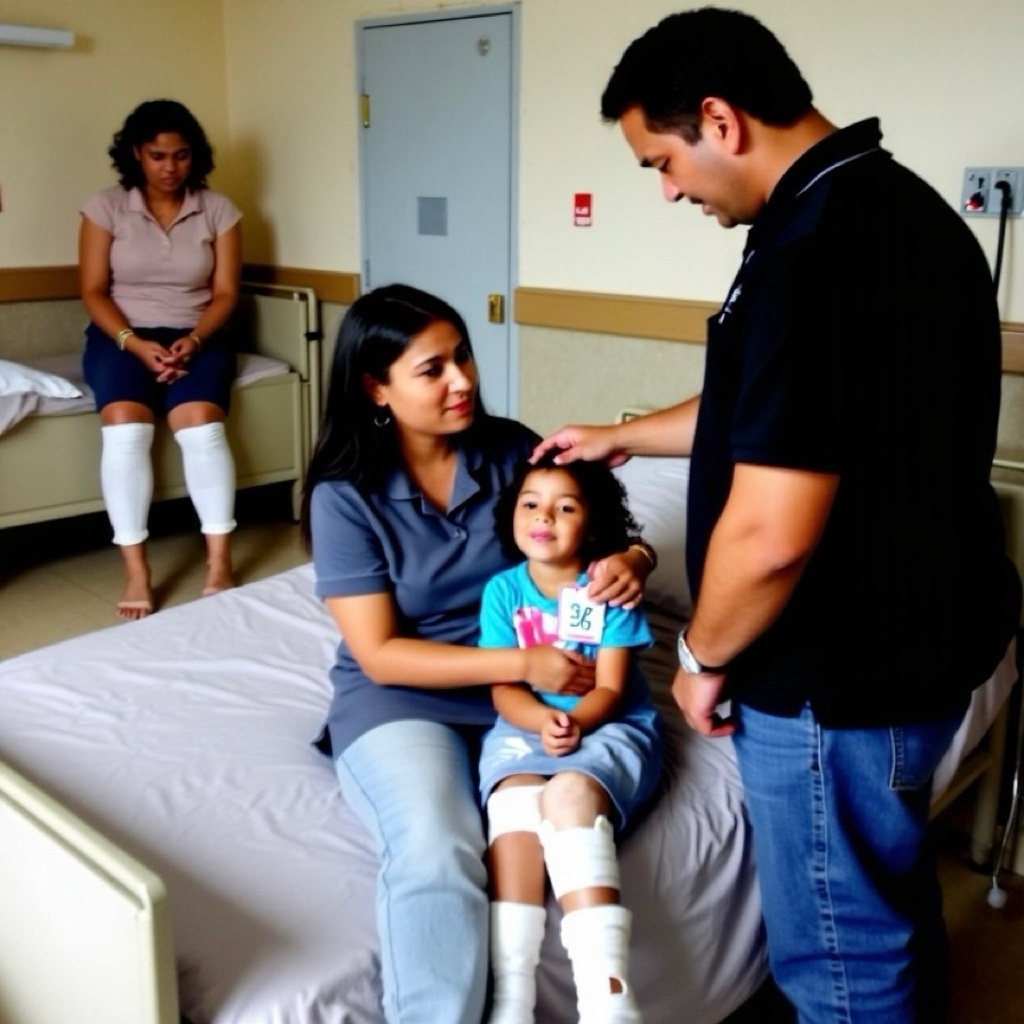}

\newcommand{\ExtTwoBaseA}{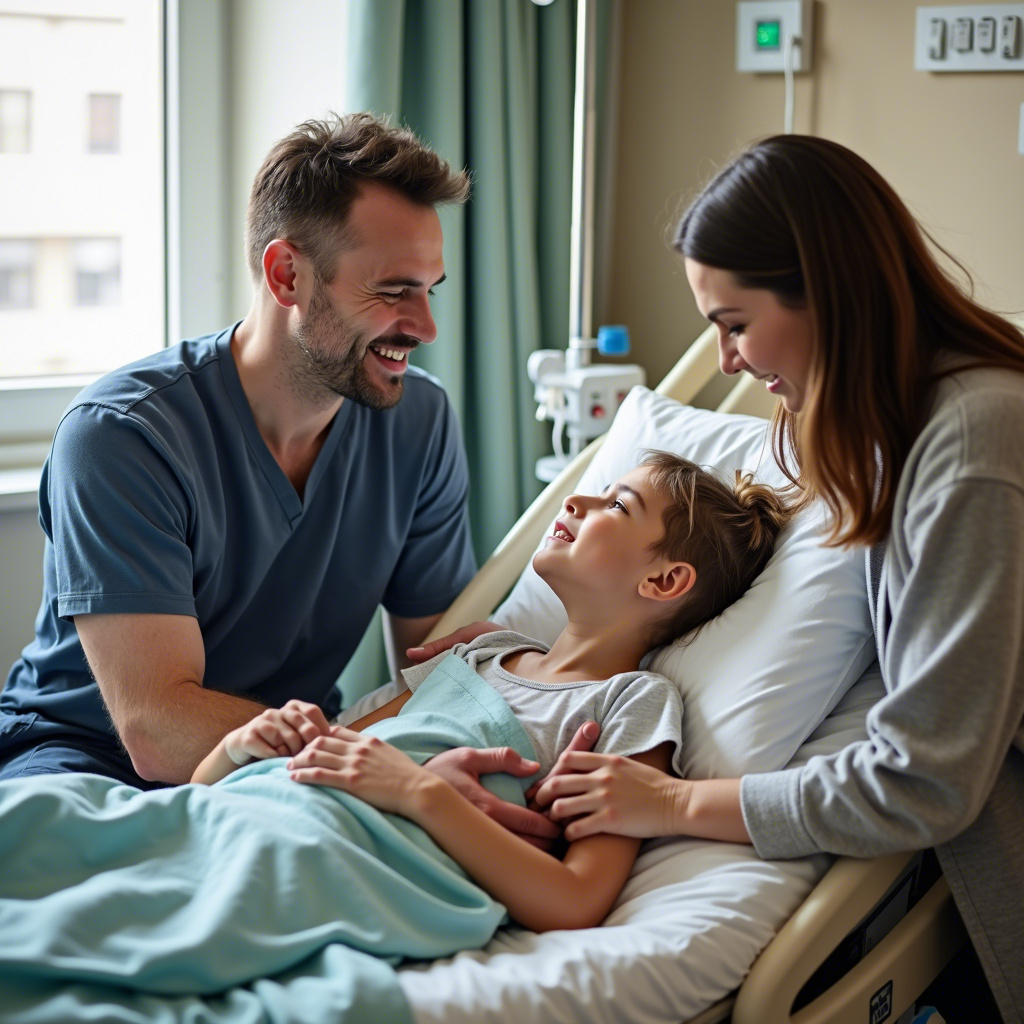}
\newcommand{\ExtTwoBaseB}{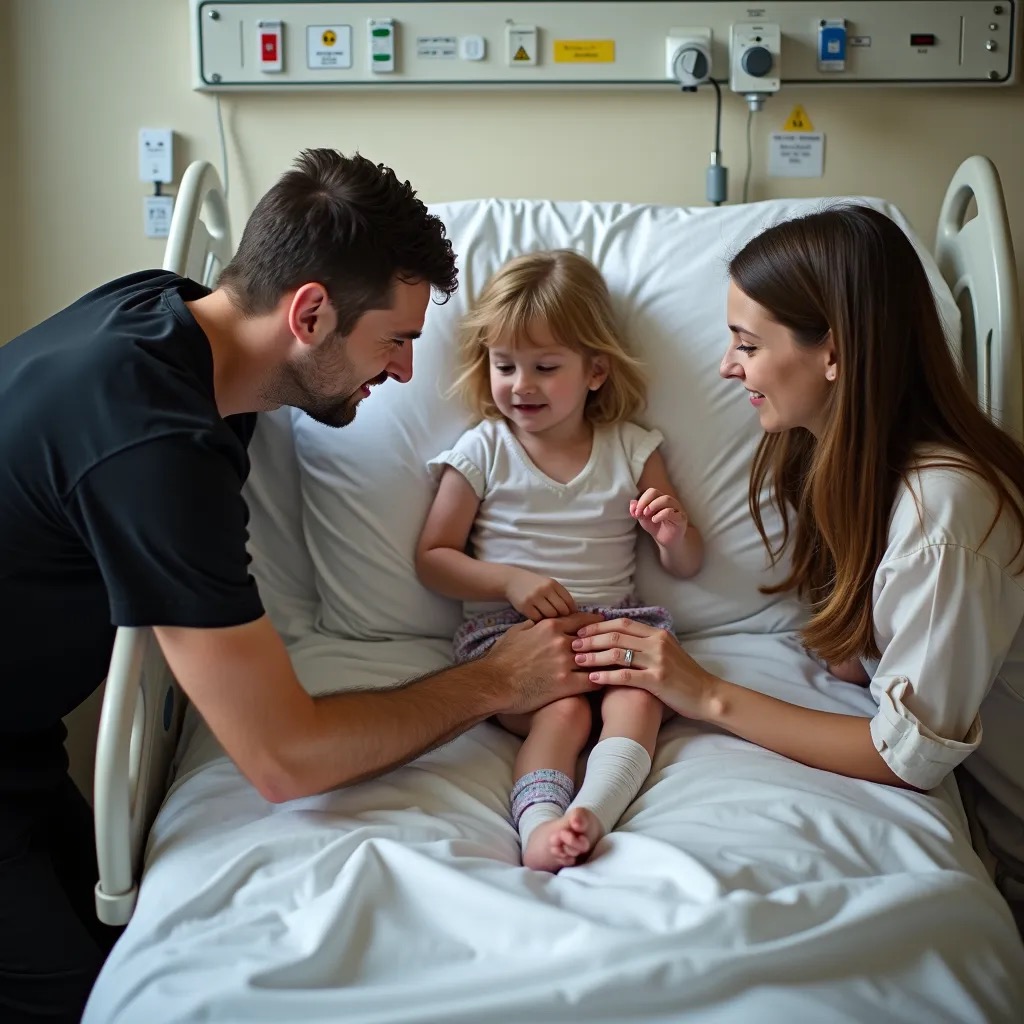}
\newcommand{\ExtTwoBaseC}{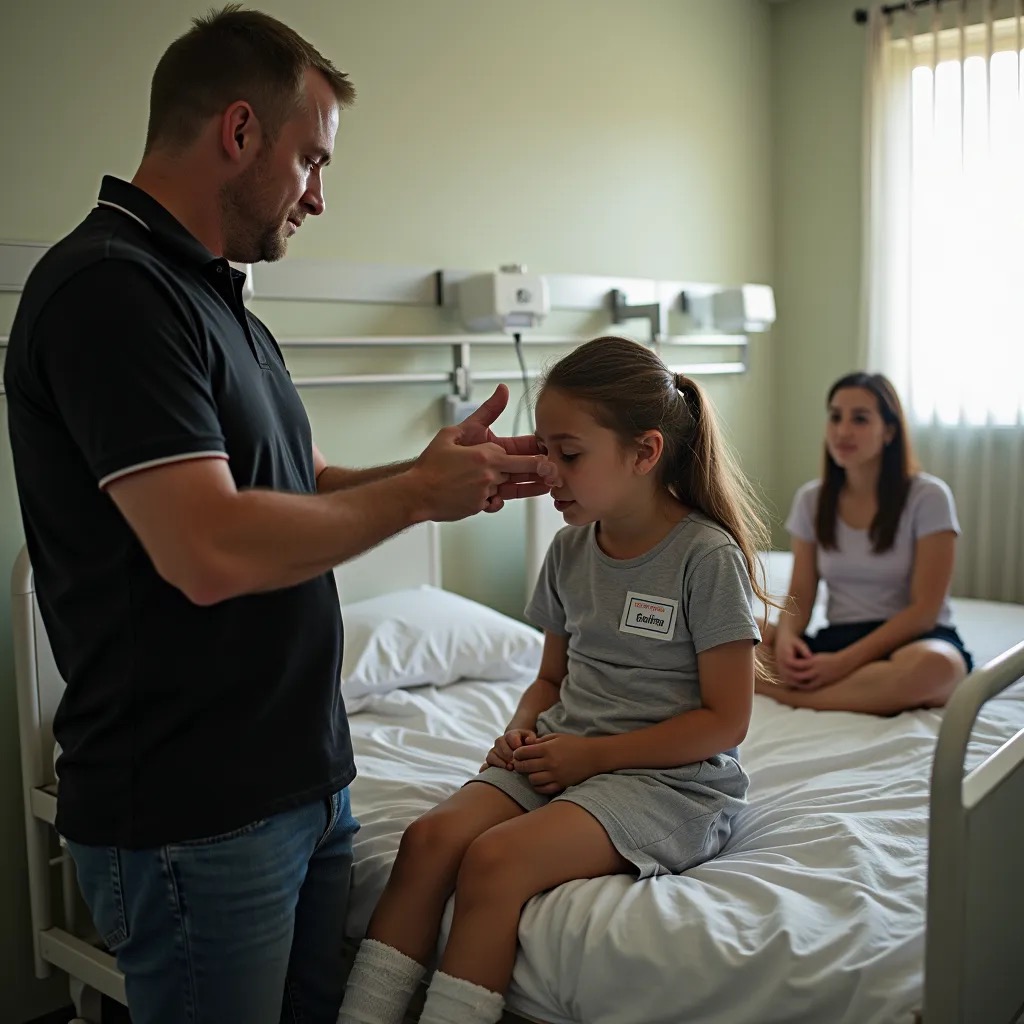}

\newcommand{\ExtTwoSDLargeA}{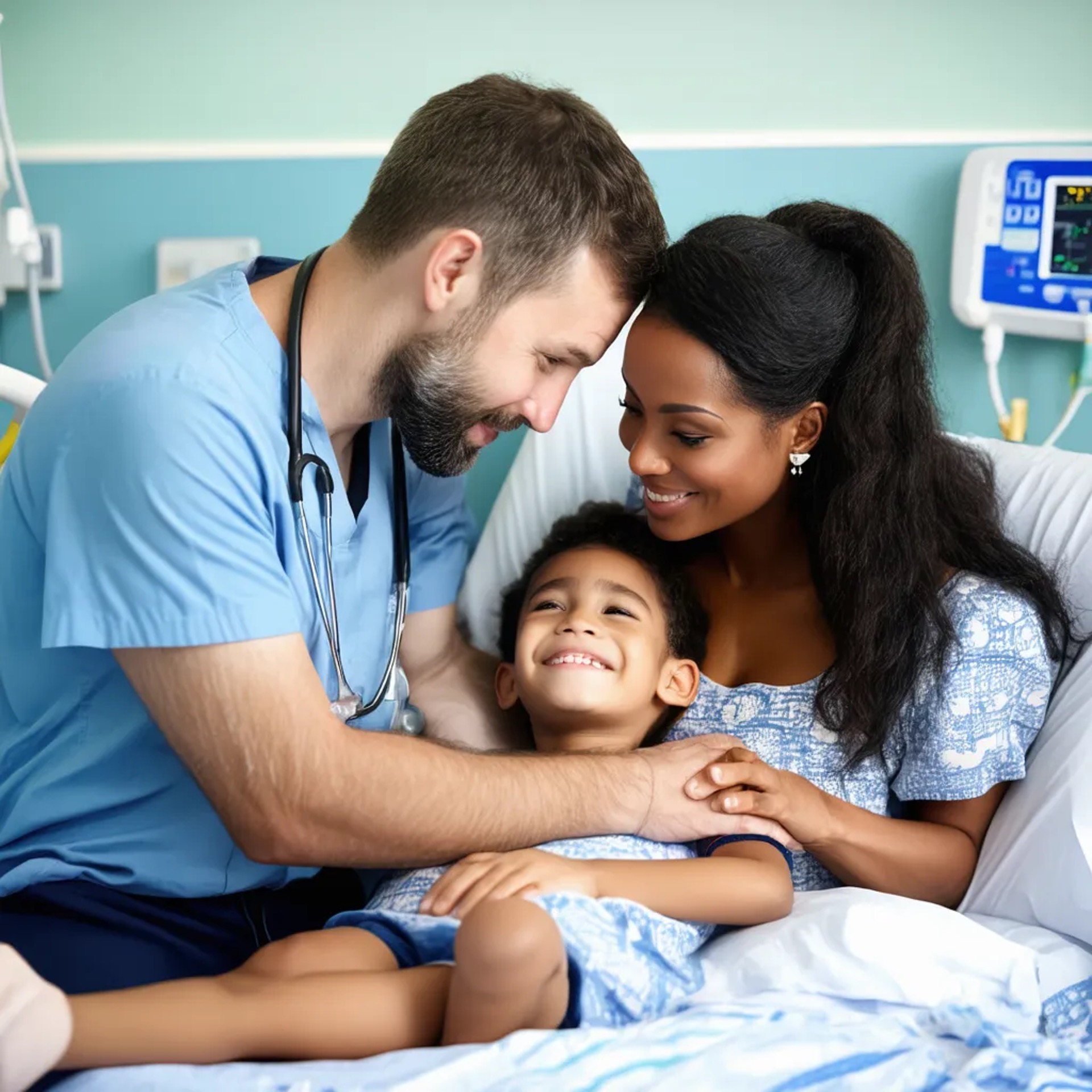}
\newcommand{\ExtTwoSDLargeB}{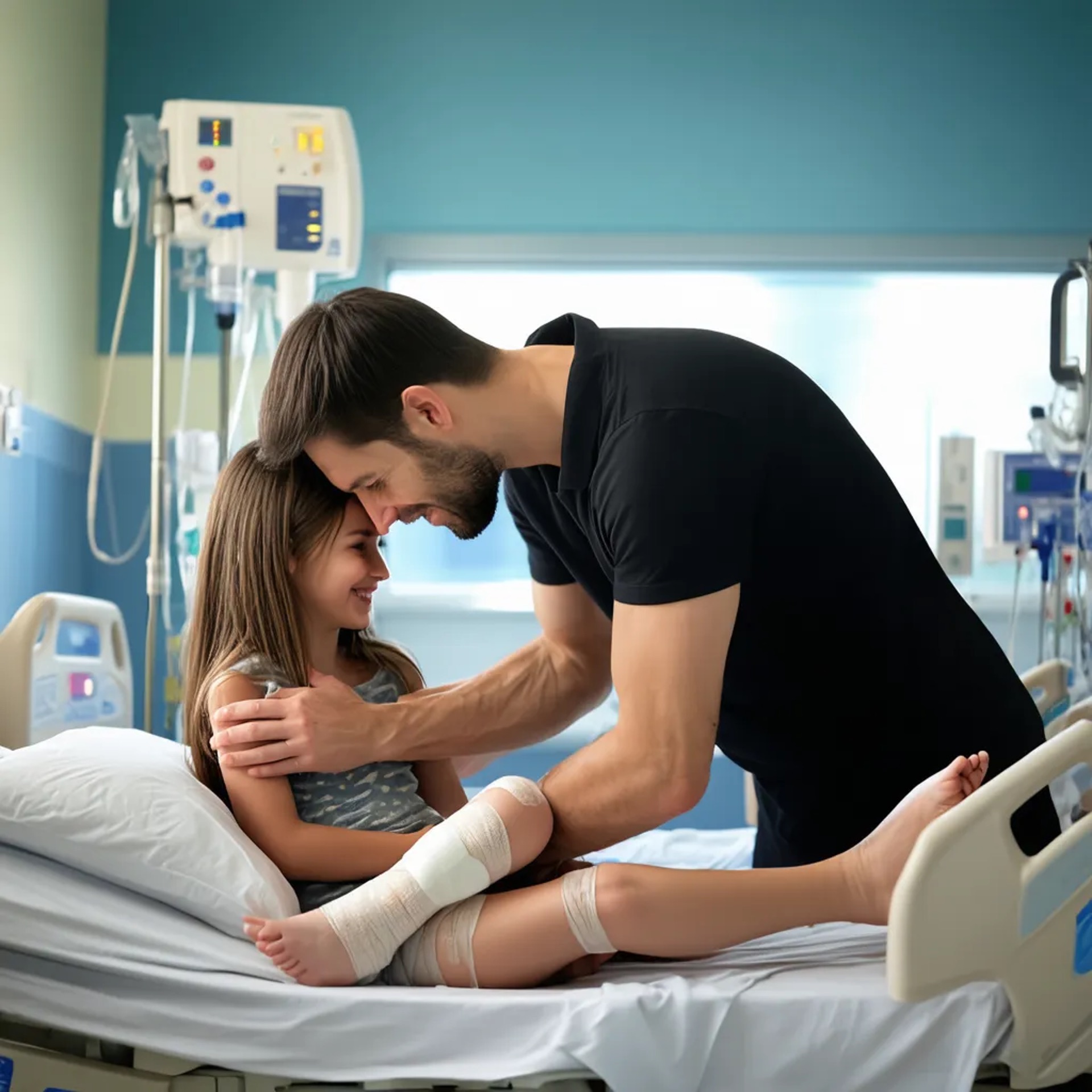}
\newcommand{\ExtTwoSDLargeC}{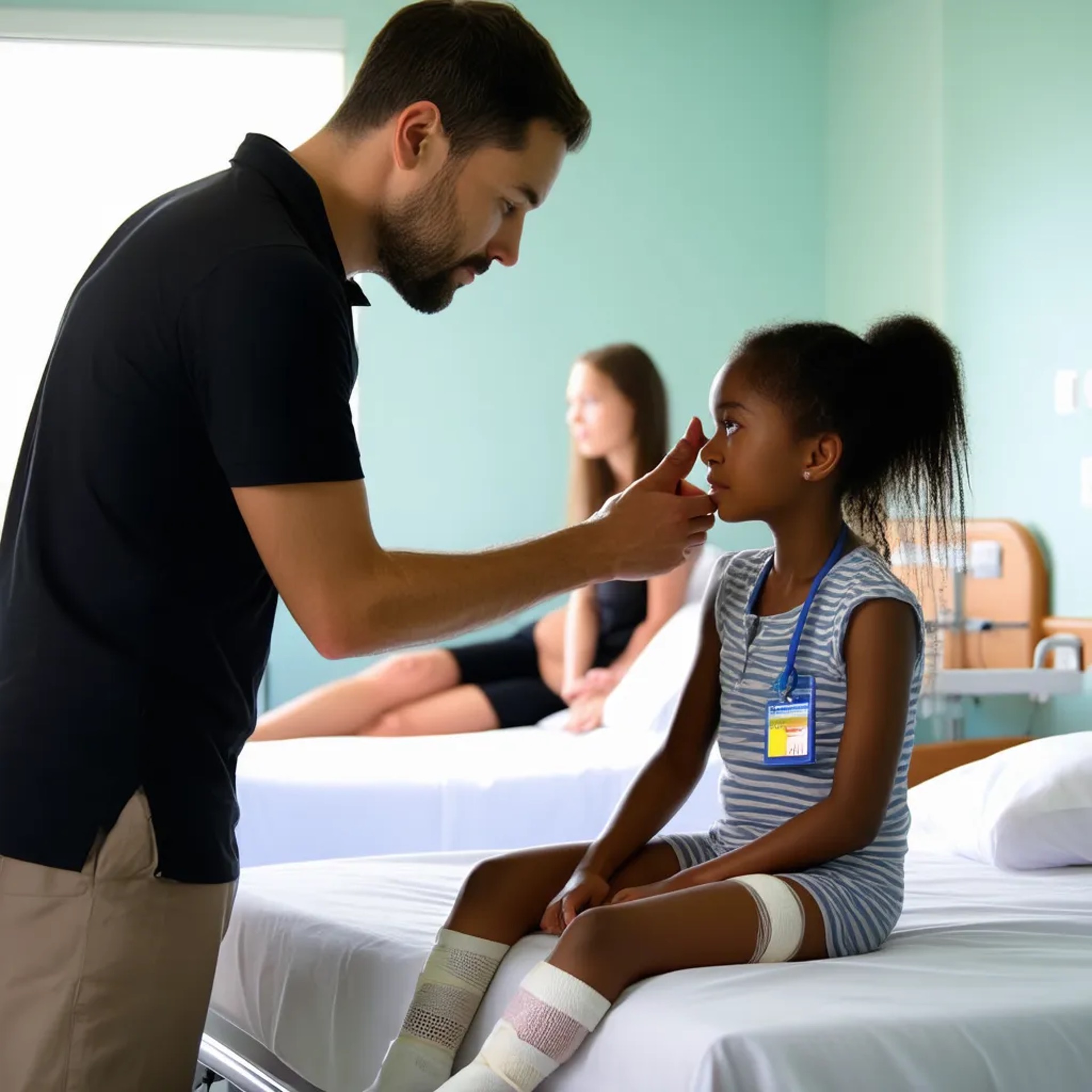}

\newcommand{\ExtTwoSDXLA}{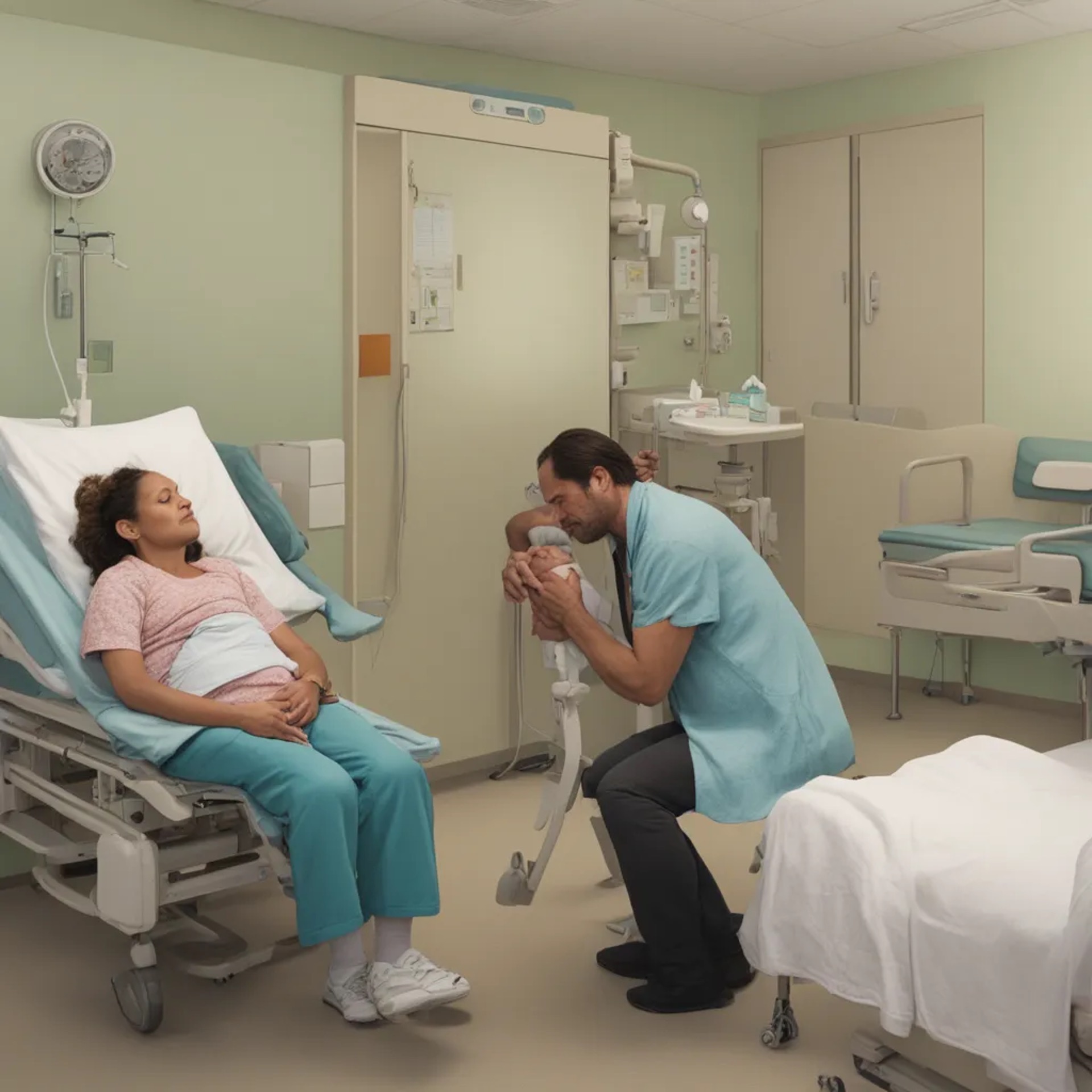}
\newcommand{\ExtTwoSDXLB}{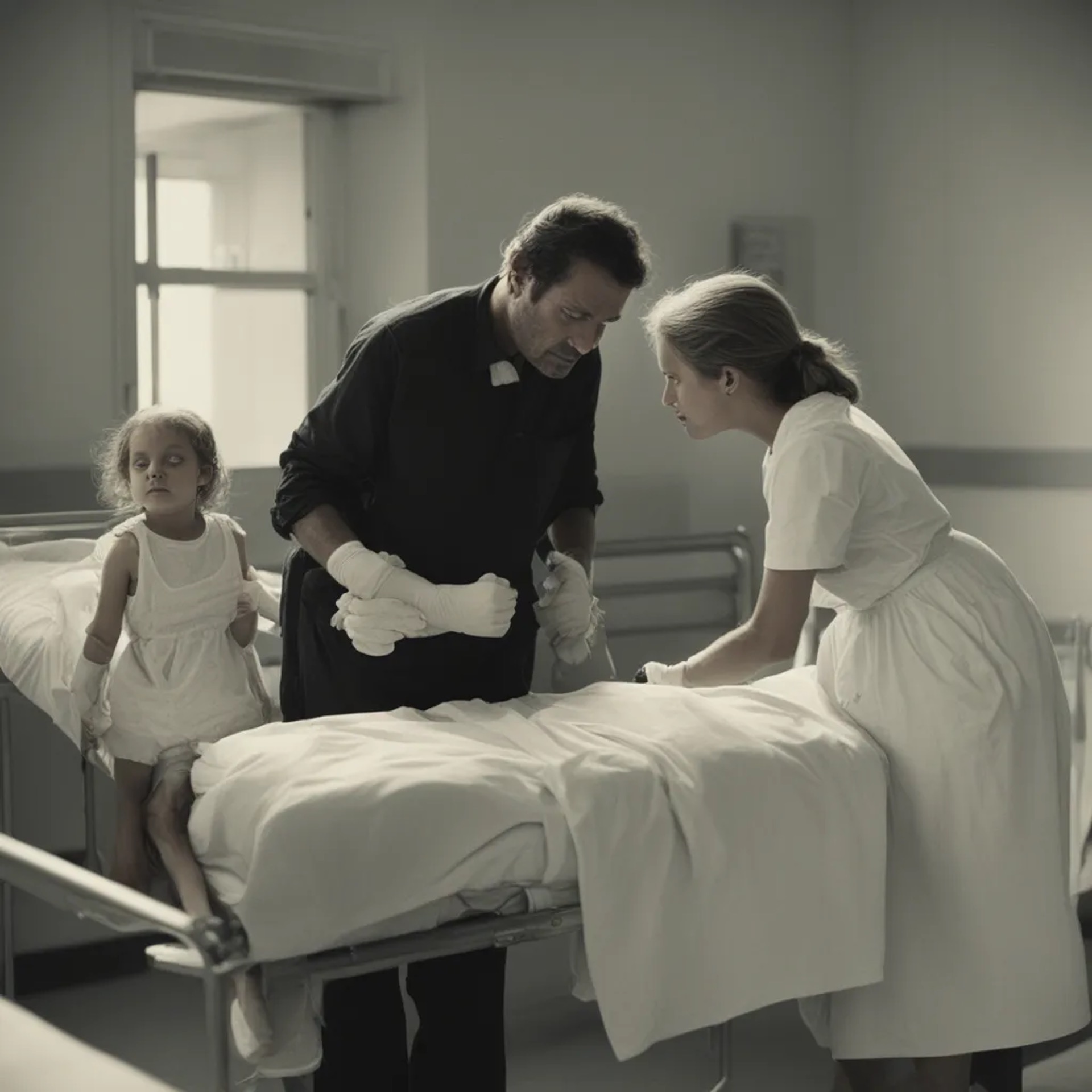}
\newcommand{\ExtTwoSDXLC}{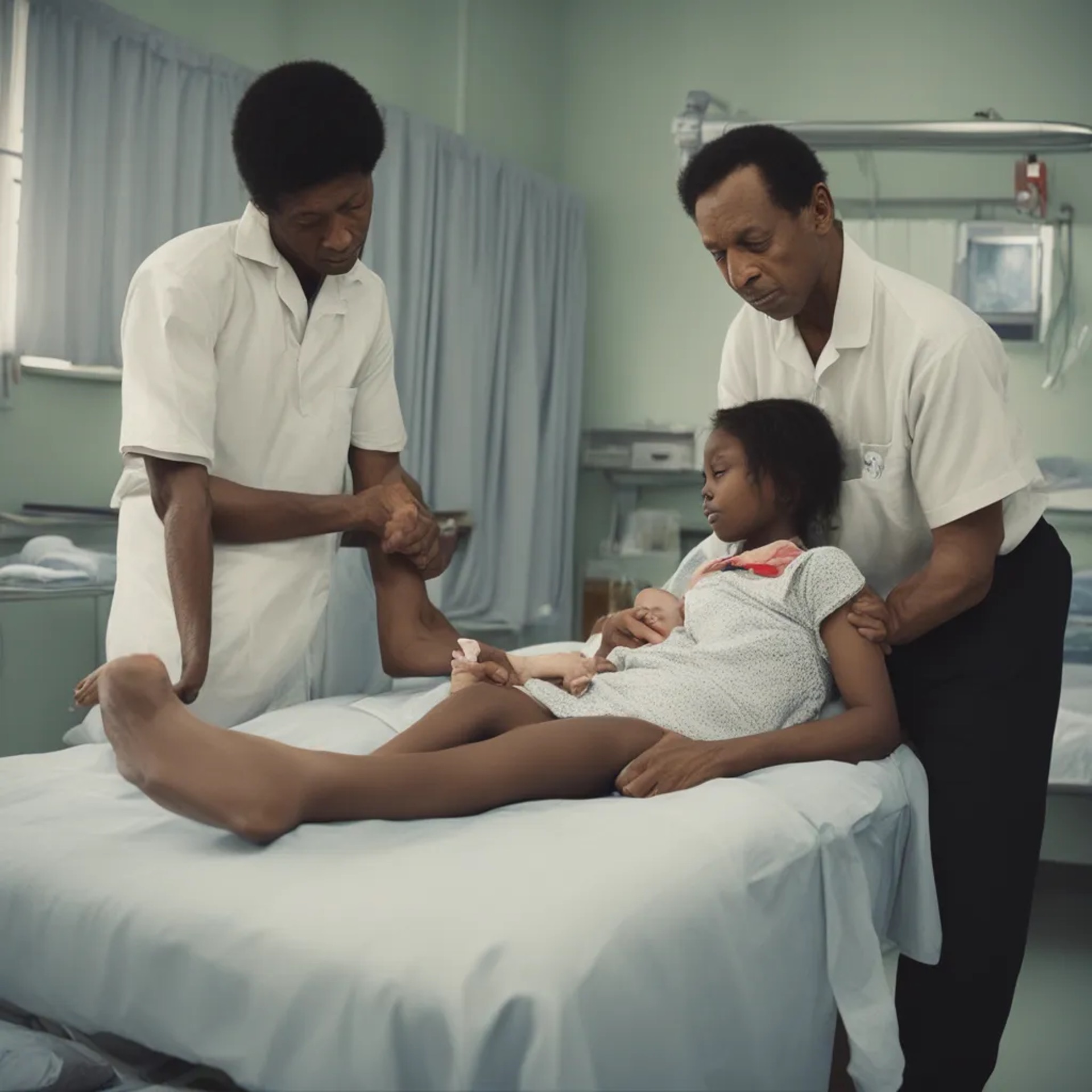}

\newcommand{\ExtTwoFLUXKA}{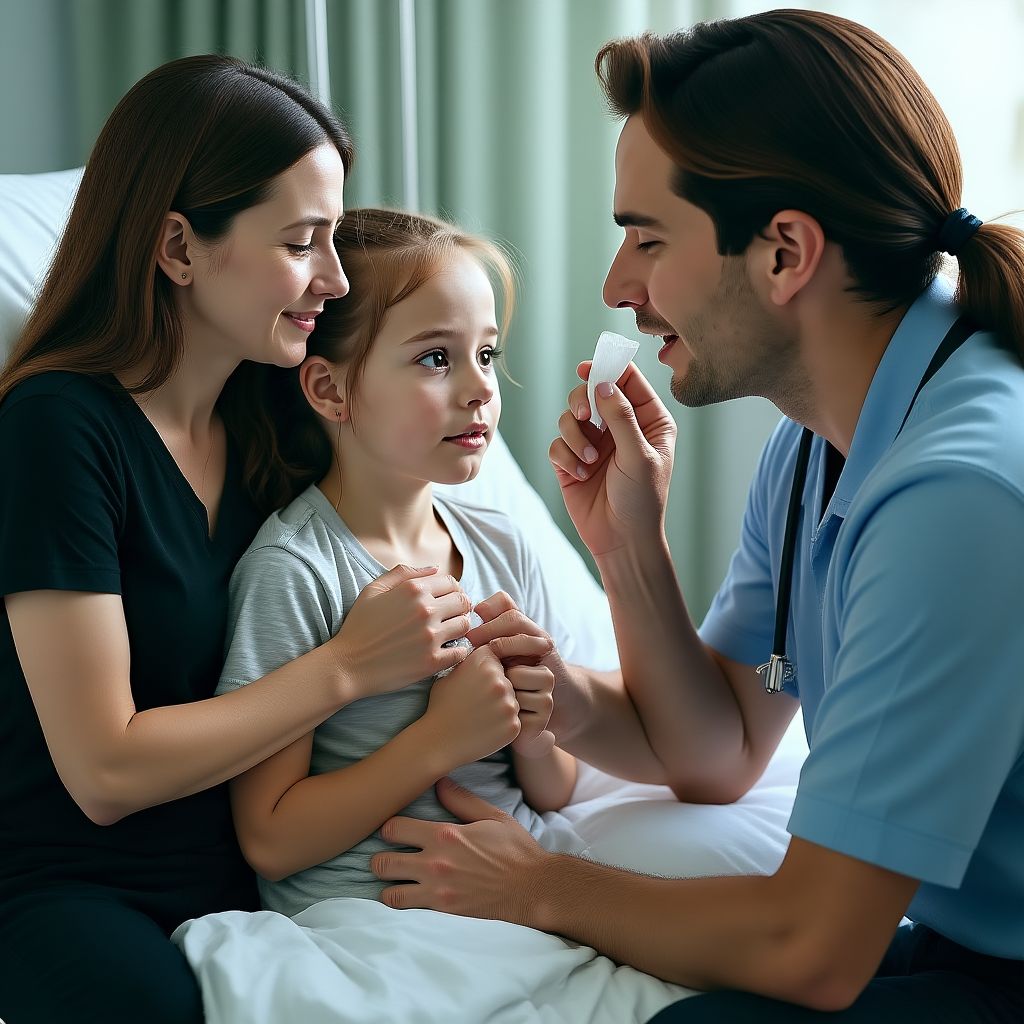}
\newcommand{\ExtTwoFLUXKB}{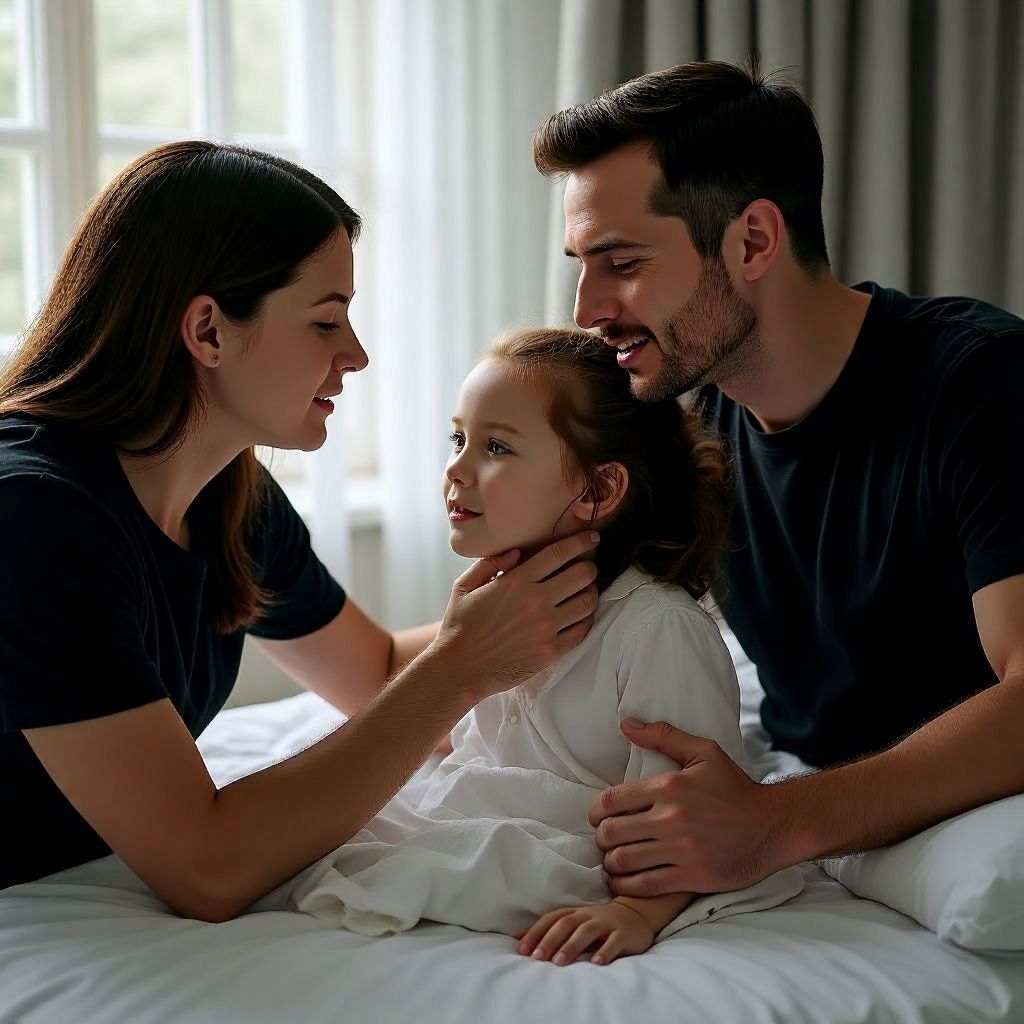}
\newcommand{\ExtTwoFLUXKC}{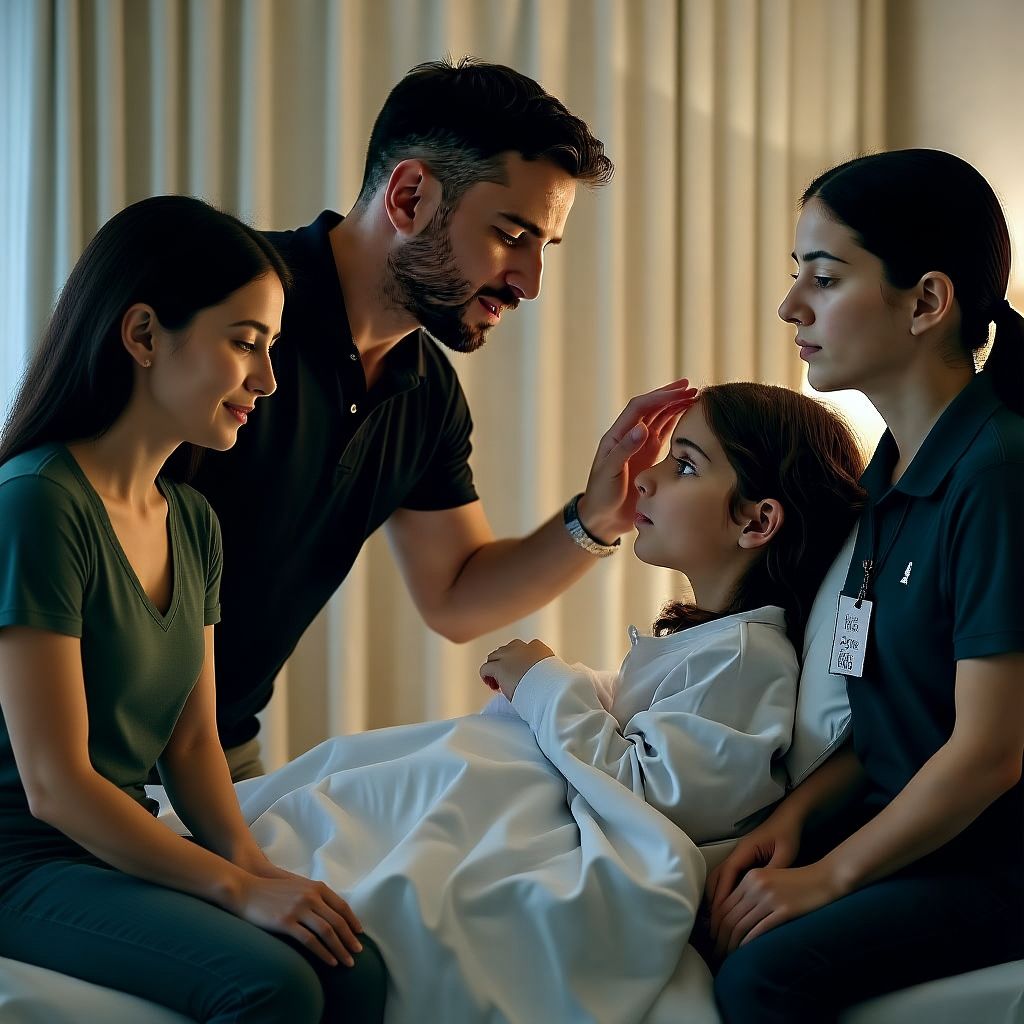}

\newcommand{\ExtTwoFLUXFA}{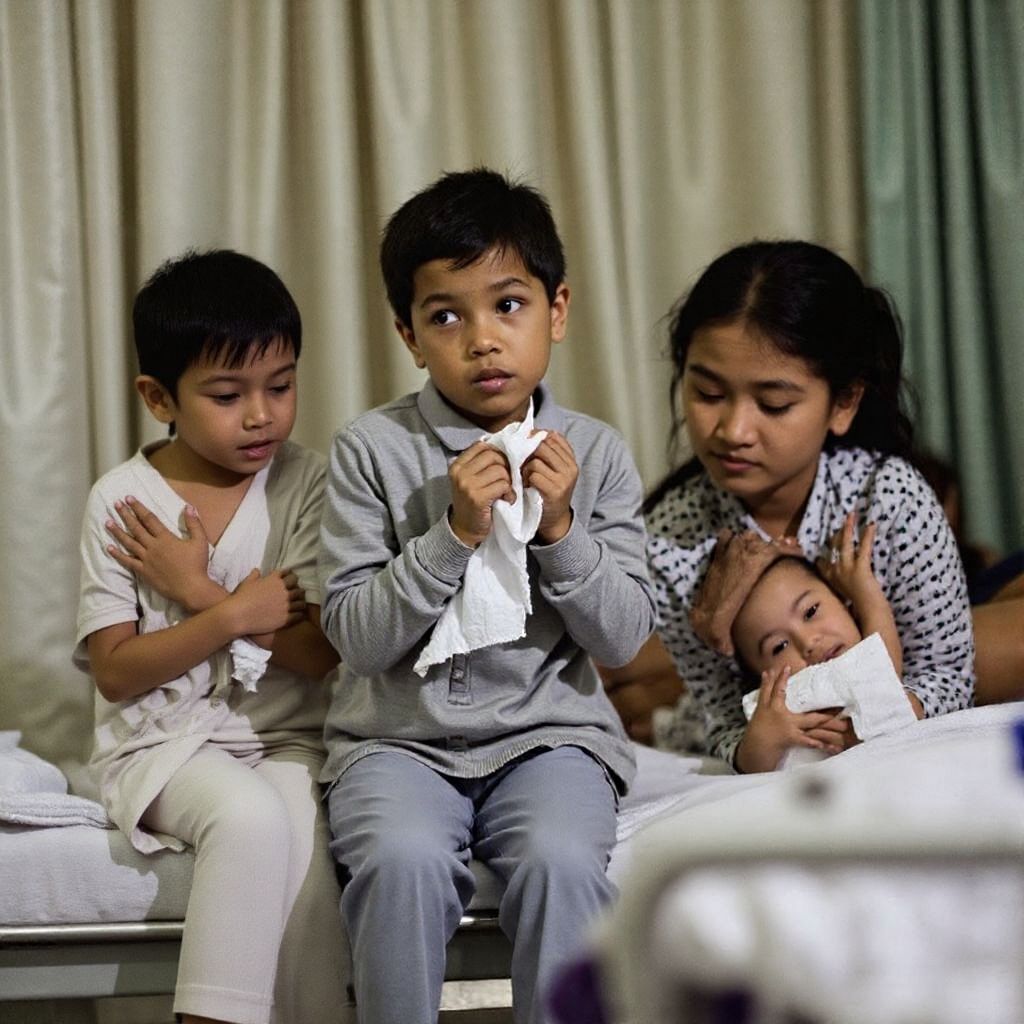}
\newcommand{\ExtTwoFLUXFB}{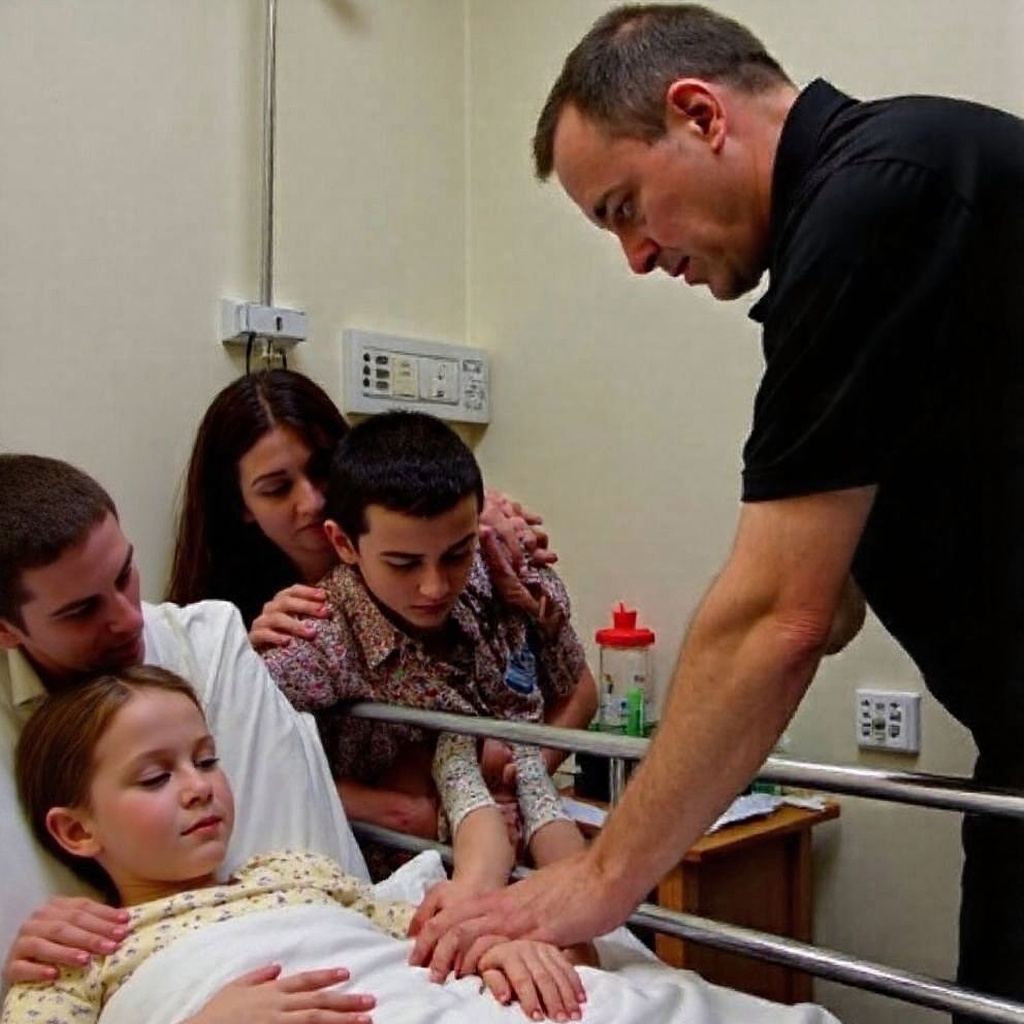}
\newcommand{\ExtTwoFLUXFC}{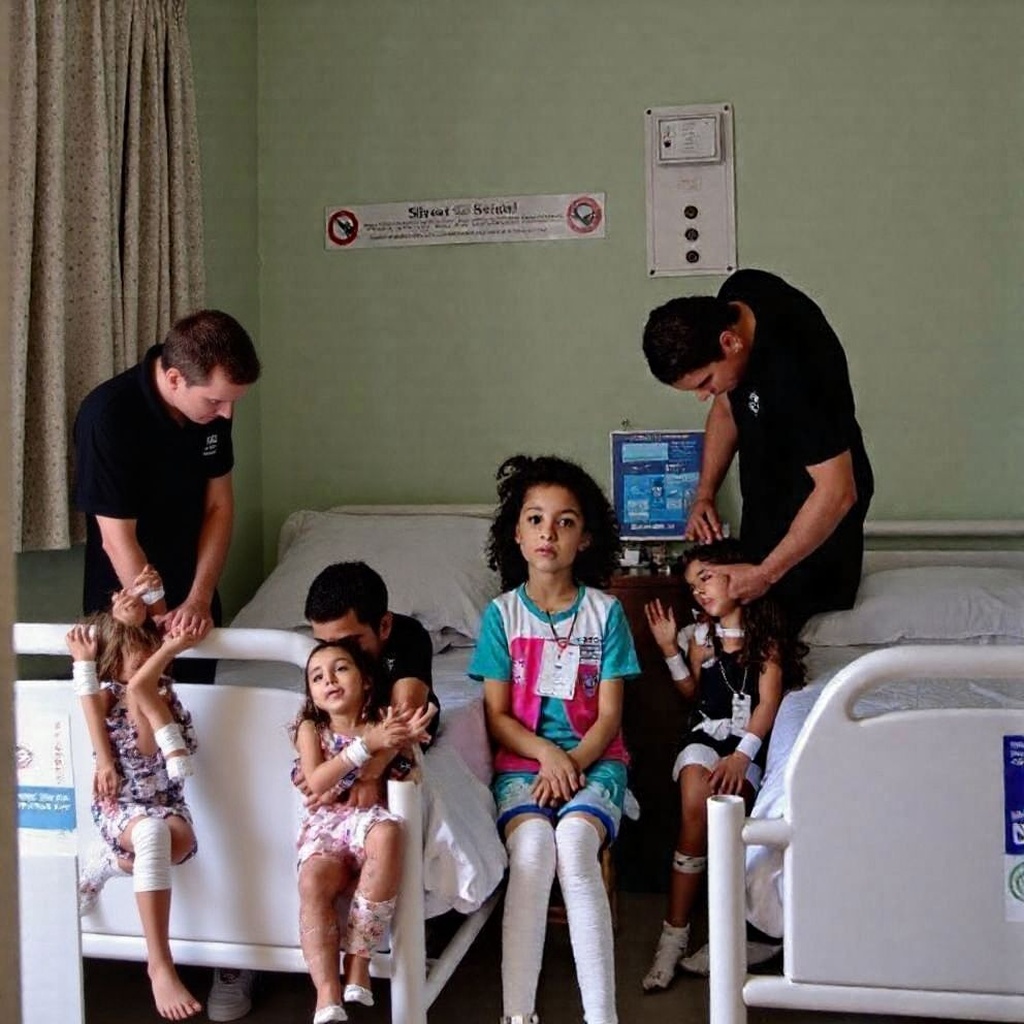}

\newcommand{\ExtTwoCreatiA}{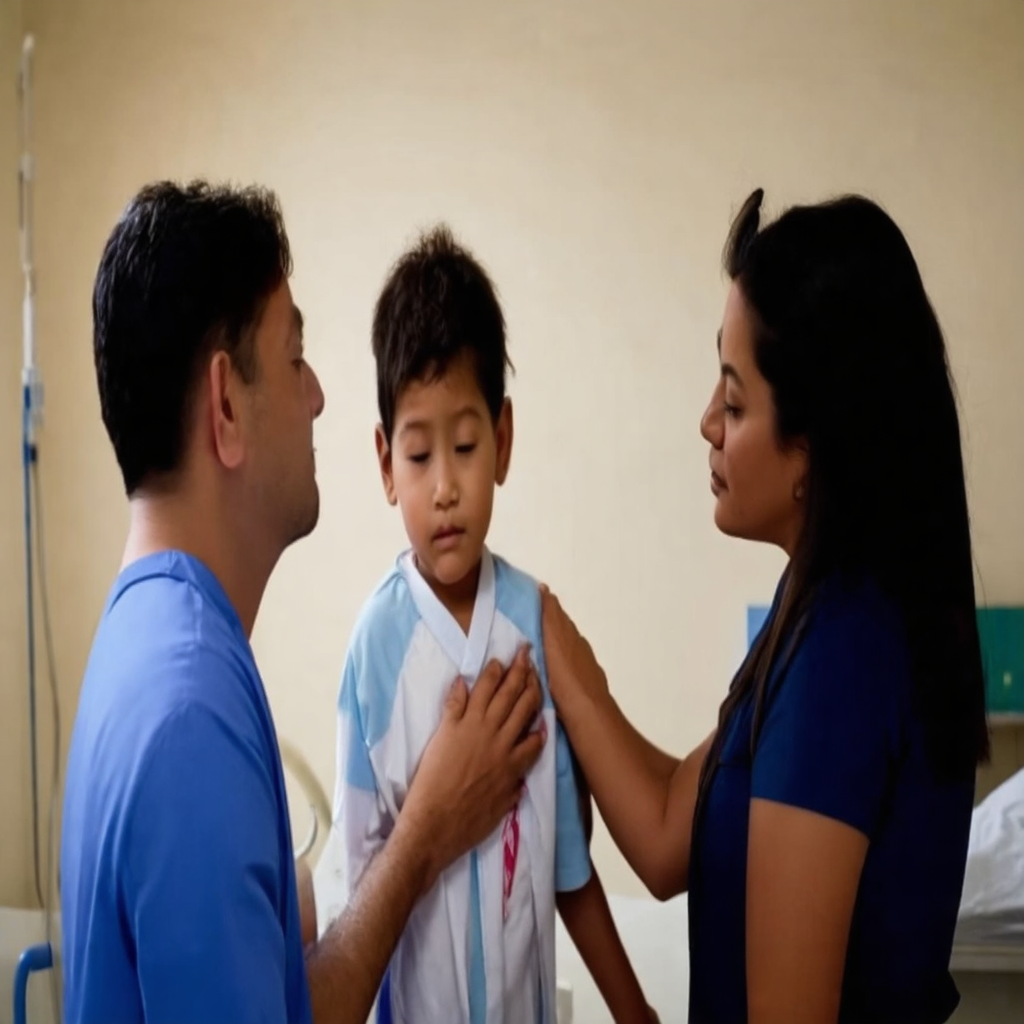}
\newcommand{\ExtTwoCreatiB}{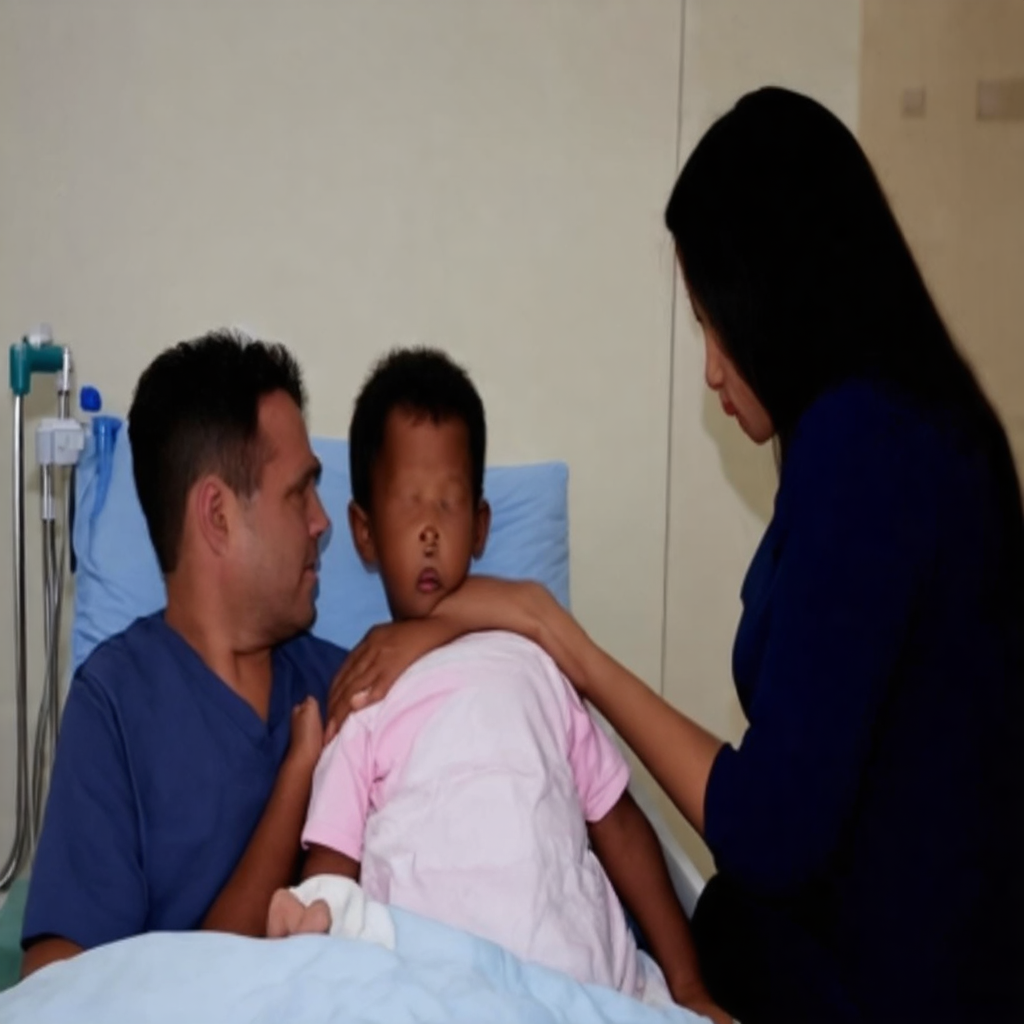}
\newcommand{\ExtTwoCreatiC}{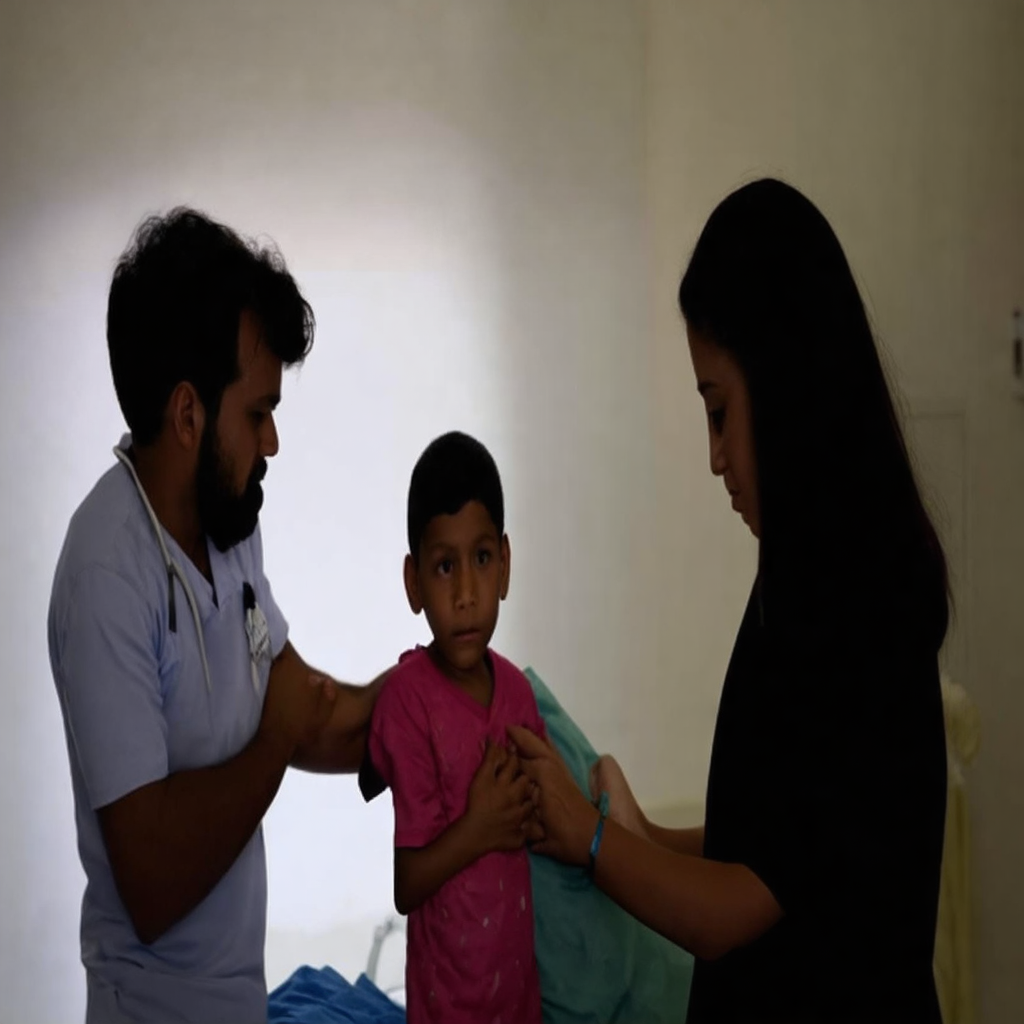}

\newcommand{\ExtTwoRealCPA}{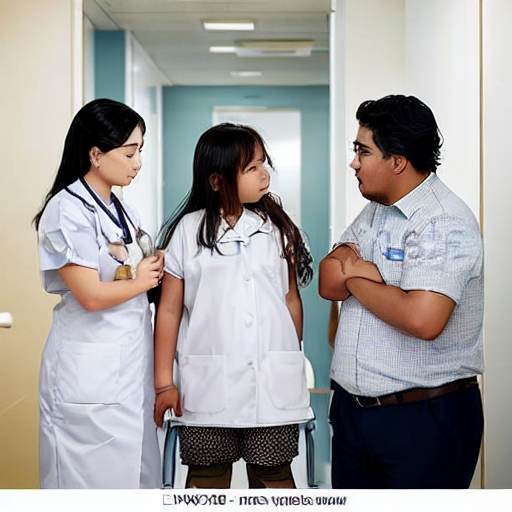}
\newcommand{\ExtTwoRealCPB}{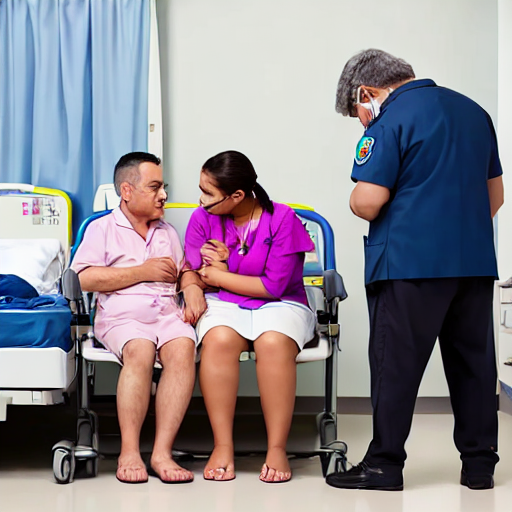}
\newcommand{\ExtTwoRealCPC}{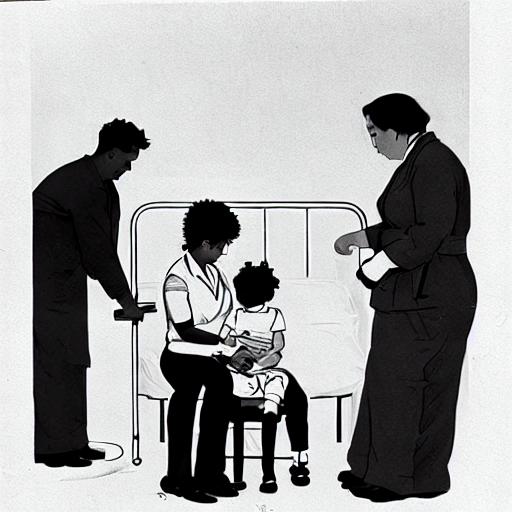}

\begin{figure*}[t]
\centering
\captionsetup{font=footnotesize,skip=2pt}
\resizebox{\textwidth}{!}{%
\begin{minipage}{\textwidth}
\centering

\begin{minipage}{0.12\linewidth}
  \textbf{Tier A:}
\end{minipage}%
\begin{minipage}{0.86\linewidth}
  \footnotesize
  Prompt: \emph{``\counttext{Two uniformed men escort two suited men} through an honor cordon as snow falls.''} | 
  Question (A): \emph{``Does this figure show \{Prompt\}?''}
\end{minipage}

\minipageimage{\ExtOursA}{\pass{} \quad \counttext{$\blacksquare$}}\hfill%
\minipageimage{\ExtBaseA}{\fail{} \quad \counttext{$\square$}}\hfill%
\minipageimage{\ExtSDLargeA}{\fail{} \quad \counttext{$\square$}}\hfill%
\minipageimage{\ExtSDXLA}{\fail{} \quad \counttext{$\square$}}\hfill%
\minipageimage{\ExtFLUXKA}{\fail{} \quad \counttext{$\square$}}\hfill%
\minipageimage{\ExtFLUXFA}{\fail{} \quad \counttext{$\square$}}\hfill%
\minipageimage{\ExtCreatiA}{\fail{} \quad \counttext{$\square$}}\hfill%
\minipageimage{\ExtRealCPA}{\fail{} \quad \counttext{$\square$}}%

\begin{minipage}{0.10\linewidth}
  \textbf{Tier B:}
\end{minipage}%
\begin{minipage}{0.86\linewidth}
  \footnotesize
  Prompt: \emph{``\counttext{Two uniformed service members escort two suited men}, \attrtext{each service member holding an umbrella for one of the men} as they walk on the wet stone steps.''} | 
  Question (B): \emph{``Does this figure show \{Prompt\}?''}
\end{minipage}

\minipageimage{\ExtOursB}{\pass{} \quad \counttext{$\blacksquare$} \attrtext{$\blacksquare$}}\hfill%
\minipageimage{\ExtBaseB}{\fail{} \quad \counttext{$\square$} \attrtext{$\square$}}\hfill%
\minipageimage{\ExtSDLargeB}{\fail{} \quad \counttext{$\square$} \attrtext{$\square$}}\hfill%
\minipageimage{\ExtSDXLB}{\fail{} \quad \counttext{$\square$} \attrtext{$\square$}}\hfill%
\minipageimage{\ExtFLUXKB}{\fail{} \quad \counttext{$\blacksquare$} \attrtext{$\square$}}\hfill%
\minipageimage{\ExtFLUXFB}{\fail{} \quad \counttext{$\square$} \attrtext{$\square$}}\hfill%
\minipageimage{\ExtCreatiB}{\fail{} \quad \counttext{$\blacksquare$} \attrtext{$\square$}}\hfill%
\minipageimage{\ExtRealCPB}{\fail{} \quad \counttext{$\blacksquare$} \attrtext{$\square$}}%

\begin{minipage}{0.10\linewidth}
  \textbf{Tier C:}
\end{minipage}%
\begin{minipage}{0.86\linewidth}
  \footnotesize
  Prompt: \emph{``In the center, two men in dark suits and ties walk side by side on wet stone steps, snow falling lightly around them. The man on the left walks with an easy stride, while the older man beside him moves in step, his expression calm and composed. To their left, a uniformed escort holds an umbrella angled over the younger man, walking slightly ahead to keep pace. Another escort follows behind on the right, holding an umbrella over the older man. The four move together in a measured, formal rhythm, their umbrellas dusted with snow as they ascend.''}\\ 
  \tiny
  Decomposed Questions (C): \emph{\counttext{a.} ``Does the image show two men in dark suits and ties walking side by side in the center, with one uniformed escort on each side accompanying them as they walk, in a snowy setting on stone steps?''} \emph{\attrtext{b.} ``Among the four people, is there a younger man in formal attire walking in the center-left, with a uniformed escort to the left holding an umbrella over him while walking slightly ahead?''} \emph{\spatialtext{c.} ``Among the four people, is there an older man walking in the center-right, with a uniformed escort following slightly behind on the right while holding an umbrella over him?''}
\end{minipage}

\minipageimage{\ExtOursC}{\pass{} \quad \counttext{$\blacksquare$} \attrtext{$\blacksquare$} \spatialtext{$\blacksquare$}}\hfill%
\minipageimage{\ExtBaseC}{\fail{} \quad \counttext{$\blacksquare$} \attrtext{$\square$} \spatialtext{$\square$}}\hfill%
\minipageimage{\ExtSDLargeC}{\fail{} \quad \counttext{$\square$} \attrtext{$\square$} \spatialtext{$\square$}}\hfill%
\minipageimage{\ExtSDXLC}{\fail{} \quad \counttext{$\square$} \attrtext{$\square$} \spatialtext{$\square$}}\hfill%
\minipageimage{\ExtFLUXKC}{\fail{} \quad \counttext{$\blacksquare$} \attrtext{$\square$} \spatialtext{$\blacksquare$}}\hfill%
\minipageimage{\ExtFLUXFC}{\fail{} \quad \counttext{$\square$} \attrtext{$\square$} \spatialtext{$\square$}}\hfill%
\minipageimage{\ExtCreatiC}{\fail{} \quad \counttext{$\blacksquare$} \attrtext{$\square$} \spatialtext{$\square$}}\hfill%
\minipageimage{\ExtRealCPC}{\fail{} \quad \counttext{$\blacksquare$} \attrtext{$\square$} \spatialtext{$\square$}}%

\begin{minipage}{0.12\linewidth}
  \textbf{Tier A:}
\end{minipage}%
\begin{minipage}{0.86\linewidth}
  \footnotesize
  Prompt: \emph{``In a hospital room, \counttext{a man gently comforts a young patient while a woman supports the child} beside her.''} | 
  Question (A): \emph{``Does this figure show \{Prompt\}?''}
\end{minipage}

\minipageimage{\ExtTwoOursA}{\pass{} \quad \counttext{$\blacksquare$}}\hfill%
\minipageimage{\ExtTwoBaseA}{\fail{} \quad \counttext{$\square$}}\hfill%
\minipageimage{\ExtTwoSDLargeA}{\fail{} \quad \counttext{$\square$}}\hfill%
\minipageimage{\ExtTwoSDXLA}{\fail{} \quad \counttext{$\square$}}\hfill%
\minipageimage{\ExtTwoFLUXKA}{\fail{} \quad \counttext{$\square$}}\hfill%
\minipageimage{\ExtTwoFLUXFA}{\fail{} \quad \counttext{$\square$}}\hfill%
\minipageimage{\ExtTwoCreatiA}{\fail{} \quad \counttext{$\square$}}\hfill%
\minipageimage{\ExtTwoRealCPA}{\fail{} \quad \counttext{$\square$}}%

\begin{minipage}{0.10\linewidth}
  \textbf{Tier B:}
\end{minipage}%
\begin{minipage}{0.86\linewidth}
  \footnotesize
  Prompt: \emph{``In a hospital ward, \counttext{a man in a black shirt leans over a bed and touches a young girl’s head} in a comforting gesture. \attrtext{The girl, with bandages on her legs, is held by a woman}.''} | 
  Question (B): \emph{``Does this figure show \{Prompt\}?''}
\end{minipage}

\minipageimage{\ExtTwoOursB}{\pass{} \quad \counttext{$\blacksquare$} \attrtext{$\blacksquare$}}\hfill%
\minipageimage{\ExtTwoBaseB}{\fail{} \quad \counttext{$\square$} \attrtext{$\square$}}\hfill%
\minipageimage{\ExtTwoSDLargeB}{\fail{} \quad \counttext{$\square$} \attrtext{$\square$}}\hfill%
\minipageimage{\ExtTwoSDXLB}{\fail{} \quad \counttext{$\square$} \attrtext{$\square$}}\hfill%
\minipageimage{\ExtTwoFLUXKB}{\fail{} \quad \counttext{$\blacksquare$} \attrtext{$\square$}}\hfill%
\minipageimage{\ExtTwoFLUXFB}{\fail{} \quad \counttext{$\square$} \attrtext{$\square$}}\hfill%
\minipageimage{\ExtTwoCreatiB}{\fail{} \quad \counttext{$\blacksquare$} \attrtext{$\square$}}\hfill%
\minipageimage{\ExtTwoRealCPB}{\fail{} \quad \counttext{$\blacksquare$} \attrtext{$\square$}}%

\begin{minipage}{0.10\linewidth}
  \textbf{Tier C:}
\end{minipage}%
\begin{minipage}{0.86\linewidth}
  \footnotesize
  Prompt: \emph{``Inside a small hospital ward with two beds, a man in a black polo stands at the bedside and reaches out to gently touch a young girl’s forehead in comfort. The girl sits on a woman’s lap at the edge of the bed, with bandages on both knees, wearing socks and an ID tag. The woman sits close behind her with an arm around the girl’s waist, steadying and consoling her. On the adjacent bed, another female patient with a wrapped leg sits upright and watches. The scene centers on the man’s comforting gesture toward the child, supported by the woman at her side.''}\\
  \tiny
   Decomposed Questions (C): \emph{\counttext{a.} ``Does the image show a man standing beside a hospital bed reaching out to gently touch a young girl’s forehead in a comforting manner?''} \emph{\attrtext{b.} ``Is there a young girl sitting on a woman’s lap at the edge of a hospital bed, wearing socks and an ID tag, with bandages on both knees, while the woman has an arm around the girl’s waist to steady and console her?''} \emph{\spatialtext{c.} ``Is there a female patient sitting upright on one of the hospital beds with a wrapped leg, positioned on the bed next to where a man and a woman are attending to a young girl?''}
\end{minipage}

\minipageimage{\ExtTwoOursC}{\pass{} \quad \counttext{$\blacksquare$} \attrtext{$\blacksquare$} \spatialtext{$\blacksquare$}}\hfill%
\minipageimage{\ExtTwoBaseC}{\fail{} \quad \counttext{$\blacksquare$} \attrtext{$\square$} \spatialtext{$\square$}}\hfill%
\minipageimage{\ExtTwoSDLargeC}{\fail{} \quad \counttext{$\square$} \attrtext{$\square$} \spatialtext{$\blacksquare$}}\hfill%
\minipageimage{\ExtTwoSDXLC}{\fail{} \quad \counttext{$\square$} \attrtext{$\square$} \spatialtext{$\square$}}\hfill%
\minipageimage{\ExtTwoFLUXKC}{\fail{} \quad \counttext{$\blacksquare$} \attrtext{$\square$} \spatialtext{$\square$}}\hfill%
\minipageimage{\ExtTwoFLUXFC}{\fail{} \quad \counttext{$\square$} \attrtext{$\square$} \spatialtext{$\square$}}\hfill%
\minipageimage{\ExtTwoCreatiC}{\fail{} \quad \counttext{$\square$} \attrtext{$\square$} \spatialtext{$\square$}}\hfill%
\minipageimage{\ExtTwoRealCPC}{\fail{} \quad \counttext{$\square$} \attrtext{$\blacksquare$} \spatialtext{$\square$}}%

\input{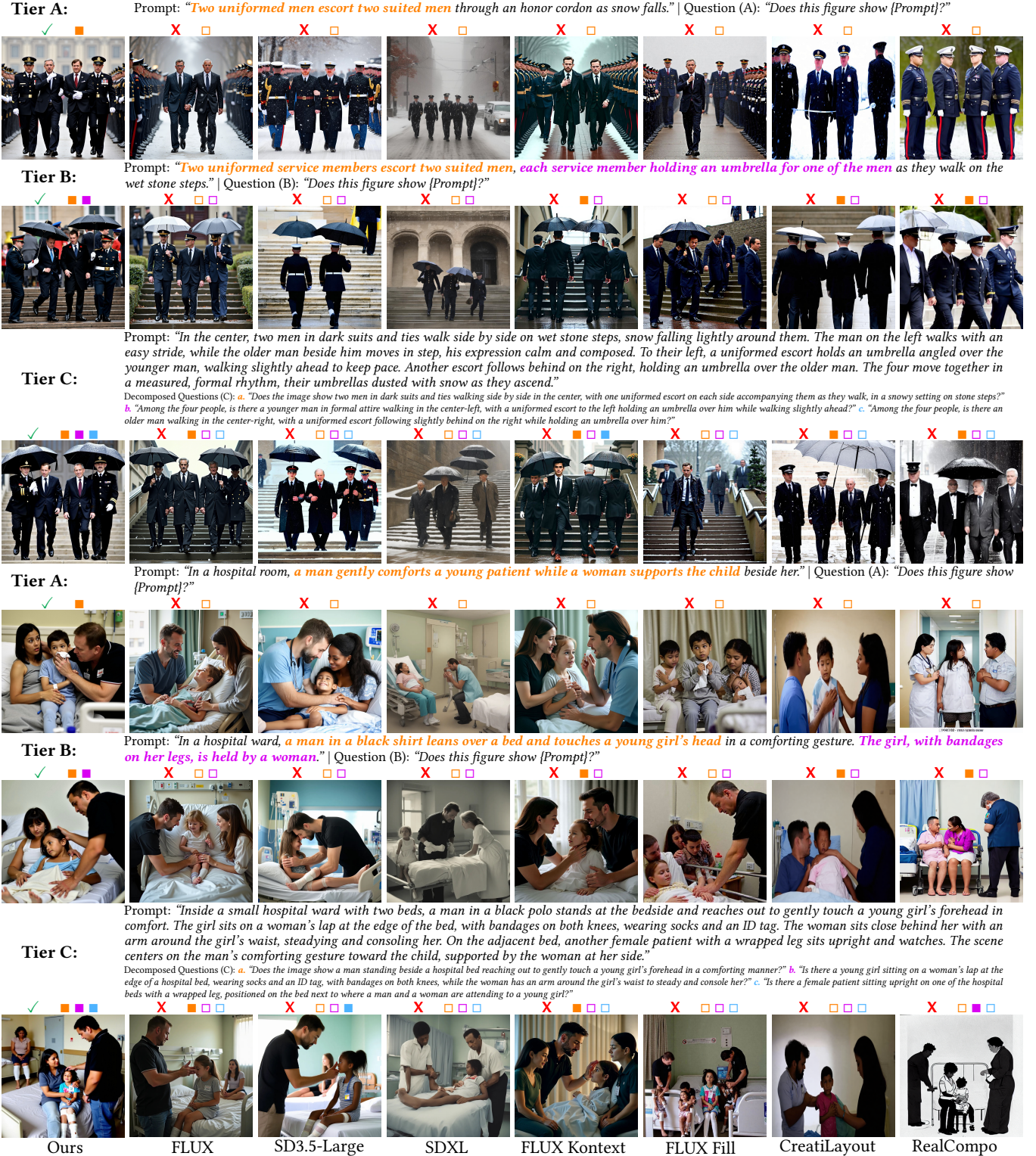}

\end{minipage}
}%

\caption{\textbf{Qualitative comparison.}
Additional qualitative results on prompts sampled from \evaldataset{}, covering Tier A/B/C, together with per-sample VQA Accuracy evaluation.
For each tier, an image is considered \textit{correct} (\pass{}) if a MLLM (GPT-5.2) returns positive answers to \emph{all} evaluation questions associated with that tier.
We compare our method against baselines from three model families:
T2I models (FLUX~[dev]~\cite{flux2024}, SD3.5-Large~\cite{esser2024scaling}, SDXL~\cite{podell2023sdxl}), editing / inpainting models (FLUX Kontext~[dev]~\cite{labs2025flux1kontextflowmatching}, FLUX Fill~[dev]~\cite{flux2024}),
and layout-controlled models (CreatiLayout~\cite{zhang2025creatilayout}, RealCompo~\cite{zhang2024realcompo}).}
\label{fig:comparison}
\Description[Qualitative comparison across prompt tiers]{Full-page qualitative comparison for two prompts across prompt tiers A, B, and C and eight methods. Correctness markers indicate VQA alignment for each generated image.}
\end{figure*}

\newcommand{\GalleryScale}{1.00}         
\newcommand{\GalleryImgH}{2.85cm}        
\newcommand{\GalleryMinGap}{2.2mm}       
\newcommand{\GalleryCaptionSize}{\scriptsize}
\newcommand{\GalleryLabelSize}{\scriptsize}
\newcommand{\GalleryLabelGap}{0.0mm}

\newcommand{\LabOrig}{Reference Image}
\newcommand{\LabFlux}{FLUX}
\newcommand{\LabProc}{Generation Process}
\newcommand{\LabOurs}{Ours}

\newcommand{\mempty}{}  

\newsavebox{\GalleryTmpImg}
\newcommand{\GalleryPanelBox}[3]{%
  \sbox{\GalleryTmpImg}{\includegraphics[height=\GalleryImgH]{#1}}%
  \vtop{%
    \hbox to \wd\GalleryTmpImg{%
      \hfil {\GalleryLabelSize #3}\hfil
    }%
    \kern 0.3mm  
    \hbox{\usebox{\GalleryTmpImg}}%
    \kern\GalleryLabelGap
    \hbox to \wd\GalleryTmpImg{%
      \hfil {\GalleryLabelSize\textbf{#2}}\hfil
    }%
  }%
}
\newcommand{\GalleryStripRow}[7]{%
  \par\noindent
  {\GalleryCaptionSize\raggedright #1\par}
  \noindent
  \resizebox{\linewidth}{!}{%
    \hbox{%
      \GalleryPanelBox{#2}{\LabOrig}{\mempty}\hspace{\GalleryMinGap}%
      \GalleryPanelBox{#3}{\LabFlux}{#6}\hspace{\GalleryMinGap}%
      \GalleryPanelBox{#4}{\LabProc}{\mempty}\hspace{\GalleryMinGap}%
      \GalleryPanelBox{#5}{\LabOurs}{#7}%
    }%
  }\par
}

\begin{figure*}[t]
  \centering
  \captionsetup{font=footnotesize,skip=2pt}

  \resizebox{0.93\textwidth}{!}{%
    \scalebox{\GalleryScale}{%
      \begin{minipage}{\textwidth}

      \GalleryStripRow
        {\emph{\counttext{A man} in a brown shirt \counttext{bends from the waist} beside the chair, \counttext{head lowered and facing right toward the elderly man}, visible from the waist up, paying respects as \attrtext{he grips the elderly man’s hand with both hands and brings it close to his forehead}. \spatialtext{The elderly man}, wearing a white shirt and glasses, sits facing him and warmly acknowledges the gesture, \spatialtext{enclosing the younger man’s hands with both of his}.}}
        {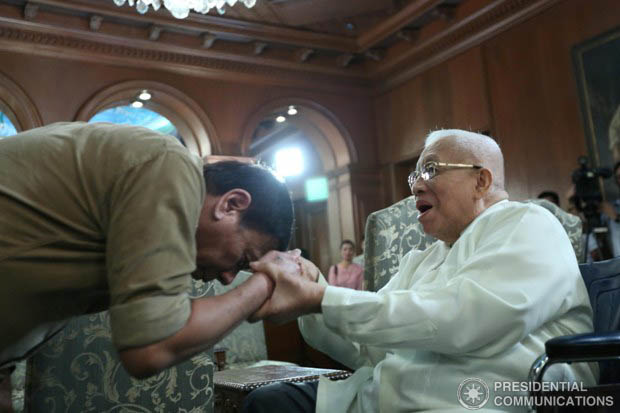}
        {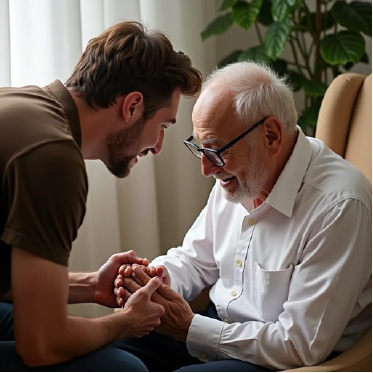}
        {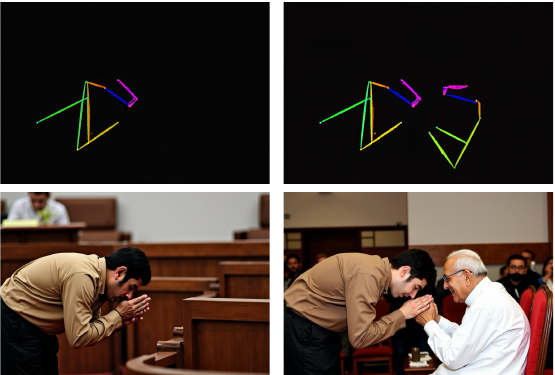}
        {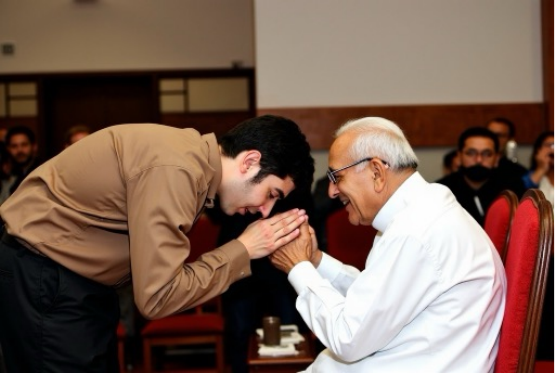}
        {{\fail{} \quad \counttext{$\blacksquare$} \attrtext{$\square$} \spatialtext{$\blacksquare$}}} 
      {{\pass{} \quad \counttext{$\blacksquare$} \attrtext{$\blacksquare$} \spatialtext{$\blacksquare$}}} 

      \GalleryStripRow
        {\emph{At the front of a chamber, a formal swearing-in ceremony is taking place. \counttext{On the left, a man} in a dark suit \counttext{raises his right hand to take the oath, while placing his left hand on a large book held by a woman} standing beside him in the center. \attrtext{The woman}, dressed in a black suit with white trim, looks on attentively, \attrtext{supporting the book with both hands}. \spatialtext{Facing them on the right, another man} in a gray suit \spatialtext{raises his right hand and reads from a sheet of paper as he administers the oath}. The three stand close together, focused and composed, with a microphone angled toward them as the ceremony unfolds.}}
        {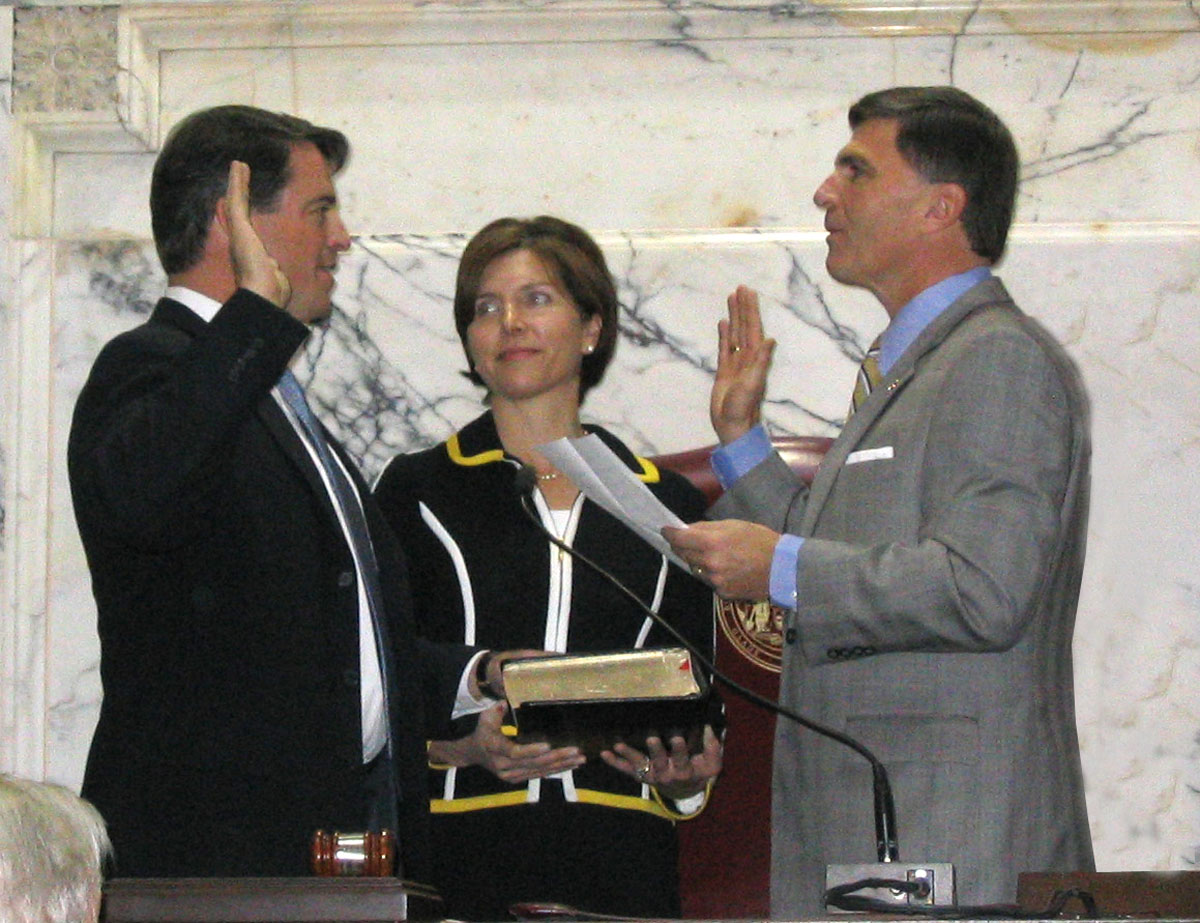}
        {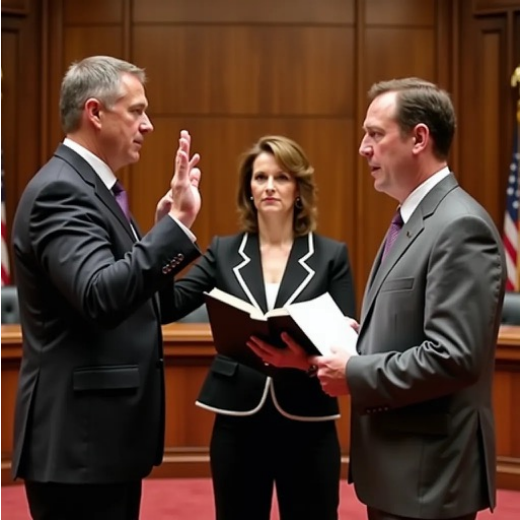}
        {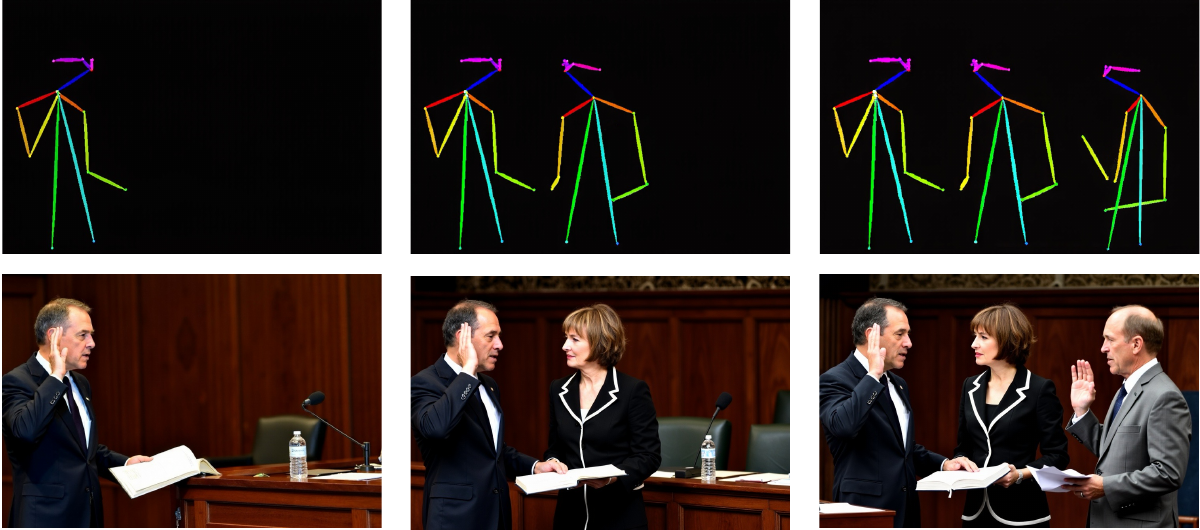}
        {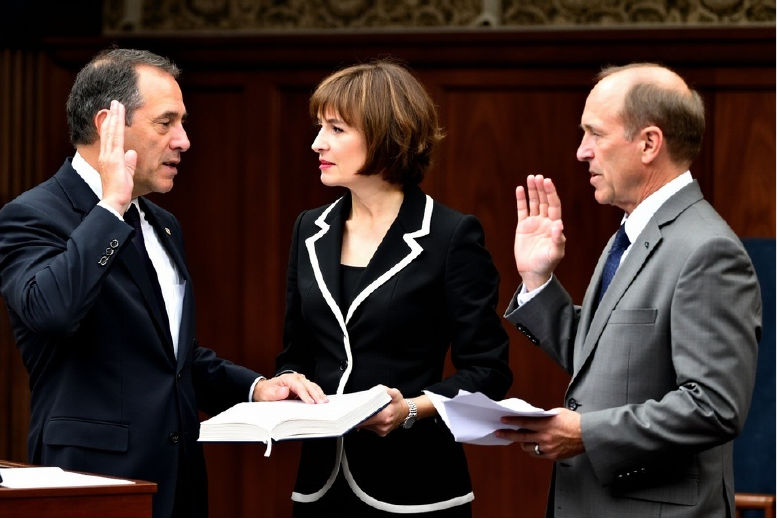}
        {{\fail{} \quad \counttext{$\square$} \attrtext{$\square$} \spatialtext{$\square$}}} 
      {{\pass{} \quad \counttext{$\blacksquare$} \attrtext{$\blacksquare$} \spatialtext{$\blacksquare$}}} 

      \GalleryStripRow
        {\emph{Under stadium lights, a coach in a blue shirt and khaki pants, headset on, steps beside a player in a gold helmet and dark jersey number 18. \counttext{The coach brings a hand up near the player’s shoulder and talks directly into his ear}, actively coaching him as \attrtext{the player listens and gestures toward the field}. \spatialtext{Another headset-wearing coach and a second uniformed teammate stand just behind them, watching the exchange}.}}
        {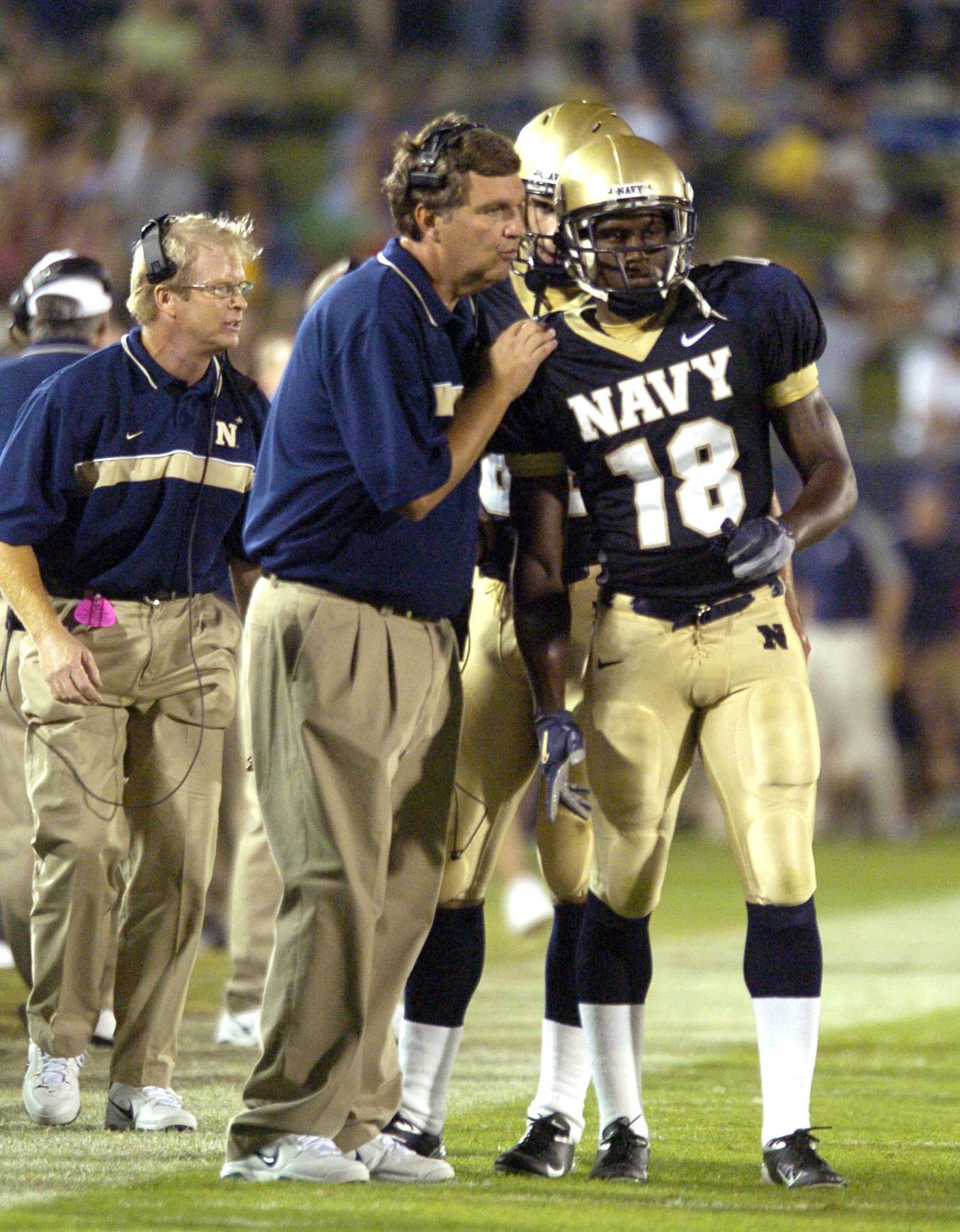}
        {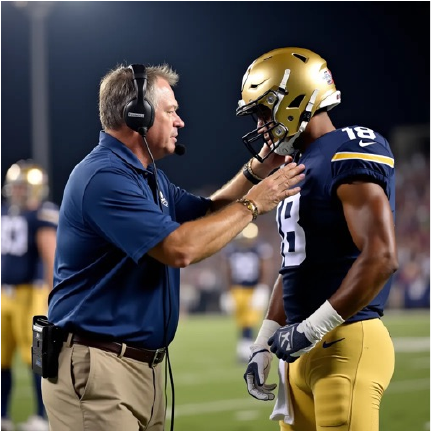}
        {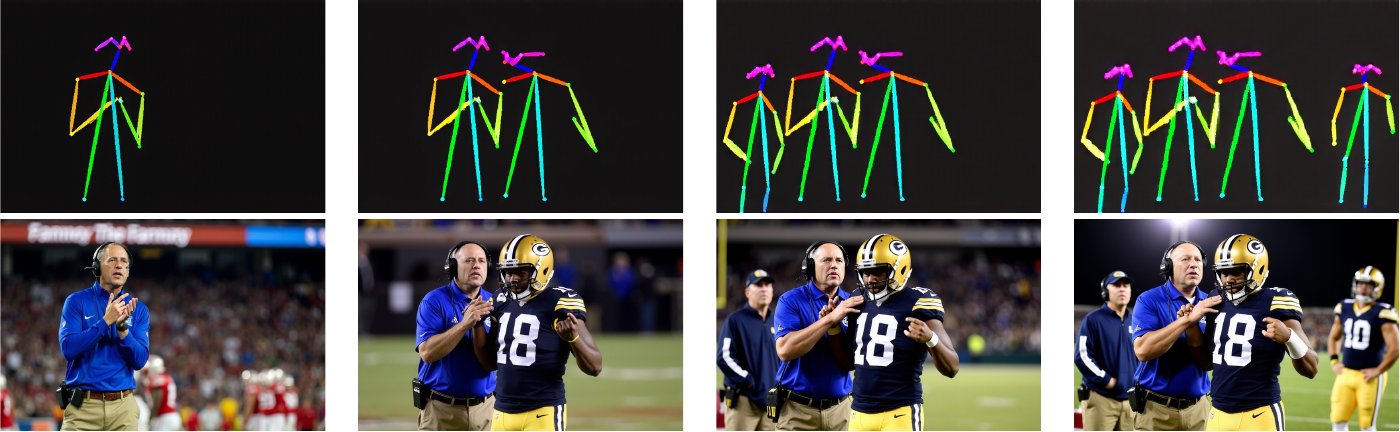}
        {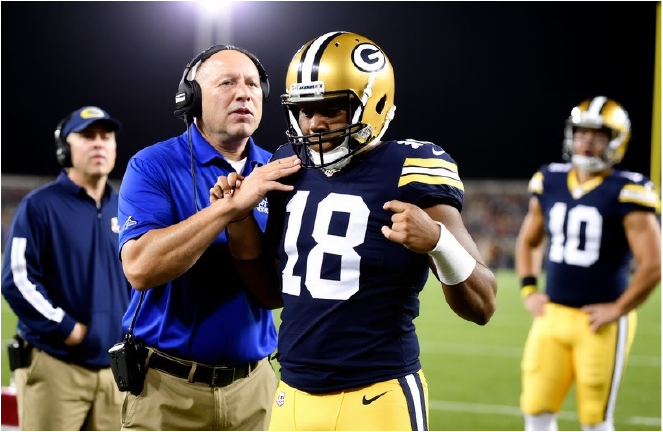}
        {{\fail{} \quad \counttext{$\square$} \attrtext{$\square$} \spatialtext{$\square$}}} 
      {{\pass{} \quad \counttext{$\blacksquare$} \attrtext{$\blacksquare$} \spatialtext{$\blacksquare$}}} 

      \GalleryStripRow
        {\emph{During a formal ceremony, \counttext{a Marine} in a dress uniform leans forward from the left, \counttext{holding a wrapped bouquet tied in both hands as he extends it to a woman} in red seated near the aisle. \attrtext{She smiles and reaches out with both hands to receive the flowers}, \spatialtext{while the man} in a blue dress uniform seated to her right \spatialtext{claps and watches}, \spatialtext{and a woman} in a blue dress farther to the right also \spatialtext{claps as she looks on}.}}
        {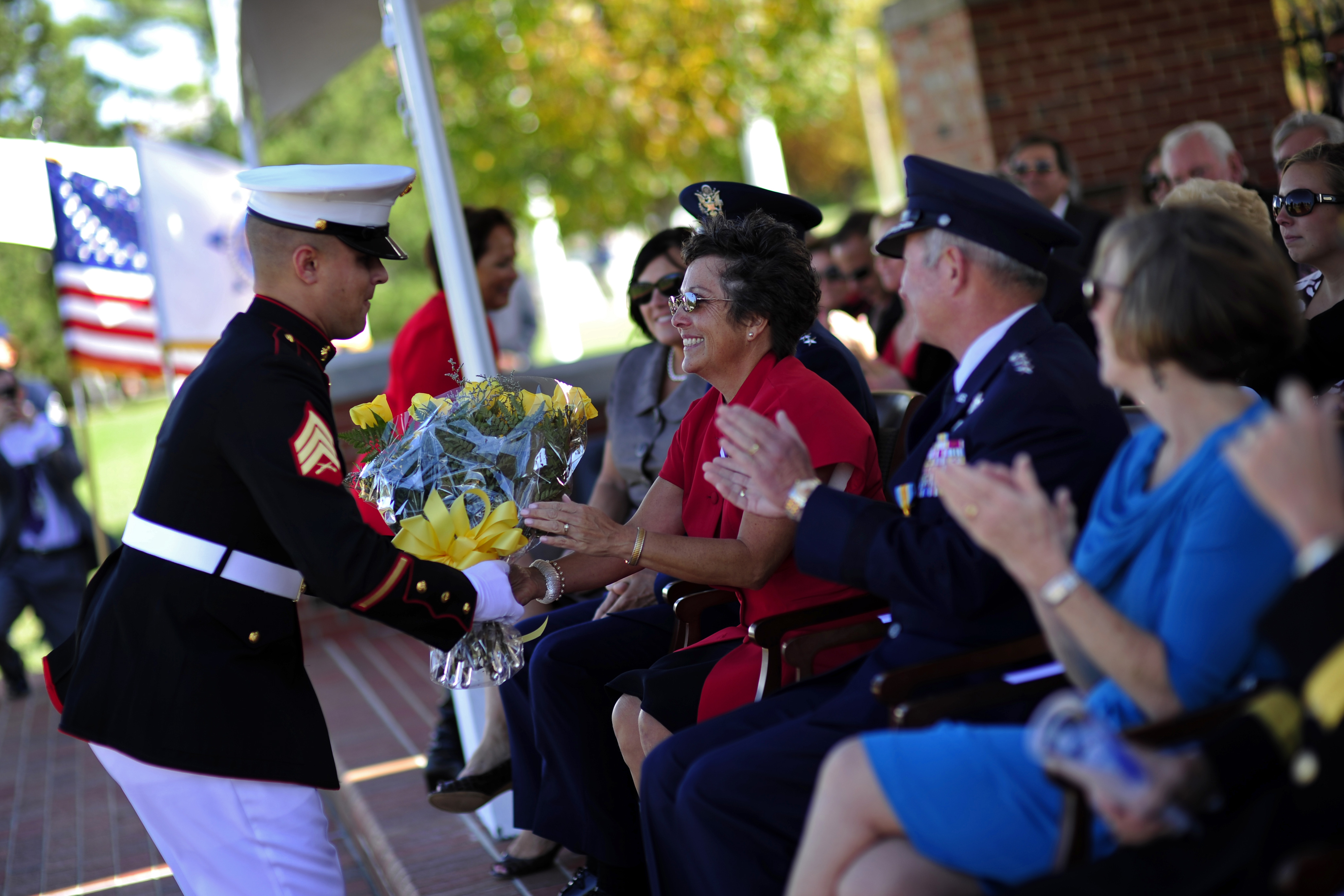}
        {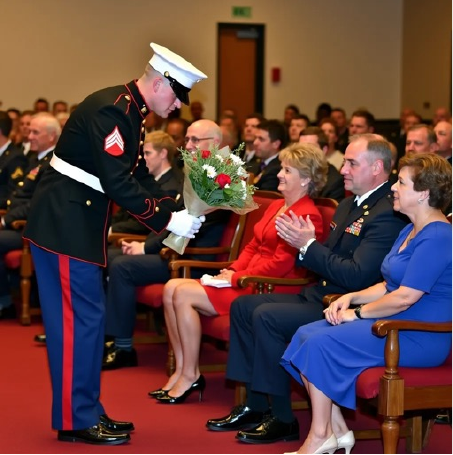}
        {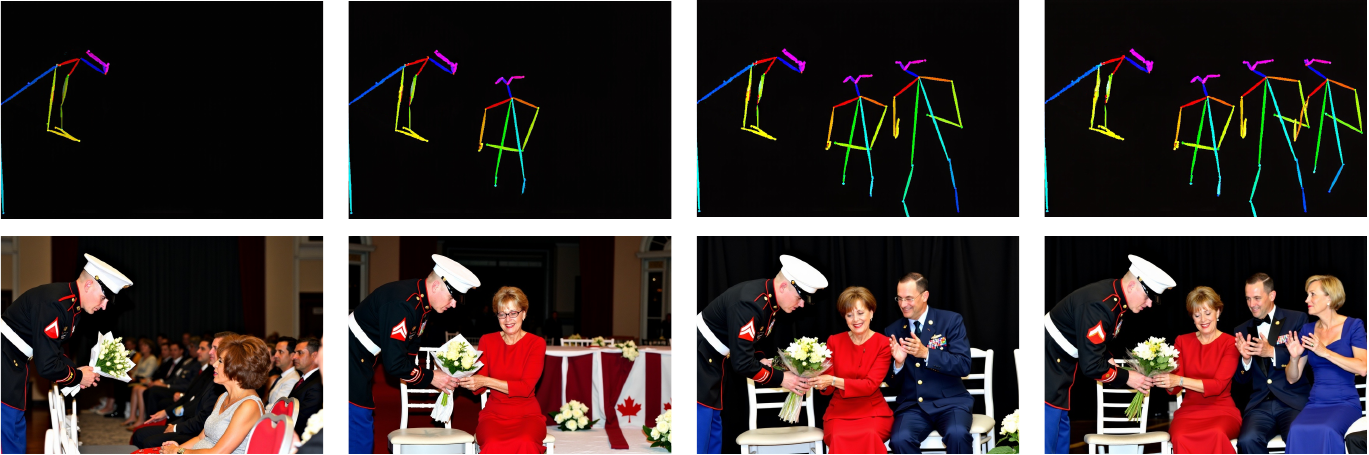}
        {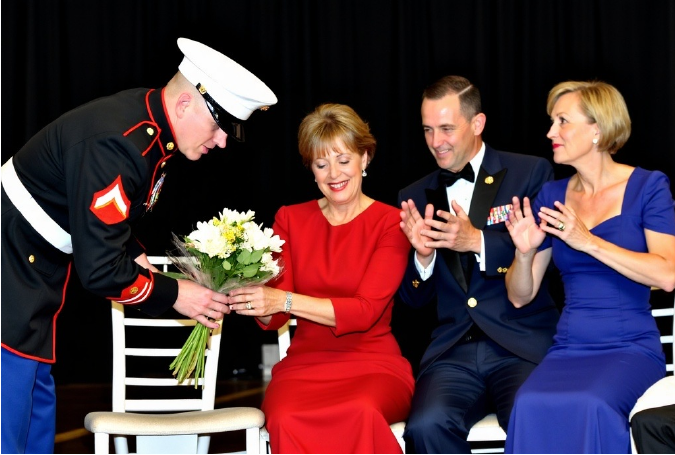}
        {{\fail{} \quad \counttext{$\square$} \attrtext{$\square$} \spatialtext{$\blacksquare$}}} 
      {{\pass{} \quad \counttext{$\blacksquare$} \attrtext{$\blacksquare$} \spatialtext{$\blacksquare$}}} 

      \GalleryStripRow
        {\emph{Behind a draped table, \counttext{a man in a dark suit and four men in military dress uniforms stand shoulder to shoulder, smiling as they lean forward together to cut a large rectangular cake}. \attrtext{The suited man stands at the center, gripping the knife firmly with both hands}, while \spatialtext{the uniformed men on either side lean in from both directions, each placing a hand over or beside his on the handle} so that their hands overlap in a shared gesture. Forming a tight semicircle around the cake, they bend slightly toward it, the moment capturing a blend of ceremony, unity, and camaraderie.}}
        {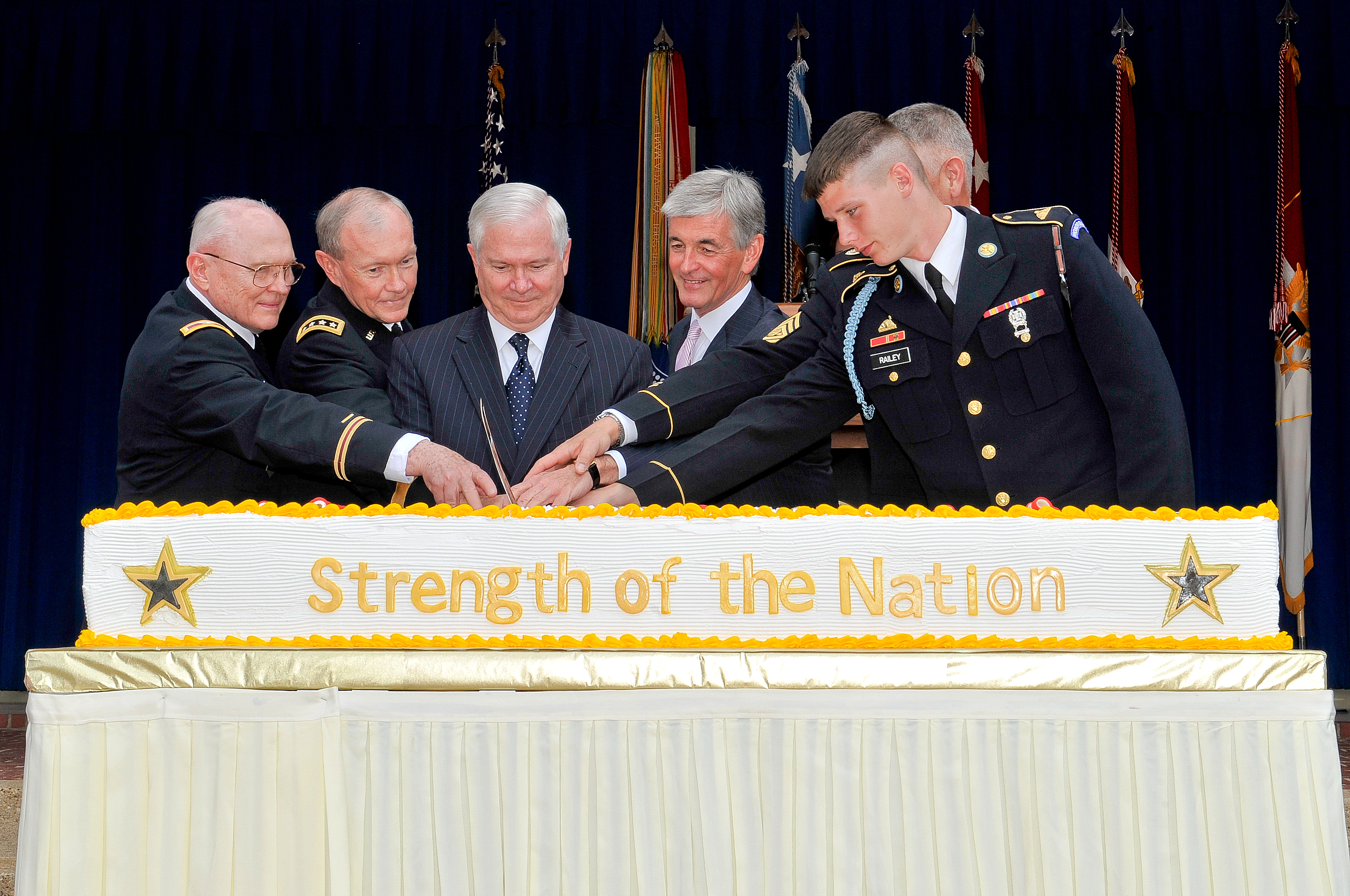}
        {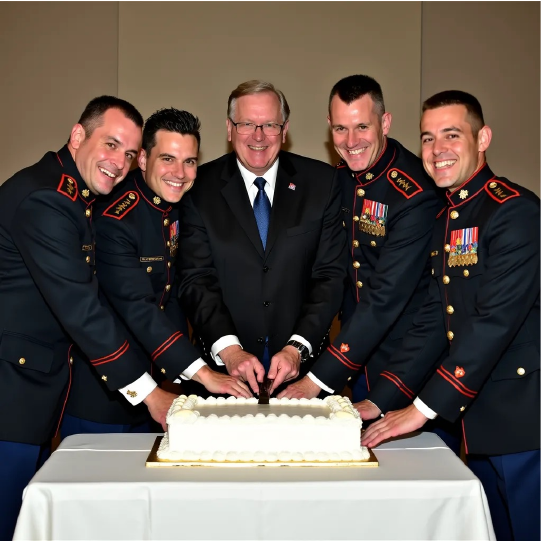}
        {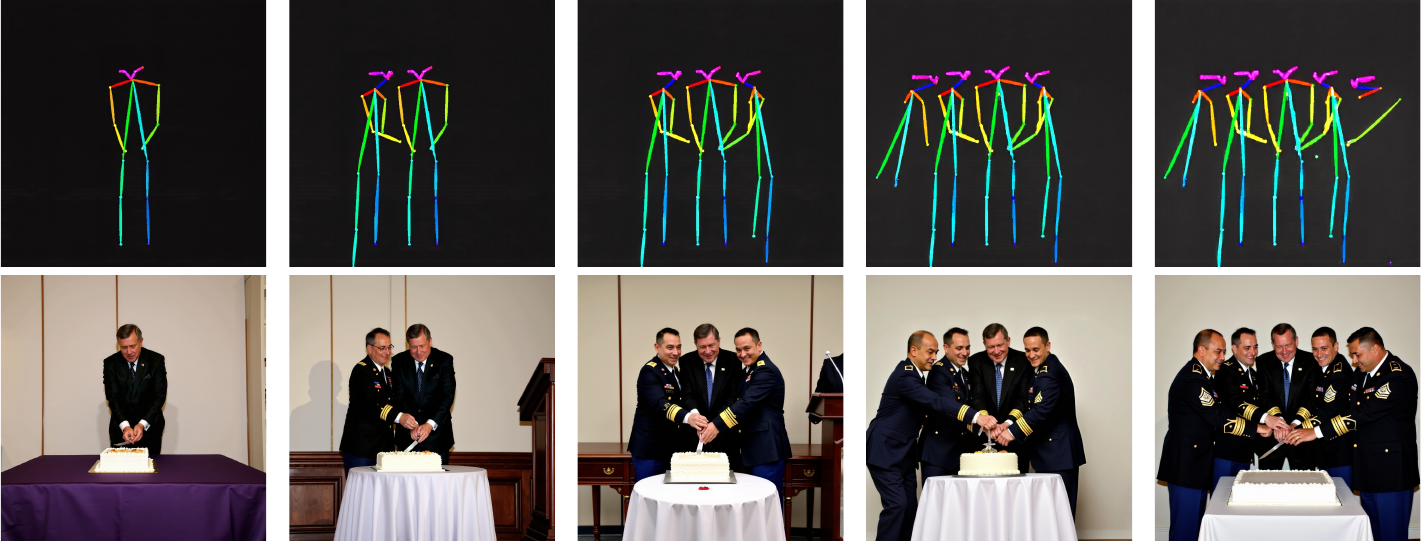}
        {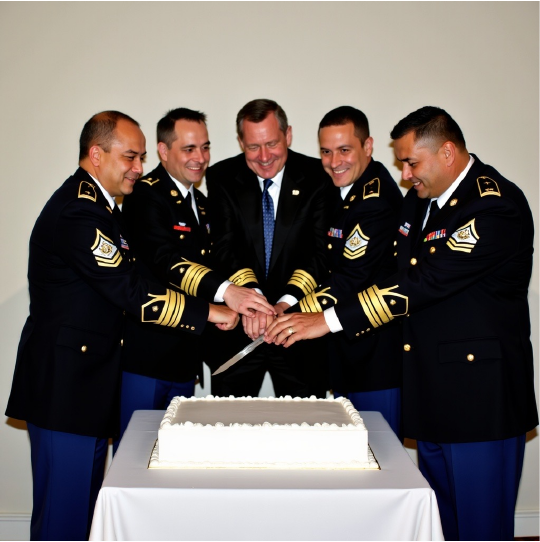}
        {{\fail{} \quad \counttext{$\blacksquare$} \attrtext{$\blacksquare$} \spatialtext{$\square$}}} 
      {{\pass{} \quad \counttext{$\blacksquare$} \attrtext{$\blacksquare$} \spatialtext{$\blacksquare$}}} 

      \end{minipage}%
    }
  }%

  \caption{
    \textbf{Gallery of Tier~C multi-person interaction examples.}
    Each row corresponds to a Tier~C test example sampled from our \evaldataset{}.
    For each example, the caption shown above is a fine-grained Tier~C prompt constructed by describing the \emph{reference image}, specifying all important people and their interactions in detail. We then compare the images generated by the \emph{FLUX} baseline and \emph{our method} using this prompt.
    We additionally visualize the \emph{iterative} pose--image generation process of our approach. Overall, our method more faithfully realizes the interactions specified by the prompt, while the baseline often fails to capture the full interaction content.
    Reference-image credits, top to bottom: KING RODRIGUEZ/PPD, public domain; Doug Gansler, CC BY 2.0, unmodified; U.S. Navy/Damon J. Moritz, public domain; U.S. Air Force/Jacob N. Bailey, public domain; Eboni L. Myart/U.S. Army, public domain.
    }

  \label{fig:gallery}
  \Description[Gallery of Tier C interaction examples]{Full-page gallery of five Tier C examples comparing reference images, FLUX results, the iterative generation process, and final outputs from our method.}
\end{figure*}

\clearpage
\section*{Supplementary Material}
Interactive visualizations, prompt examples, instruction prompts, and additional qualitative results are available on the project page.
The viewer presents 50 randomly sampled generation results from our \evaldataset{}
benchmark. Each panel shows the original reference image and interaction label from the
Waldo and Wenda dataset, along with three corresponding prompt tiers (A/B/C). For each
tier, users can interactively click to reveal generation results, toggle model columns
(ours, FLUX~[dev], FLUX Kontext~[dev]) for side-by-side comparison, and view the
evaluation questions used for that tier. Clicking on any image reveals the VQA model's
answer and score for the corresponding prompt--image pair.

\section{Technical Details}
\label{supp_tech}

\subsection{Prompt Parsing and Coarse Layout Extraction}
\label{supp:stage1_prompt}
As described in the main paper, we use an LLM to convert a user-provided
interaction prompt into a \emph{structured interaction specification}
consisting of a global interaction description $c_g$ and an ordered set of
per-person descriptions with associated bounding boxes $\{(c_i, b_i)\}$.

In our implementation, we apply GPT-5.2 to generate this structured
representation. In addition to the per-person bounding boxes, the LLM is also
asked to specify the target image resolution, with the longer edge fixed to
1024 pixels. This allows our generation pipeline to support variable aspect
ratios while remaining compatible with the multi-resolution training setup
used by FLUX.

The LLM is encouraged to arrange the per-person descriptions in a \emph{semantically coherent order}.
The ordering typically begins with the primary actor or interaction initiator, then follows individuals who are spatially or functionally close to the central action, and finally proceeds to background or peripheral figures.
This ordering naturally aligns with the bounding box layout and supports our \revadded{iterative} generation pipeline. 



\subsection{Model Implementation}
\smallskip \noindent \textbf{Implementation Setup.}
We implement our model on top of the official FLUX.1-dev codebase released by Black Forest Labs~\cite{flux2024}, which is a DiT-based latent diffusion model supporting multi-resolution synthesis.

\smallskip \noindent \textbf{Trainable Parameters.}
All image and text stream parameters are frozen throughout training.
The only trainable components are:
\begin{itemize}
    \item Pose-side input/output projection layers (copied from the original image projection layers and left trainable)
    \item LoRA modules (rank 512) inserted into:
    \begin{itemize}
        \item Query, key, value, and output projections in the self-attention modules of the pose stream
        \item Feed-forward layers in the pose stream
        \item Pose-side modulation layers for scale, shift, and gating
    \end{itemize}
\end{itemize}

\smallskip \noindent \textbf{Training Setup.}
To support multi-resolution generation and preserve natural human proportions, we adopt a multi-resolution training setup.
Each training image and its corresponding pose input is resized so that the longer edge is fixed to 1024 pixels while preserving the original aspect ratio.

We optimize using AdamW with a learning rate of \(3 \times 10^{-4}\) and weight decay of 0.01.
Training is conducted with a batch size of 32 using gradient accumulation and bf16 mixed-precision.
\revadded{We use Fully Sharded Data Parallel (FSDP) across 8$\times$A100 GPUs (80\,GB each), reaching our 5k-step checkpoint in approximately one day.}

\smallskip \noindent \textbf{Training Procedure.}
We supervise the pose and image streams using the flow-matching~\cite{chen2024diffusion} objective.
In flow matching, the model learns to predict the velocity that maps a noisy input toward a clean target under linear interpolation.
Given a clean target tensor \( \mathbf{x}_1 \), a noise sample \( \mathbf{x}_0 \sim \mathcal{N}(0, I) \), and a timestep \( t \sim \mathcal{U}(0,1) \), we construct the interpolated input:
\[
\mathbf{x}_t = (1 - t)\,\mathbf{x}_1 + t\,\mathbf{x}_0,
\]
and train the model to predict the velocity $u = x_0 - x_1$ under an \( \ell_2 \) loss.

In our setup, each training sample corresponds to a specific \revadded{iterative} scene stage \( i \).
Let \( \mathbf{P}_{i} \) denote the rendered pose image for the current scene state containing \( i \) people, and let \( \mathbf{I} \) be the final RGB image for the full scene with \( n \) people.

At training stage \( i \), the inputs to the dual pose–image stream are \( \{\mathbf{P}_{i},\; \mathbf{0}\} \) for intermediate stages (\( i < N \)) and \( \{\mathbf{P}_{i},\; \mathbf{I} \} \) for the final stage (\( i = N \)).

The previous pose image \( \mathbf{P}_{i-1} \), representing the scene state up to person \( i{-}1 \), is encoded using the same VAE and pose-side projection layers as the target pose \( \mathbf{P}_{i} \), but remains unnoised and serves purely as structural context. 
Meanwhile, \( \mathbf{P}_{i} \) is perturbed with Gaussian noise and used as the prediction target under the flow-matching objective.

Following the diffusion-forcing strategy~\cite{chen2024diffusion}, we apply independent noise levels to the pose and image targets to allow the model to learn their interdependence.
As a result, in intermediate stages, the image stream receives pure-noise inputs after perturbation by noise under the flow-matching formulation, and since no clean image target is available, no loss is applied to the image stream for these samples.

Therefore, for a training sample derived from stage \( i \), the full multimodal input sequence is:
\[
[\mathcal{C}_{i};\; \mathbf{P}_{i-1}^{\text{clean}};\; \mathbf{P}_{i}^{(t_{\text{pose}})};\; \mathbf{I}^{(t_{\text{img}})}],
\]
where \( \mathcal{C}_{i} \) is the concatenated prompt consisting of the first \( i \) per-person descriptions, \( \mathbf{P}_{i-1} \) is the clean structural pose context, \( \mathbf{P}_{i}^{(t_{\text{pose}})} \) is the noised pose input, and \( \mathbf{I}^{(t_{\text{img}})} \) is either the noised image (if \( i = N \)) or pure noise (if \( i < N \)).

Letting \( u_{\text{pose}} = x^{\text{pose}}_0 - P_{i} \) and \( u_{\text{img}} = x^{\text{img}}_0 - I \), we denote the predicted velocities as:
\[
\hat{u}_{\text{pose}} = v_{\theta}^{\text{pose}}(P_{i}^{(t_p)}, t_p), 
\qquad
\hat{u}_{\text{img}} = v_{\theta}^{\text{img}}(I^{(t_i)}, t_i).
\]
The per-stage loss is given by:
\[
\mathcal{L}_i =
\lambda_{\text{pose}}
\big\|
  \hat{u}_{\text{pose}} - u_{\text{pose}}
\big\|_2^2
+
\mathbf{1}_{i=n}\,\lambda_{\text{img}}
\big\|
  \hat{u}_{\text{img}} - u_{\text{img}}
\big\|_2^2.
\]
Each training sample corresponds to a single \revadded{iterative} stage from an \( n \)-person scene, and stages are treated independently and randomly shuffled during training.
Pose supervision is applied at every \revadded{iterative} stage, while image supervision is applied only at the final stage.

\smallskip \noindent \textbf{Inference Configuration.}
At inference time, generation proceeds in an \revadded{iterative} loop over the ordered person descriptions.
At each step \( i \), the previously generated pose latent \( P_{i-1} \) is reused as structural context and concatenated directly into the multimodal token sequence for predicting the next-stage pose latent \( P_{i} \) and image latent \( I_i \).
We use 50 denoising steps and set the guidance scale to 4.0 across all experiments.

\smallskip \noindent \textbf{Run Time.}
Generating one image step (i.e., synthesizing a new person conditioned on the current scene state) takes approximately 50 seconds on a single NVIDIA A100 GPU with 80GB memory. Since our generator proceeds \revadded{iteratively} over all \( n \) people in the scene, the total generation time scales linearly with \( n \). For example, synthesizing a 3-person scene takes approximately 150 seconds. All timings include both pose and image prediction with 50 denoising steps.

\subsection{\revadded{Iterative Generation: Design Details}}
\label{supp:iterative_details}

\smallskip \noindent \textbf{Pose-only state propagation.}
\revadded{A natural alternative to our design would be to recursively pass forward the generated RGB image and edit it at every step. We intentionally avoid this: reusing RGB tends to accumulate texture mistakes, local edit seams, and other pixel-level artifacts over multiple rounds. Passing forward only the pose state preserves the structural scaffold needed for composition while keeping the recurrent state clean and lightweight. Our comparisons to FLUX Kontext and FLUX Fill directly instantiate this editing-based alternative; their lower quality scores (Table~\ref{tab:quality_preservation}) and visible artifact accumulation (Figure~\ref{fig:artifact_accumulation}) confirm that recursive RGB editing degrades both alignment and perceptual quality.}

\smallskip \noindent \textbf{Occlusion and bounding-box overlap.}
\revadded{Occlusion is handled implicitly by the training distribution: target pose renderings come from real scenes, so partially visible people naturally appear with missing or truncated limbs when appropriate, and the model learns to produce such partial poses without explicit occlusion rules. For overlapping layout boxes, later bounding-box assignments overwrite earlier $\tau$ indices in the shared visual grid. This simple strategy works well in typical cases but severe overlap remains a limitation of the current formulation.}

\smallskip \noindent \textbf{Visual consistency across stages.}
\revadded{Visual consistency across iterative stages is not enforced by a dedicated identity-preservation module; when it appears, it mainly arises from shared noise initialization and similar pose conditioning. Because the image and text streams of the pretrained backbone remain frozen, the model preserves the backbone's broader visual prior throughout the iterative process.}

\section{Experimentation Details}
\label{supp_experiments}
This section provides full experimental details, including benchmark construction, baseline implementations, and metric definitions.

\subsection{\evaldataset{} Benchmark Construction}
\label{supp:benchmark}
\evaldataset{} is built from the test split of the Waldo and Wenda dataset~\cite{alper2023learning}, which contains 272 image–caption pairs depicting diverse human–human interaction scenes.
Each original annotation in the Waldo and Wenda dataset provides an original image along with an interaction label that summarizes the main interaction depicted in the scene.
We use these image–label pairs as semantic anchors for prompt construction for our benchmark.

For each sample, we query a multimodal LLM (GPT-5.2) with the reference image and its corresponding interaction label.
We design a system instruction prompt to guide the model in producing a set of three text descriptions that reflect the same scene from different levels of detail.

During prompt generation, the model is instructed to focus on the main interacting participants and any directly relevant surrounding individuals, while avoiding unnecessary descriptions of background figures.
It is also encouraged to incorporate the original interaction label where appropriate, ensuring consistency with the source annotation.

GPT-5.2 produces the following three tiers of prompts for each sample:

\begin{itemize}
    \item \textbf{Tier~A} is a short, interaction-focused description that captures the core activity and the participating people in compact form.
    \item \textbf{Tier~B} adds moderate detail, such as basic spatial layout, individual roles, or actions, while remaining concise.
    \item \textbf{Tier~C} is a long, image-specific description that details all visually relevant participants, their appearances, actions, spatial relations, and nearby context. The goal is for a model faithfully following this prompt to generate a scene resembling the original image.
\end{itemize}

All three tiers are generated jointly in a single LLM call to maintain semantic consistency across tiers and alignment with the original interaction label.
These tiered prompts are used as evaluation inputs for semantic alignment (Tiers~A/B/C) and diversity (Tiers~A/B).

\subsection{Baseline Models}
\label{supp:baselines}

\subsubsection{Text-to-Image Models}
For all text-to-image (T2I) baselines, we input the original benchmark prompts (Tier~A/B/C) to these models to generate images.
For each prompt, we generate 5 different images using different random seeds and evaluate all results for semantic alignment and diversity under the same protocol used for our model.

\paragraph{\textbf{FLUX~[dev]}.}
We use the official FLUX.1-dev implementation released by Black Forest Labs~\cite{flux2024}, following their default inference setup provided in the public repository.
This model is a rectified-flow latent diffusion model based on a pure DiT backbone, with separate modality-specific input streams for text and image tokens.

\paragraph{\textbf{Stable Diffusion 3.5~Large (SD3.5-Large)}}
We use the \path{stabilityai/stable-diffusion-3.5-large} configuration available on HuggingFace.
This model also follows a transformer-based rectified-flow architecture and supports high-resolution synthesis.

\paragraph{\textbf{Stable Diffusion XL (SDXL)}}
We use the \path{stabilityai/stable-diffusion-xl-base-1.0} pipeline available on HuggingFace.
This model adopts a UNet-based diffusion backbone and represents the prior generation paradigm before transformer-based SD3.5.

\paragraph{\textbf{FLUX~[dev] + Scaled Group Inference (SGI)}}
We further include \emph{FLUX~[dev] + SGI}~\cite{parmar2025scaling} as a strong diversity-enhancing baseline. SGI is an inference-time strategy that selects a subset of diverse, high-quality outputs from a larger candidate pool by solving a quadratic assignment problem. Specifically, it defines a unary quality term (e.g., CLIP similarity with the prompt) and a binary diversity term (e.g., DINO feature distance between image pairs), then optimizes a group selection objective balancing both.
We follow the official implementation and default hyperparameters provided by the authors. As required by SGI, the user must specify the group size $K$ and initial candidate size $M$ for each inference. To match our five-seed evaluation protocol, we set $K=5$ and sample $M=64$ candidates using FLUX~[dev]. SGI then prunes this candidate set progressively using intermediate denoising predictions, achieving efficient selection of a diverse image group with improved perceptual quality.

\subsubsection{Editing and Inpainting Models}
All editing and inpainting baselines are evaluated using the same structured prompts that serve as input to our pose–image generator---that is, the outputs from our LLM-based structured prompt expansion stage. For each sample and prompt tier (A/B/C), we generate 5 images using the 5 prompt variants produced by the LLM, following the same sampling protocol as our model. All image variants use the same width and height specified by the structured prompt expansion stage.

\paragraph{\textbf{FLUX Kontext~[dev]}}
We use the official implementation of FLUX Kontext provided by Black Forest Labs~\cite{labs2025flux1kontextflowmatching}, and follow the default inference configuration described in their public repository. FLUX Kontext is a text-driven local editing model that accepts an input image and a textual instruction describing the desired edit.

To construct multi-person scenes, we adopt an iterative editing strategy. We initialize the process with an image generated by FLUX~[dev], conditioned on the first person’s description. Then, for each remaining person, we apply a local edit with the instruction: \emph{“Add the following person: [caption]”}. The editing process is repeated sequentially until all people in the structured prompt have been added.

\paragraph{\textbf{FLUX Fill~[dev]}}
We use the official implementation of FLUX Fill from Black Forest Labs~\cite{flux2024}, using the default inpainting configuration. FLUX Fill accepts both a textual instruction and a binary spatial mask to perform region-constrained content insertion.

Because FLUX Fill requires the initial frame to contain the first subject at the correct spatial location, we need to provide an image where the first person is already positioned appropriately. To achieve an equivalent setup, we initialize the process using our model’s first-person output, which ensures that the first subject is placed according to the target bounding box. We then reapply this same first-person prompt and bounding box mask to FLUX Fill, allowing it to refine the region before proceeding to add the remaining people one at a time.
Then for each additional person, we construct a binary mask from the person’s bounding box and provide it to FLUX Fill along with the instruction \emph{“Add the following person: [caption]”}. This procedure is repeated for all remaining individuals, enabling a stepwise inpainting trajectory that parallels the person-by-person composition used in our model.

\subsubsection{Layout-Controlled Models}
All layout-controlled baselines are evaluated using the same structured input used by our pose–image generator—that is, the outputs of the LLM-based prompt expansion stage.
Each model receives the complete set of bounding boxes and per-person descriptions and synthesizes the full scene in a single forward pass.
For each prompt tier (A/B/C), we generate 5 images per sample using 5 prompt variants.

We include two recent representative layout-conditioned generation models: \textbf{CreatiLayout}~\cite{zhang2025creatilayout} and \textbf{RealCompo}~\cite{zhang2024realcompo}.
For each, we follow the default inference configuration provided in their official GitHub implementations.
These models take as input the entire scene specification—bounding boxes and associated textual descriptions—and produce the final image in one shot without intermediate composition steps.
As they only support square outputs, we use their default image resolution settings during evaluation.

\subsection{Evaluation Metrics}
\label{supp:metrics}
As detailed in the baseline section, all comparison models produce five output images for each benchmark sample under the evaluation protocol. Our generator also generates five outputs using the five structured prompt variants produced by our LLM-based prompt expansion stage. All evaluation metrics operate on these groups of five outputs: for semantic alignment, we compute the metric scores on all five images and average them to obtain a sample-level alignment score; for diversity, we treat the five outputs as a set and compute pairwise differences within the group to obtain a single diversity value per sample. All reported numbers are averaged over samples within each tier.

\subsubsection{Semantic Alignment Metrics}

\paragraph{\textbf{Question Design.}}
For \evaldataset{}, we use a tier-aligned question design.
Tier~A and Tier~B prompts each yield one verification question of the form ``Does this figure show [caption]?'', while Tier~C prompts, which contain fine-grained multi-person specifications, are decomposed into multiple sub-questions targeting individual actions and inter-person relationships. We generate these sub-questions using a custom instruction prompt to GPT-5.2, designed to extract one interpretable visual fact per described person or interaction.
\revadded{For \evaldatasetsupp{}, we adopt the benchmark's original question--answer protocol: simple prompts use one verification question, while complex prompts use multiple questions covering different scene aspects.}

\paragraph{\textbf{VQA Similarity (VQA Sim)}}
We adopt the official implementation of VQA-Score~\cite{lin2024evaluating} and use GPT-5.2 as the answering model to obtain soft probability scores for yes/no answers.
For each prompt or sub-question, we compute the probability that the VLM answers "Yes" given the generated image and query. 
When a prompt is associated with multiple questions, we average the predicted probabilities to obtain a single score per image.

\paragraph{\textbf{VQA Accuracy (VQA Acc)}}
We use the same VLM and question protocol as VQA Sim but record discrete ``Yes''/``No'' answers.
An image is marked correct only if all associated questions are answered ``Yes,'' yielding a strict compositional correctness measure.

\subsubsection{Diversity Metrics}
\paragraph{\textbf{Preprocessing for Feature-Based Metrics}}
For DINO Diff and LPIPS, all images belonging to the same prompt are first converted to tensors in the $[0,1]$ range and then linearly mapped to $[-1,1]$ to match the input conventions of our preprocessing pipeline. Because image resolutions may vary across generations in one prompt group, we identify the maximum height and width among the five samples and \emph{center-pad} all images to a common size using a constant value of $-1.0$. This ensures that each metric receives spatially aligned tensors. The padded images are finally resized to $256{\times}256$ before feature extraction, following standard practice for both DINO and LPIPS.

\paragraph{\textbf{DINO Feature Diversity (DINO Diff)}}
Following~\cite{parmar2025scaling}, we compute feature-space diversity using the ViT-based \path{facebook/dinov2-base} model. After preprocessing, images are normalized with ImageNet statistics and passed through DINOv2 to obtain patch-token embeddings. We discard the CLS token, L2-normalize the patch features, and compute cosine distances for every image pair by averaging $1{-}\cos(\cdot)$ across all patches. The upper-triangular entries of the resulting $5{\times}5$ distance matrix are summarized to produce the per-prompt DINO Diff score.

\paragraph{\textbf{Perceptual Diversity (LPIPS)}}
We adopt the standard VGG-based LPIPS metric~\cite{zhang2018unreasonable}. After padding and resizing, LPIPS is applied to each pair of images to obtain a perceptual distance. These distances populate a symmetric $5{\times}5$ matrix with zeros on the diagonal, and we report the mean of the upper-triangular entries as the per-prompt LPIPS diversity score.

\paragraph{\textbf{VLM-Based Semantic Diversity (GRADE)}}
Following the procedure of Rassin \emph{et al.}~\cite{rassin2024grade}, GRADE evaluates diversity by deriving semantically meaningful visual aspects for each prompt and measuring the normalized entropy of their realizations across the generated images. 
For every concept, we use an LLM (GPT-4o in the original implementation) to generate a small set of aspect-specific questions together with categorical answer sets; these specifications are cached and reused for all prompts sharing the same concept. 
Each generated image is then passed to a VLM-based VQA model, which selects one categorical answer per question, producing a per-prompt distribution over aspect values. 
GRADE computes normalized Shannon entropy over these distributions to obtain the semantic diversity score.
Because our setting focuses on multi-human interactions rather than object-centric concepts, we adapt GRADE's \emph{question-generation} stage so that the
derived aspects focus on human-centric interaction semantics (e.g., roles,
actions, and relationships between people), rather than primarily on
object-level or attribute-level variations. All other components of GRADE,
including VQA answering, value normalization, entropy computation, and caching,
remain identical to the original methodology.

\section{Additional Experiments}
\label{supp_additional_experiments}

\subsection{Ablation: Prompt Parsing vs.\ Full Model}
\label{supp:ablation_prompt_parsing}
\begin{table*}[t]
\centering
\caption{
\textbf{Prompt Parsing Ablation.}
We evaluate each text-to-image model under two input settings: using the original benchmark prompt (Tier~A/B/C) and using our LLM-based expanded parsed prompts (denoted as ``+ Prompt Parsing''), where the expanded prompt is concatenated into a single text input.
All models are evaluated for alignment against the original prompt using VQA-based metrics, as well as diversity using DINO Diff, LPIPS, and GRADE.
}
\label{tab:abl_structured_prompt}
\small
\renewcommand{\arraystretch}{1.2}
\setlength{\tabcolsep}{4.0pt}
\resizebox{0.8\textwidth}{!}{%
\begin{tabular}{l cc cc cc ccc c}
\toprule
{\normalsize \textbf{Method}} &
\multicolumn{2}{c}{\textbf{DINO Diff} $\uparrow$} &
\multicolumn{2}{c}{\textbf{LPIPS} $\uparrow$} &
\multicolumn{2}{c}{\textbf{GRADE} $\uparrow$} &
\multicolumn{3}{c}{\textbf{VQA Acc} $\uparrow$} &
\textbf{VQA Sim} $\uparrow$ \\
\cmidrule(l{4pt}r{4pt}){2-3}
\cmidrule(l{4pt}r{4pt}){4-5}
\cmidrule(l{4pt}r{4pt}){6-7}
\cmidrule(l{6pt}r{6pt}){8-10}
& \textbf{A} & \textbf{B} &
  \textbf{A} & \textbf{B} &
  \textbf{A} & \textbf{B} &
  \textbf{A} & \textbf{B} & \textbf{C} & \\

\midrule

\multicolumn{11}{l}{{\small \textit{\textbf{Text-to-Image Models}}}} \\[-2pt]
FLUX [dev]                    & 0.52 & 0.50 & 0.60 & 0.59 & 0.31 & 0.25 & 0.71 & 0.56 & 0.39 & 0.83 \\
\quad + Prompt Parsing & 0.60 & 0.57 & 0.61 & 0.59 & 0.55 & 0.54  & 0.74 & 0.65 & 0.35 & 0.85    \\
SD3.5-Large                   & 0.55 & 0.52 & 0.64 & 0.61 & 0.27 & 0.24 & 0.68 & 0.55 & 0.31 & 0.76 \\
\quad + Prompt Parsing & 0.62 & 0.59 & 0.63 & 0.62  & 0.55 & 0.50 & 0.71 & 0.52 & 0.29 & 0.75   \\
SDXL                          & 0.62 & 0.60 & 0.64 & 0.63 & 0.39 & 0.31 & 0.55 & 0.20 & 0.12 & 0.57 \\
\quad + Prompt Parsing & 0.66 & \textbf{0.66} & 0.64 & 0.63 & \textbf{0.65} & \textbf{0.59} & 0.49 & 0.15 & 0.13 & 0.53   \\

\midrule

Ours     & \textbf{0.67} & 0.63 & \textbf{0.67} & \textbf{0.65} & 0.55 & 0.54 &
\textbf{0.84} & \textbf{0.72} & \textbf{0.56} & \textbf{0.87} \\
\bottomrule
\end{tabular}
}%
\end{table*}

A question is whether our observed improvement in semantic alignment arises primarily from the use of more detailed and structured prompts, rather than from the proposed generator.

In the main paper's comparison table (see Table~\ref{tab:main_results}), all models are evaluated based on how well their generated images align with the original benchmark prompts from \evaldataset{} (Tiers~A/B/C; see Section~\ref{sec:exp} of the main paper).
\textit{Text-to-image (T2I) baselines}, including FLUX~[dev], SD3.5-Large, and SDXL, receive these benchmark prompts directly as input to generate images.

In contrast, our method first applies an LLM-based \emph{structured prompt expansion} (see Section~\ref{sec:joint-rep} in the main paper) to convert each benchmark prompt into a more detailed and compositional version, consisting of a global caption $c_g$ and a set of per-person descriptions $\{c_i\}_{i=1}^n$.
This expanded and reformatted prompt is then used as the conditioning input to our \revadded{iterative} dual pose--image generator (see Section~\ref{sec:pose_autogressive_gen} of the main paper).

To better isolate the contribution of our generator, we conduct an ablation in which we reuse the exact same parsed prompts (i.e., the same $(c_g, \{c_i\}_{i=1}^n)$ that is input to our generator), concatenate all parts into a single textual prompt, and feed it directly into each T2I baseline.
We then run the same evaluation procedure, scoring how well the resulting images align with the original benchmark prompt.

Table~\ref{tab:abl_structured_prompt} compares each T2I baseline with and without prompt parsing, alongside our full model. 
We observe that using parsed prompts leads to a slight improvement in VQA Accuracy for Tier~A across most models.
This outcome is expected, as Tier~A prompts are originally very concise, and expanding them with additional details provides clearer instructions to the image generators.
In Tier~B, FLUX~[dev] shows a larger boost in accuracy (from 0.56 to 0.65), suggesting some benefit from prompt expansion on mid-granularity scenes. 
But other models see minimal or no improvement, despite receiving the same expanded prompt content.
Overall, our method still achieves significantly higher alignment across all tiers, highlighting the effectiveness of our \revadded{iterative} dual pose--image generator in faithfully realizing the complex structure encoded in the expanded prompts.

In terms of diversity, we observe consistent gains across all baselines when using structured prompt expansion.
This aligns with expectations, as richer and more specific scene descriptions naturally encourage greater variation in generation.
Nonetheless, our full model still outperforms most T2I baselines on the diversity metrics, indicating that it not only aligns more accurately with prompts but also captures a broader range of plausible multi-human configurations.


\subsection{\revadded{Extended Baselines}}
\revadded{The main paper evaluates two representative layout-controlled models alongside T2I and editing baselines (Table~\ref{tab:main_results}). Here we broaden this analysis to six additional layout- and interaction-conditioned methods with publicly available code (Table~\ref{tab:extended_baselines}).
We report VQA Accuracy---our strictest compositional alignment metric---as the central question is whether additional spatial or interaction conditioning improves interaction grounding.
The evaluated methods include bounding-box-conditioned approaches (BoxDiff~\cite{xie2023boxdiff}, Layout Guidance~\cite{chen2024training}, GrounDiT~\cite{lee2024groundit}, R\&B~\cite{xiao2023r}), a segmentation-guided method (Seg2Any~\cite{li2025seg2any}), and an interaction-conditioned model (InteractDiffusion~\cite{hoe2024interactdiffusion}, for which we used GPT to convert our prompts into the required HOI triples).
All methods receive our parsed structured inputs and remain far below ours, especially on Tier~C, where the strongest extended baseline (Seg2Any, 0.21) still falls well short of our method (0.56).
This confirms that coarse spatial conditioning alone is insufficient for fine-grained human--human interaction generation.
Even InteractDiffusion---the only interaction-conditioned method---scores lowest, illustrating that while bounding-box-level control can suffice for localizing entities in human--object settings, human--human interactions depend on fine-grained body articulation and spatial coordination that such coarse representations cannot capture.
Additional related methods discussed in Section~\ref{sec:human_centric_generation} of the main paper (VerbDiff, Chains, PersonaCraft, DreamRenderer) could not be included due to unavailable checkpoints or incompatible input modalities.} 
\begin{table}[t]
\caption{\revadded{\textbf{Extended baselines on \evaldataset{}.} Beyond the baselines evaluated in the main paper, we survey a broader set of layout- and interaction-conditioned methods. We report VQA Accuracy, our strictest alignment metric, as the central question is whether additional spatial or interaction conditioning improves compositional interaction grounding. All methods receive our parsed structured inputs.}}
\label{tab:extended_baselines}
\revtable{%
\centering
\renewcommand{\arraystretch}{1.15}
\setlength{\tabcolsep}{5pt}
\begin{tabular}{lccc}
\toprule
& \multicolumn{3}{c}{\textbf{VQA Accuracy} $\uparrow$} \\
\cmidrule(lr){2-4}
\textbf{Method} & \textbf{Tier A} & \textbf{Tier B} & \textbf{Tier C} \\
\midrule
Seg2Any & 0.42 & 0.36 & 0.21 \\
Layout Guidance & 0.42 & 0.14 & 0.04 \\
BoxDiff & 0.40 & 0.15 & 0.05 \\
GrounDiT & 0.39 & 0.13 & 0.06 \\
R\&B & 0.29 & 0.07 & 0.03 \\
InteractDiffusion & 0.25 & 0.08 & 0.03 \\
\textbf{Ours} & \textbf{0.84} & \textbf{0.72} & \textbf{0.56} \\
\bottomrule
\end{tabular}
}%
\end{table}

\begin{table}[t]
\caption{\revadded{\textbf{Visual-quality preservation.} Our method stays close to the pretrained FLUX backbone across perceptual-quality and human-preference proxies while remaining clearly above iterative editing baselines on artifact-sensitive metrics such as TOPIQ and CLIP-IQA+.}}
\label{tab:quality_preservation}
\revtable{%
\centering
\small
\renewcommand{\arraystretch}{1.15}
\resizebox{\columnwidth}{!}{%
\begin{tabular}{lcccc}
\toprule
\textbf{Model} & \textbf{TOPIQ} $\uparrow$ & \textbf{CLIP-IQA+} $\uparrow$ & \textbf{Aesthetic} $\uparrow$ & \textbf{HPSv3} $\uparrow$ \\
\midrule
FLUX [dev] & \textbf{0.665} & 0.690 & \textbf{5.78} & \textbf{14.12} \\
Ours & 0.655 & \textbf{0.692} & 5.64 & 13.79 \\
SD3.5-Large & 0.622 & 0.681 & 5.57 & 11.29 \\
FLUX Kontext [dev] & 0.519 & 0.623 & 5.44 & 13.55 \\
FLUX Fill [dev] & 0.499 & 0.612 & 5.46 & 11.27 \\
\bottomrule
\end{tabular}
}%
}%
\end{table}

\subsection{\revadded{Visual Quality Preservation}}

\revadded{Table~\ref{tab:quality_preservation} evaluates whether our model preserves the visual quality of its pretrained FLUX backbone. We report four complementary no-reference metrics, each capturing a different aspect of image quality:
TOPIQ~\cite{chen2024topiq} is a learned quality predictor particularly sensitive to local distortions such as blur, noise, and texture artifacts---the types of degradation most likely to arise from iterative editing;
CLIP-IQA+~\cite{wang2023exploring} leverages CLIP features to assess overall perceptual quality, including sharpness and visual coherence;
LAION Aesthetic~\cite{schuhmann2022laion} is a linear probe trained on large-scale human aesthetic ratings, capturing compositional and color-harmony quality;
and HPSv3~\cite{ma2025hpsv3} is trained on human preference data to predict which images people prefer overall.
Our method achieves the best or second-best score on every metric, remaining close to the pretrained FLUX backbone. The iterative editing baselines (FLUX Kontext, FLUX Fill) show substantially larger drops, especially on TOPIQ and CLIP-IQA+, consistent with the artifact accumulation visible in Figure~\ref{fig:artifact_accumulation}. This is expected from our design: the image pathway of the pretrained backbone remains frozen, so visual quality is preserved while interaction-aware structure is learned through the pose pathway.}

\begin{figure}[t]
\revtable{%
\centering
\includegraphics[width=\columnwidth,trim={0 8pt 0 0},clip]{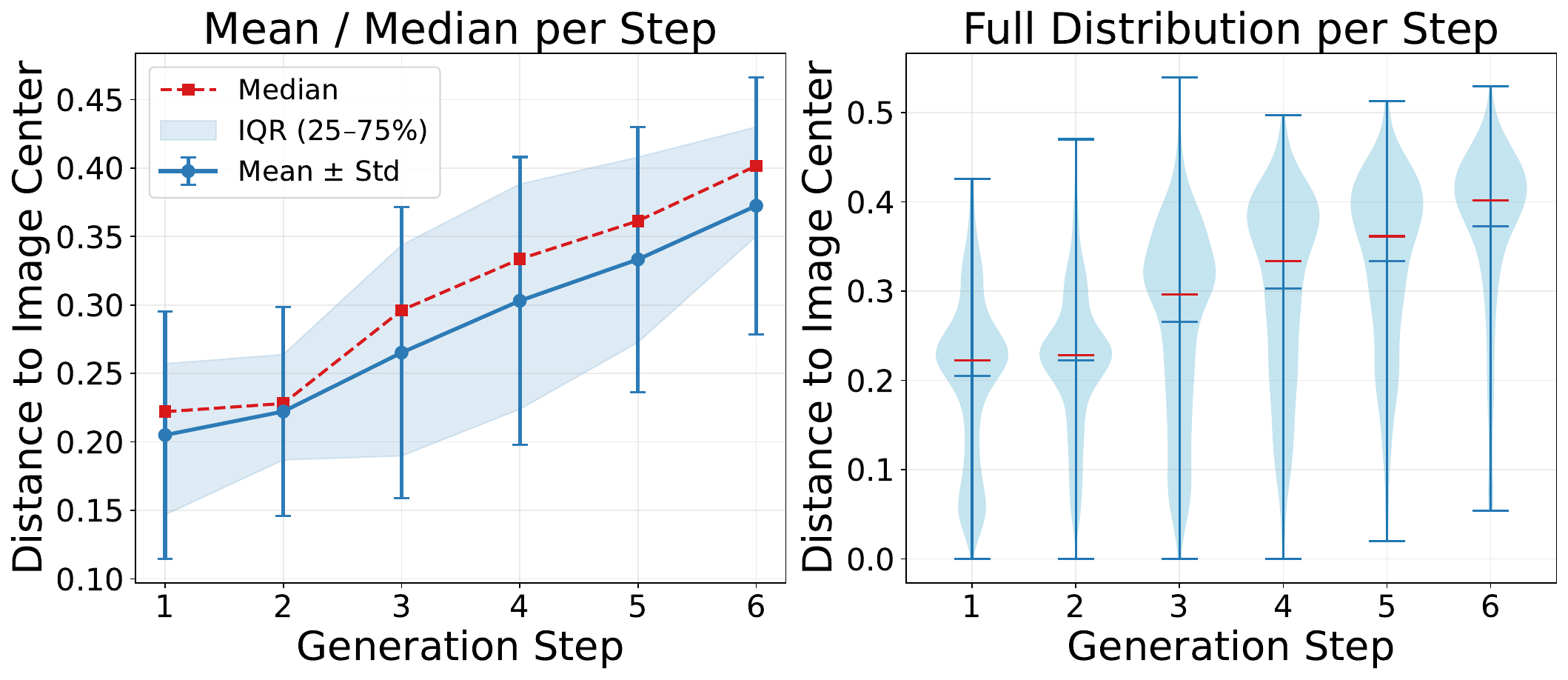}
\vspace{2pt}
\noindent\hdashrule{\columnwidth}{0.4pt}{3pt 2pt}

\vspace{4pt}
\includegraphics[width=\columnwidth,trim={0 2pt 0 0},clip]{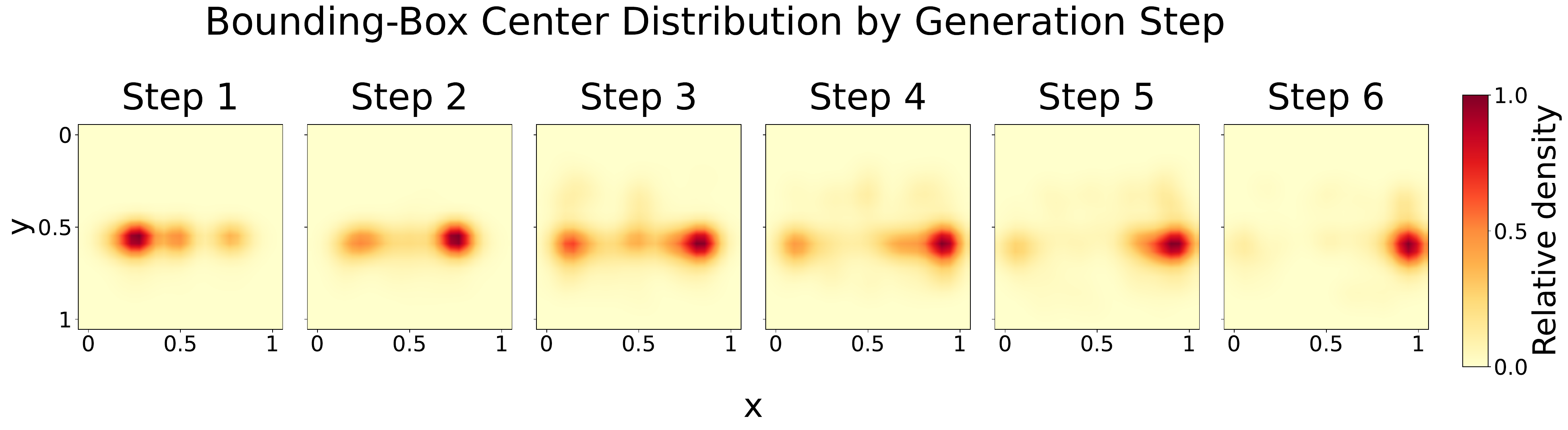}
\caption{\revadded{\textbf{Generation order analysis.}
\emph{Top:} Mean distance from bounding-box center to image center increases with generation step, suggesting a structured core-to-periphery ordering.
The left panel shows mean, median, and IQR; the right panel shows the full distribution via violin plots.
\emph{Bottom:} Spatial heatmaps of bounding-box center positions at each step, showing how later steps extend toward more peripheral regions.
}}
\label{fig:order_analysis}
}%
\end{figure}

\subsection{\revadded{Generation Order Analysis}}
\revadded{As described in Section~\ref{sec:joint-rep} of the main paper, our structured prompt parser instructs the LLM to produce a coherent person ordering: begin from the interaction core, then expand according to spatial proximity, foreground-to-background structure, or left-to-right arrangement.
To verify that this intended pattern is realized in practice, we analyze the bounding-box positions assigned by the LLM across all samples in \evaldataset{}, examining how each person's spatial position relates to its generation step.
Figure~\ref{fig:order_analysis} supports this structured ordering pattern.
The top panel shows that the mean distance from each person's bounding-box center to the image center increases with generation step.
The bottom panel visualizes the spatial distribution of bounding-box centers at each step: earlier steps tend to stay nearer the central interaction band, while later steps extend farther toward peripheral regions.}

\begin{figure}[t]
  \centering
  \includegraphics[width=\columnwidth]{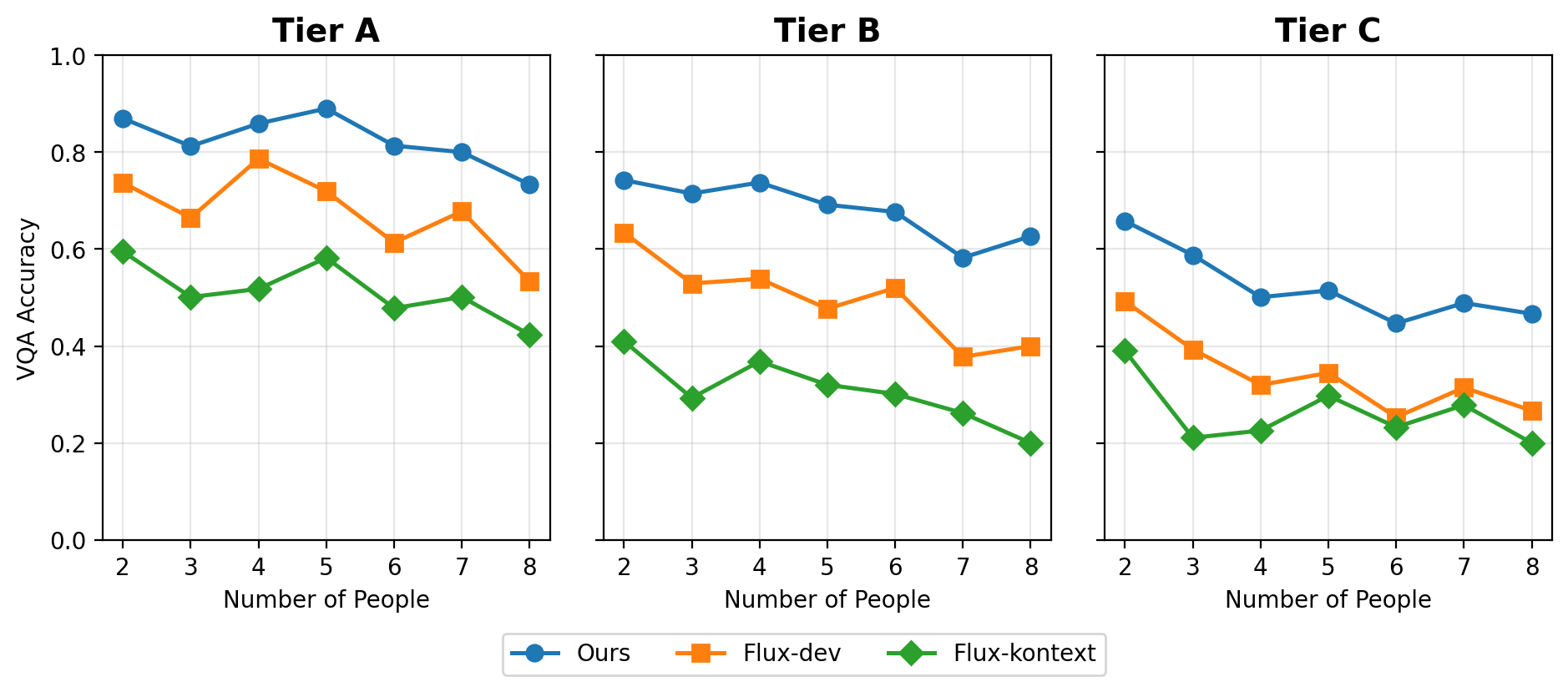}
\caption{
VQA Accuracy breakdown by number of people in the scene, across Tier~A (left), Tier~B (middle), and Tier~C (right) prompts.
As the number of people increases, model performance tends to drop due to increased compositional complexity.
Our model maintains consistently higher alignment across all tiers and shows a more gradual decline, indicating better robustness to multi-person scene complexity.
}
\label{fig:number_of_people}
\end{figure}

\subsection{\revadded{Alignment Breakdown by Scene Complexity}}
\revadded{Figure~\ref{fig:number_of_people} presents a breakdown of alignment performance across scenes with varying numbers of people. Our method consistently maintains higher alignment scores and tends to exhibit a smaller performance drop as scene complexity increases, indicating more stable alignment behavior in multi-person settings.}

\subsection{\revadded{\evaldatasetsupp{} Per-Subset Analysis}}
\revadded{As reported in the main paper, we also evaluate on \evaldatasetsupp{}~\cite{borse2025multihuman}. Here we provide a more detailed per-subset analysis.
For the Single-Person and Multi-Person Simple subsets, alignment is generally high across all models and performance differences are small.
This is largely because these subsets evaluate each prompt with a single verification question that targets coarse scene conditions (e.g., ``at a press conference'') rather than fine-grained person-level constraints such as exact person count or individual actions.
As a result, generations that satisfy the queried scene context can receive positive evaluation even when person-level details are incorrect.
Figure~\ref{fig:supp:case} illustrates this with representative examples from the Simple subset.}

\revadded{The Multi-Person Complex subset is more discriminative, as its prompts use multiple verification questions that jointly test person-level constraints.
Figure~\ref{fig:mmtest_comparison} presents qualitative comparisons on this subset across all baselines with per-sample VQA scores.
Even on this benchmark---which does not specifically target interaction reasoning---our method more reliably satisfies the compositional constraints specified by the prompts.}

\newcommand{\ArtifactRoot}{supplement/figures/supp_figures/artifact_examples/images}

\bigskip
\noindent
\begingroup
\centering
\setlength{\tabcolsep}{2pt}

\begin{tabular}{@{}cc@{}}
\footnotesize \textbf{Ours} & \footnotesize \textbf{FLUX Kontext} \\[2pt]
\includegraphics[width=0.48\columnwidth]{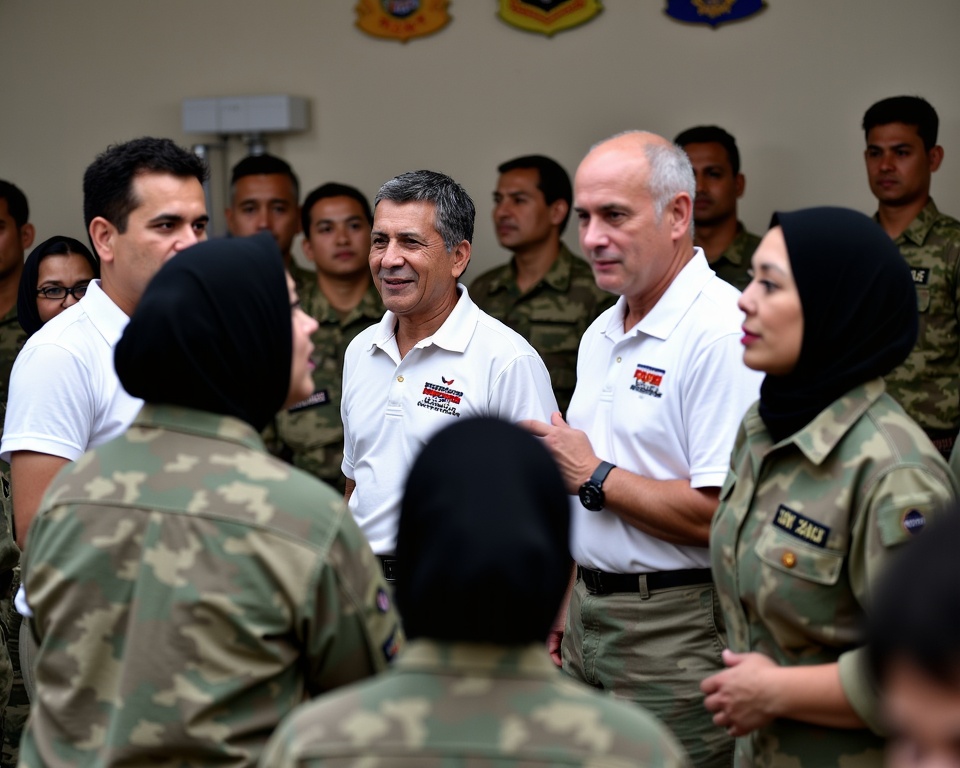} &
\includegraphics[width=0.48\columnwidth]{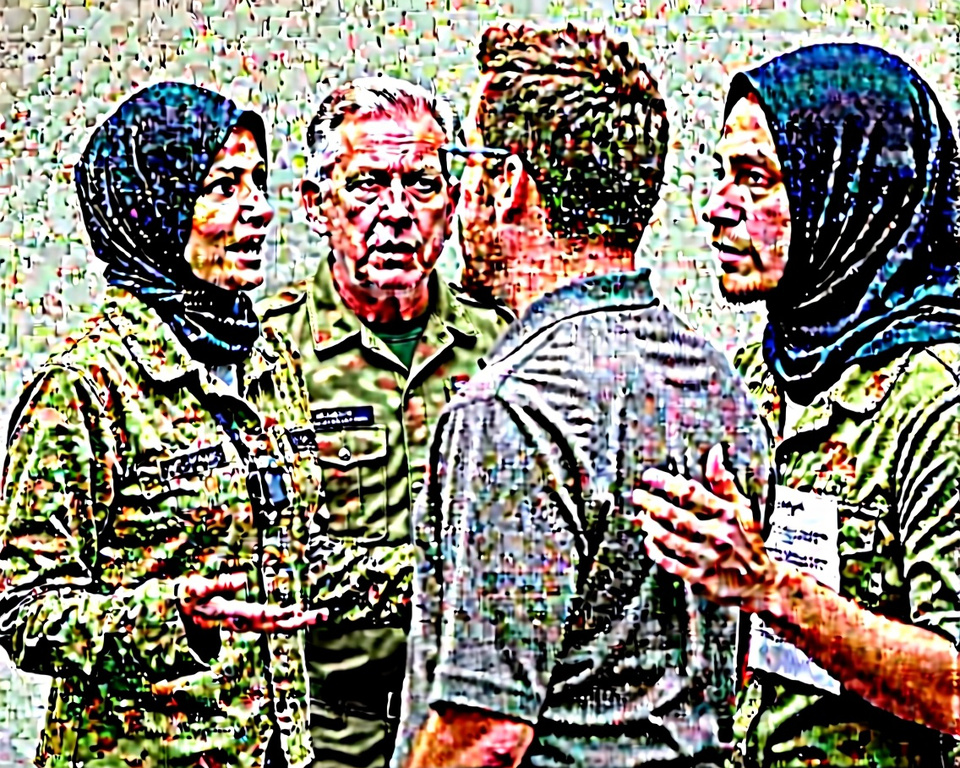} \\[2pt]
\includegraphics[width=0.48\columnwidth]{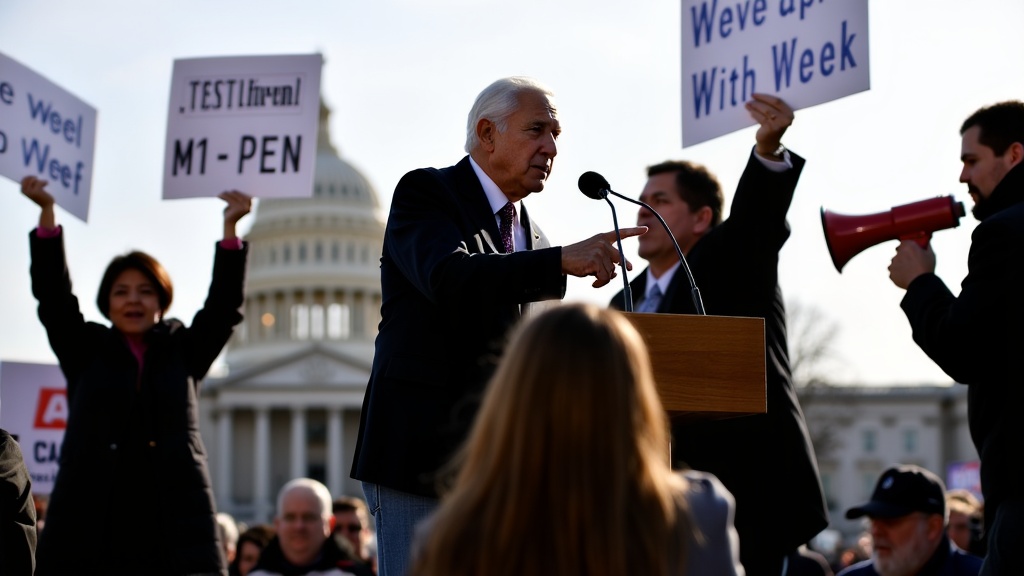} &
\includegraphics[width=0.48\columnwidth]{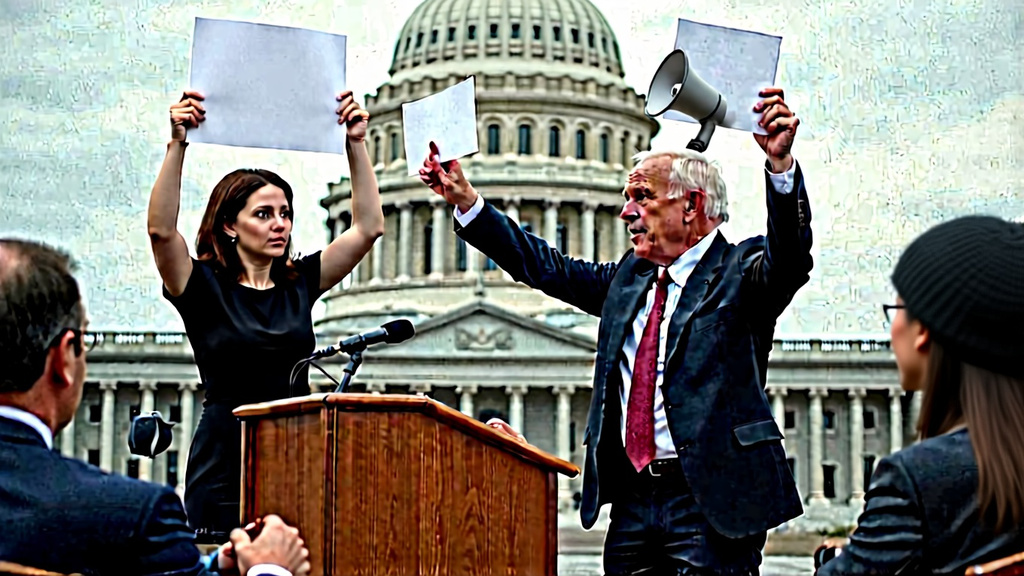} \\[2pt]
\includegraphics[width=0.48\columnwidth]{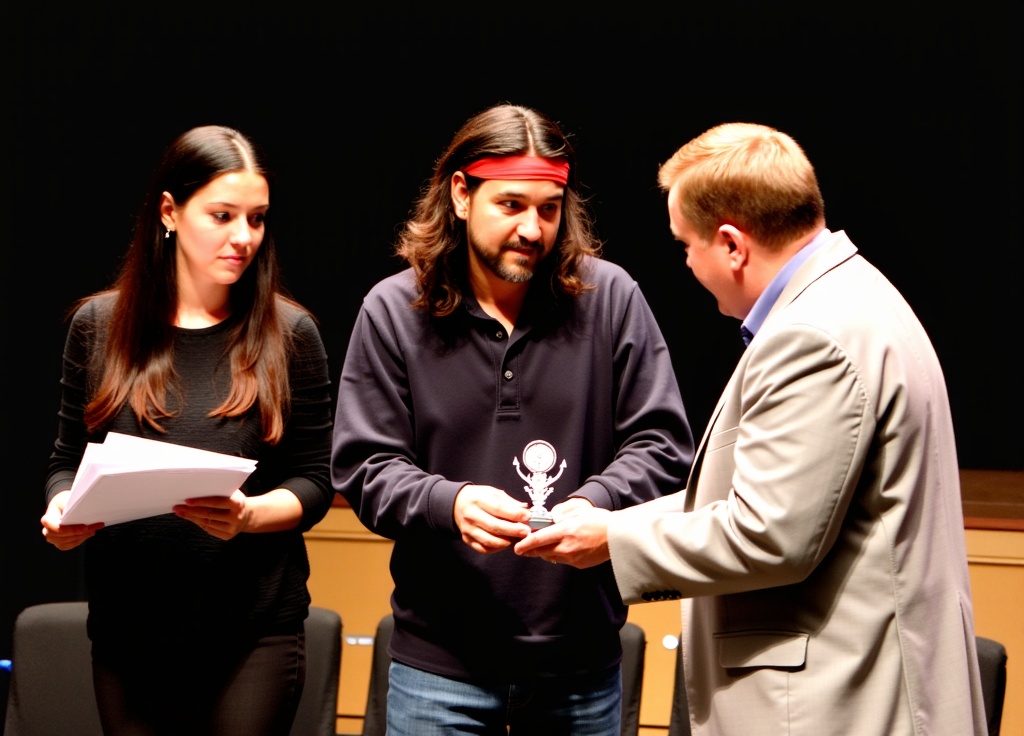} &
\includegraphics[width=0.48\columnwidth]{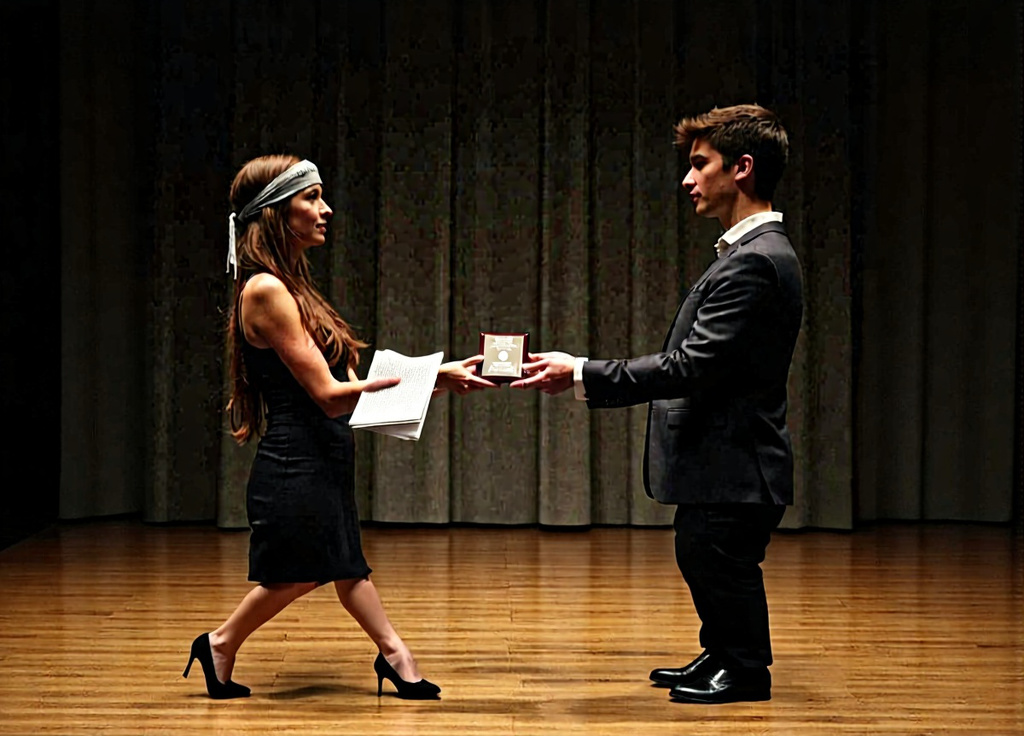} \\[2pt]
\includegraphics[width=0.48\columnwidth]{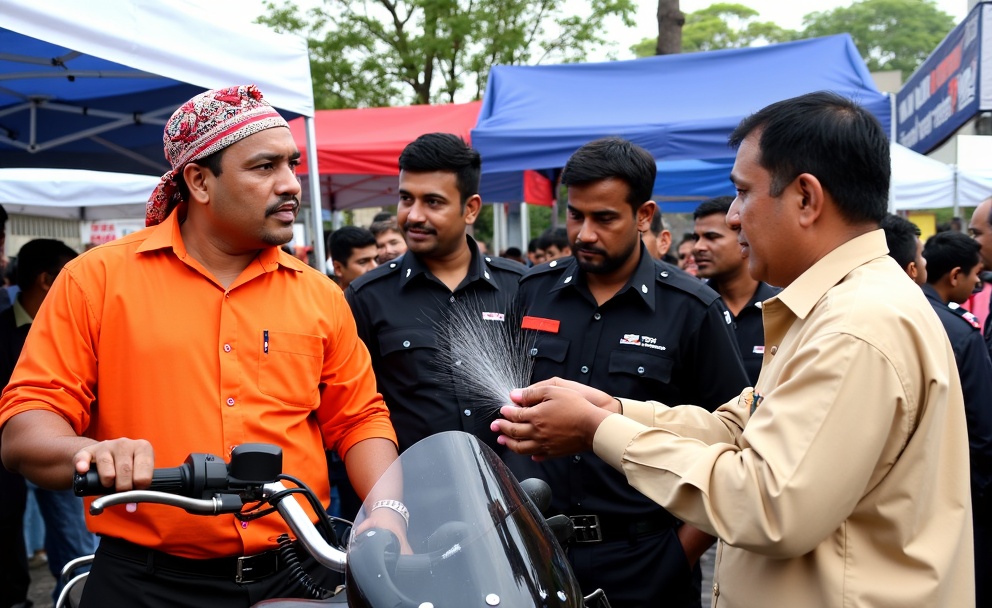} &
\includegraphics[width=0.48\columnwidth]{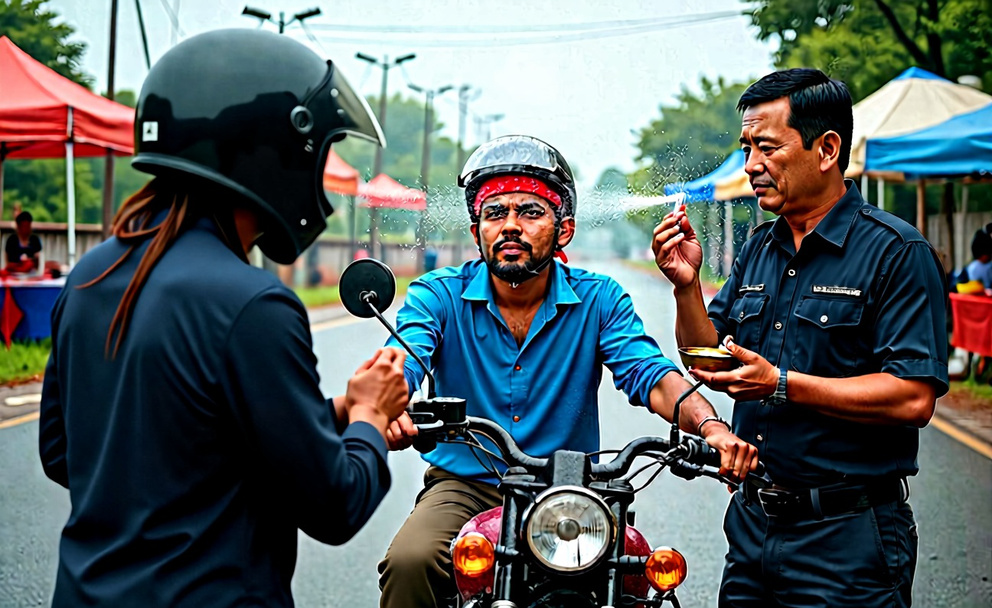} \\
\end{tabular}

\captionof{figure}{\revadded{\textbf{Artifact accumulation in iterative editing.}
The editing-based outputs (right) exhibit noisy textures, oversharpening, and texture distortion from repeated image-space modifications. Our method avoids these artifacts by propagating only the pose state between stages, keeping the RGB generation clean at each step.
Please zoom in to inspect the texture quality differences.}}
\label{fig:artifact_accumulation}
\par
\endgroup
\bigskip

\providecommand{\CaseAlignRowVQA}[4]{%
  {\scriptsize
  VQA Score: #1\\[-0.1em]
  {\color{#3}\fontsize{7}{7.4}\selectfont #4}
  }%
}

\definecolor{ForestGreenDeep}{RGB}{0,80,0}

\begin{figure}[t]
  \centering
  \setlength{\tabcolsep}{3pt}

  \noindent\textbf{Prompt:} \emph{``Portrait picture of \textbf{four} people sitting at a press conference; clear faces visible; microphones, cameras. Ultra-Realistic. 8K.''}\\[-0.1em]
  \noindent\textbf{Question:} \emph{Are the people in this image at a press conference?}

  \vspace{0.5em}

  \begin{tabular}{@{}c@{\hspace{0.8em}}c@{}}
    \begin{minipage}[t]{0.47\columnwidth}
      \centering
      \textbf{Ours}\\[0.25em]
      \includegraphics[width=\linewidth]{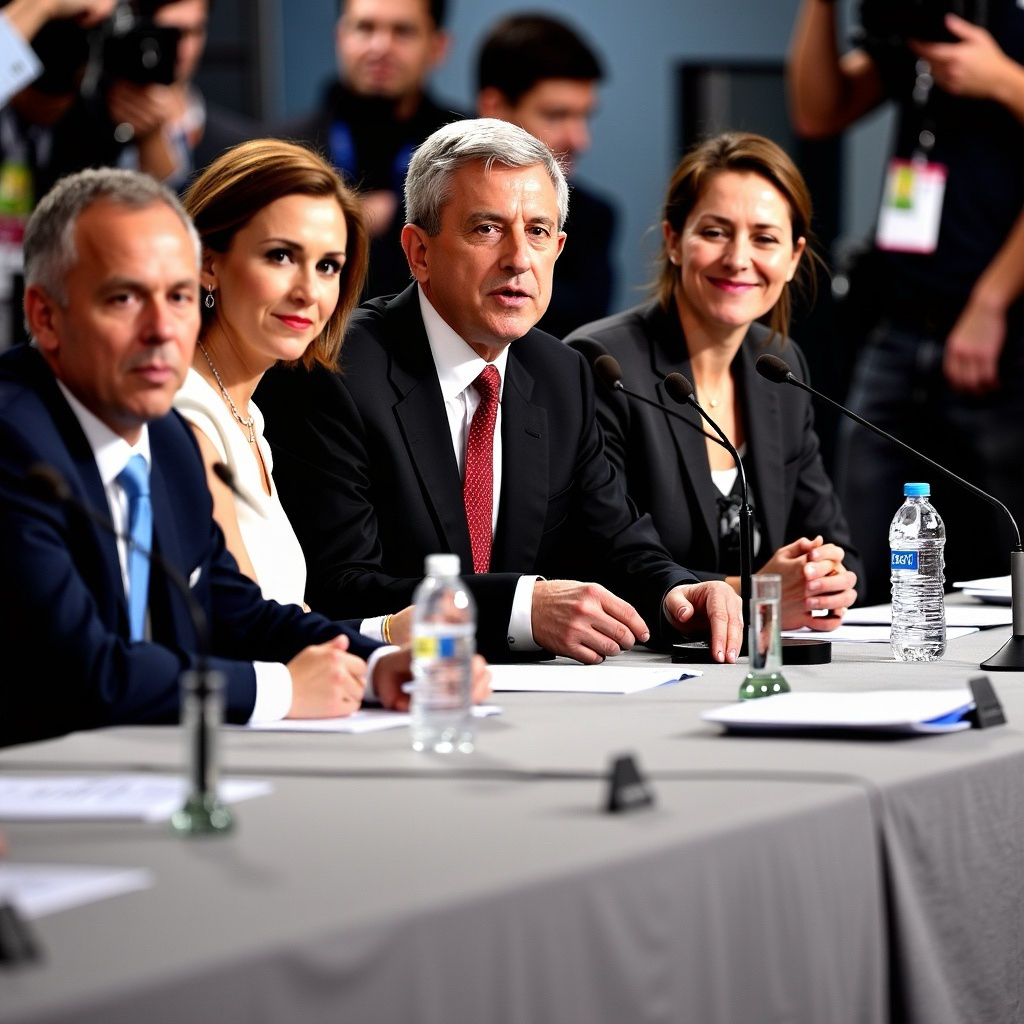}\\[0.15em]
      \CaseAlignRowVQA{\pass}{}{ForestGreenDeep}{Correctly depicts four people at a press conference, fully matching the prompt.}
    \end{minipage}
    &
    \begin{minipage}[t]{0.47\columnwidth}
      \centering
      \textbf{SD3.5-Large}\\[0.25em]
      \includegraphics[width=\linewidth]{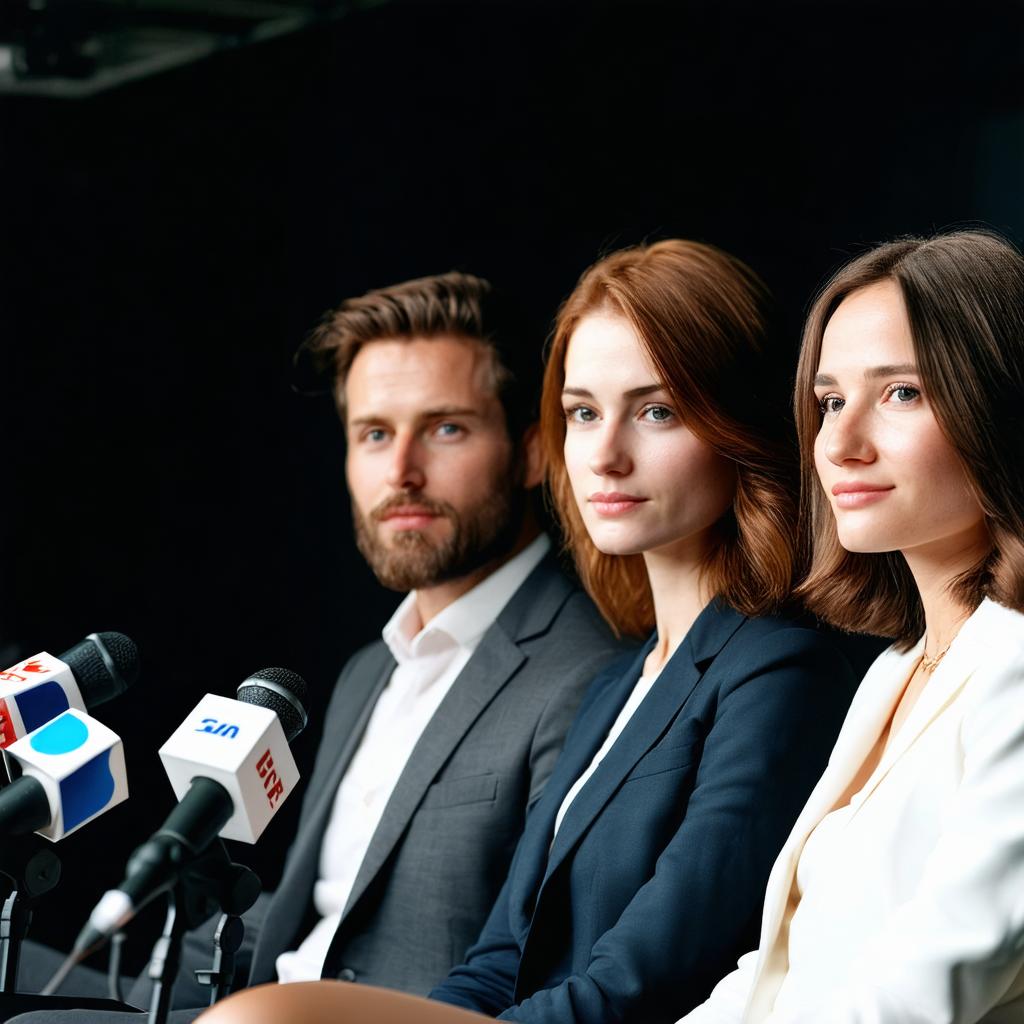}\\[0.15em]
      \CaseAlignRowVQA{\pass}{}{red}{Satisfies the VQA question, but generates three people instead of four.}
    \end{minipage}
  \end{tabular}

  \vspace{1.5em}

  \noindent\textbf{Prompt:} \emph{``\textbf{Five} warriors defending a \textbf{castle}; faces clearly visible. Portrait picture. Ultra-Realistic. 8K.''}\\[-0.1em]
  \noindent\textbf{Question:} \emph{Are the people in this image defending a castle?}

  \vspace{0.5em}

  \begin{tabular}{@{}c@{\hspace{0.8em}}c@{}}
    \begin{minipage}[t]{0.47\columnwidth}
      \centering
      \textbf{Ours}\\[0.25em]
      \includegraphics[width=\linewidth]{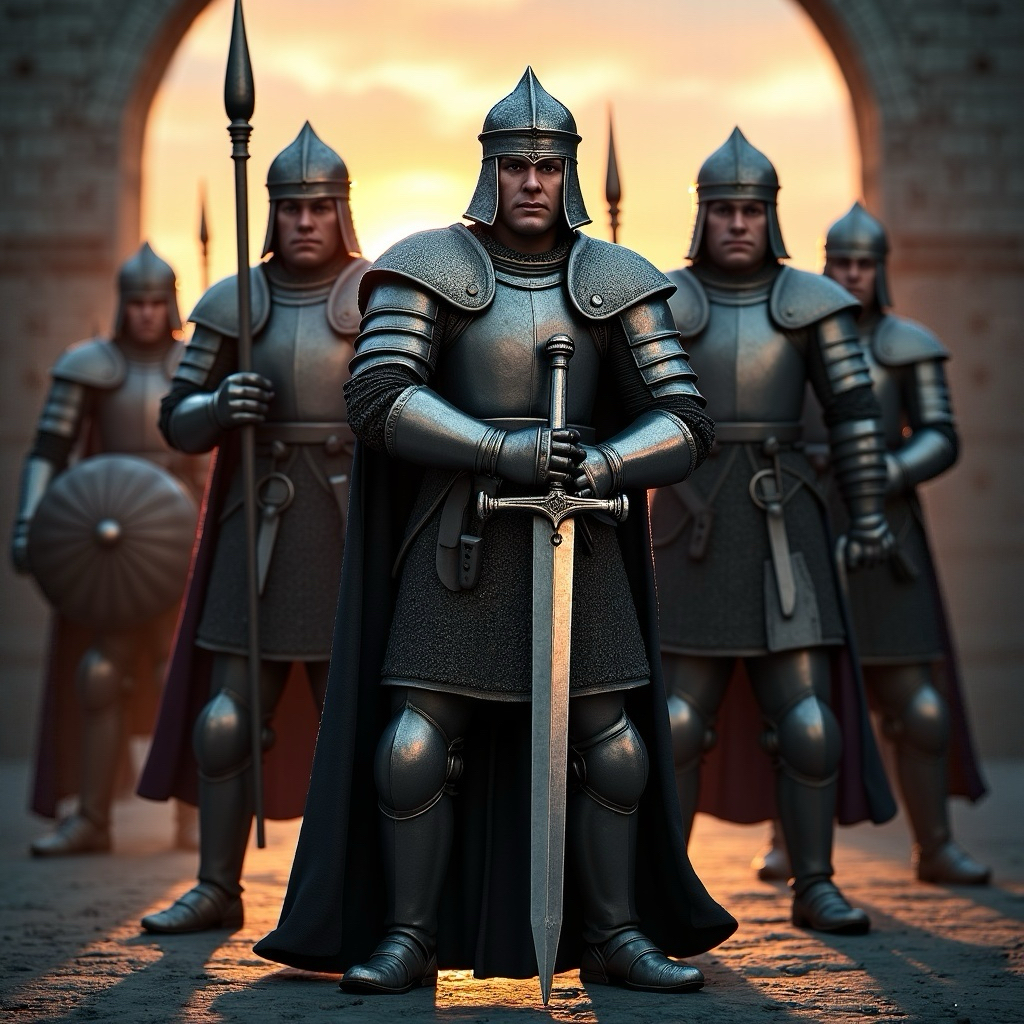}\\[0.15em]
      \CaseAlignRowVQA{\fail}{}{red}{Correct person count, but the castle background is not clearly visible, leading to a negative VQA evaluation.}
    \end{minipage}
    &
    \begin{minipage}[t]{0.47\columnwidth}
      \centering
      \textbf{SD3.5-Large}\\[0.25em]
      \includegraphics[width=\linewidth]{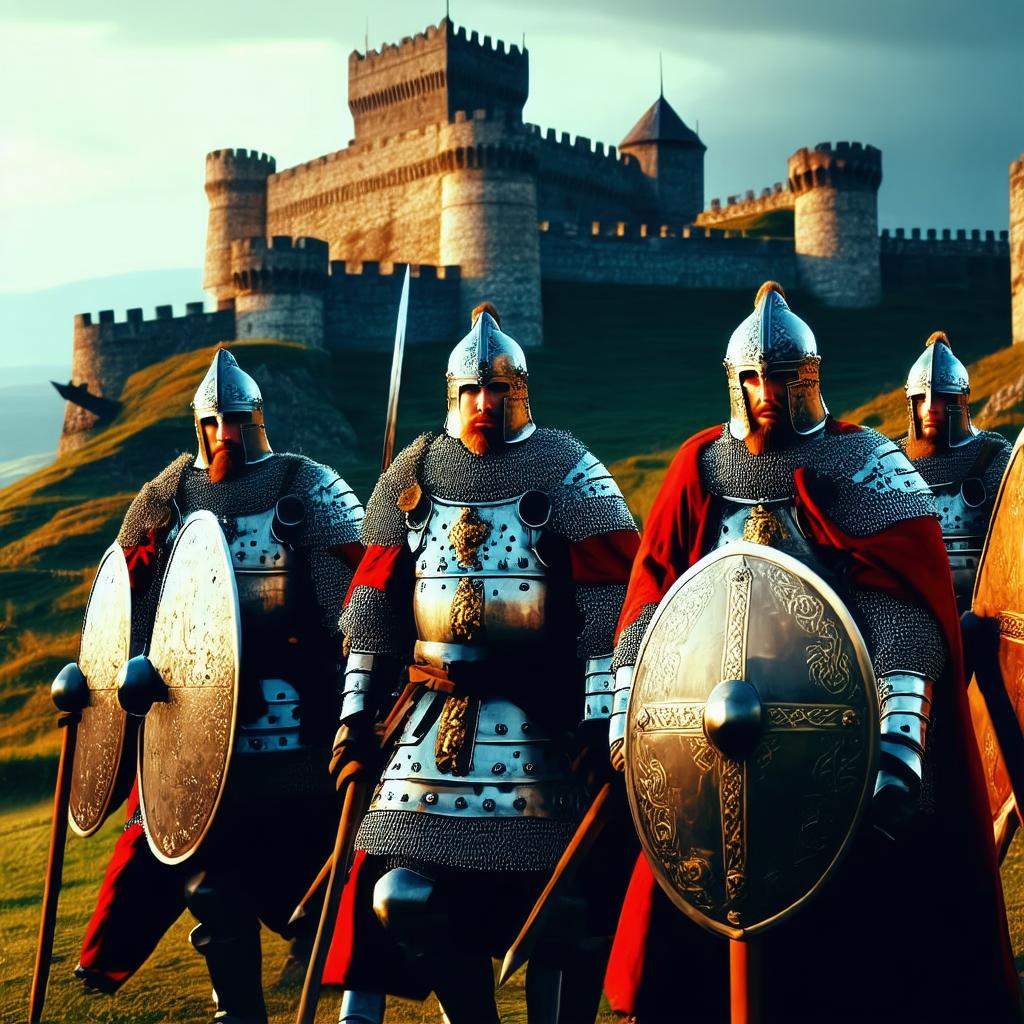}\\[0.15em]
      \CaseAlignRowVQA{\pass}{}{red}{Castle is clearly depicted, but only four people are generated instead of five.}
    \end{minipage}
  \end{tabular}

  \vspace{0.4em}
  \caption{\revadded{Illustrative examples from the \evaldatasetsupp{} \textbf{Simple} subset. The single verification question checks coarse scene conditions (e.g., press conference or castle setting) but does not explicitly enforce person count, so generations with incorrect numbers of people can still receive positive VQA evaluation.}}
  \label{fig:supp:case}
\end{figure}

\clearpage
\subsection{\revadded{Failure Cases and Limitations}}

\revadded{While our method substantially improves interaction grounding, it is not without limitations. We identify several main categories of failure modes:}

\revadded{\noindent \textbf{Missing held objects.}
Our pose-based structural prior captures body configuration and inter-person spatial relationships but does not explicitly represent hand-held objects, which lie outside the scope of the pose modality. As a result, items such as a folder or a pair of glasses may be absent from the generated image even when the overall interaction and scene layout are correctly realized (Fig.~\ref{fig:failure_cases}(a)).}

\revadded{\noindent \textbf{Incorrect fine-grained attributes.}
Subtle person-level attributes such as gaze direction can also deviate from the prompt. In particular, prompts that require multiple bystanders to direct their attention toward a shared focal point are challenging, as the pose representation does not encode gaze explicitly. Fig.~\ref{fig:failure_cases}(b) shows a case where the primary interaction is correctly grounded but bystander gaze directions do not match the prompt.}

\revadded{\noindent \textbf{Severe bounding-box overlap.}
When multiple people occupy highly overlapping spatial regions, our $\tau$-index assignment rule (which overwrites earlier indices with later ones in shared regions) can break down, leading to incorrect person-level role binding. This is a limitation of the current layout representation rather than the generation model itself.}


\definecolor{FailRed}{RGB}{200,30,30}
\newcommand{\failhl}[1]{{\color{FailRed}\textbf{#1}}}

\noindent
\begingroup
\centering

\smallskip
\begin{minipage}{\columnwidth}
\noindent\textbf{(a) Missing held objects.}\\[0.2em]
\noindent{\footnotesize\emph{``A gray-haired man in a dark suit walks arm in arm with an older woman in a white suit. She hooks her hand through his arm as they step forward side by side. \failhl{He carries a folder in his free hand, while she holds a pair of glasses in her other hand.} They pass in front of a large American flag with several onlookers, some in uniform, gathered around them.''}}

\vspace{0.4em}
\noindent
\makebox[\linewidth][c]{\includegraphics[width=0.85\columnwidth]{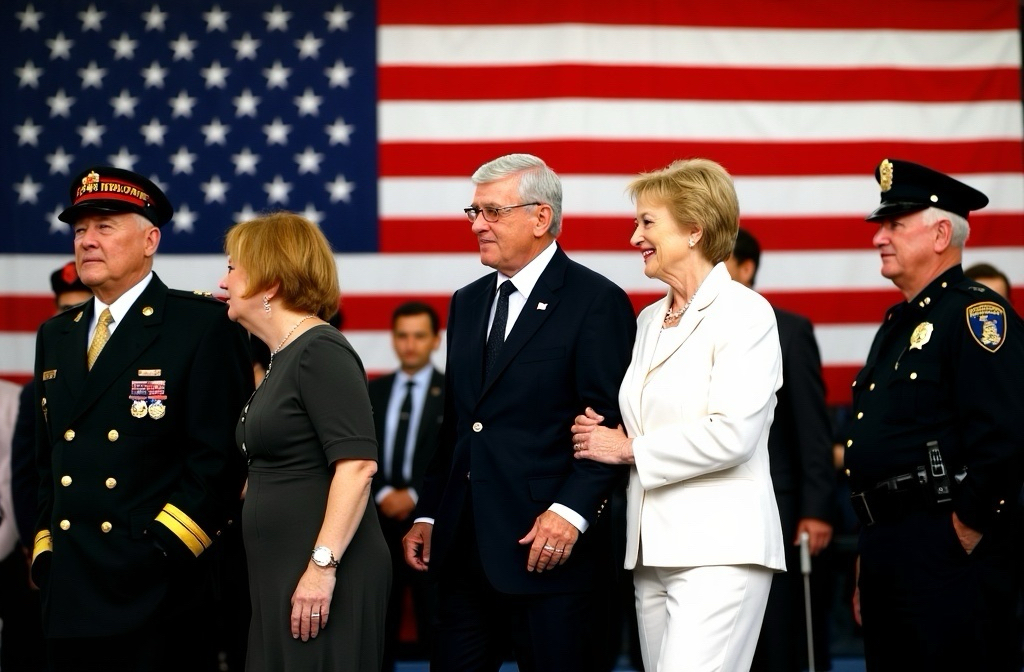}}

\vspace{0.3em}
\noindent{\footnotesize The overall scene composition, interaction (arm in arm), and context (American flag, uniformed onlookers) are correctly realized. However, the specified held objects---a folder and a pair of glasses---are missing. Small handheld items are among the most common failure modes, as they occupy very few pixels and are easily overlooked by the model when the broader scene structure is already established.}

\vspace{1.2em}
\noindent\textbf{(b) Incorrect gaze direction.}\\[0.2em]
\noindent{\footnotesize\emph{``A man in a black suit bends forward to the boy's level and greets him with a handshake as the young boy steps up to meet him. They clasp hands firmly, the suited man smiling warmly while the boy looks up with confidence. Behind the boy, a man in a red shirt smiles proudly, watching the exchange. To the left, a woman holds a phone and observes with a smile, while on the right, a man cradles a small child in his arms, looking on. \failhl{The moment radiates warmth and sincerity as everyone's attention centers on the friendly handshake between the man and the boy.}''}}

\vspace{0.4em}
\noindent
\makebox[\linewidth][c]{\includegraphics[width=0.85\columnwidth]{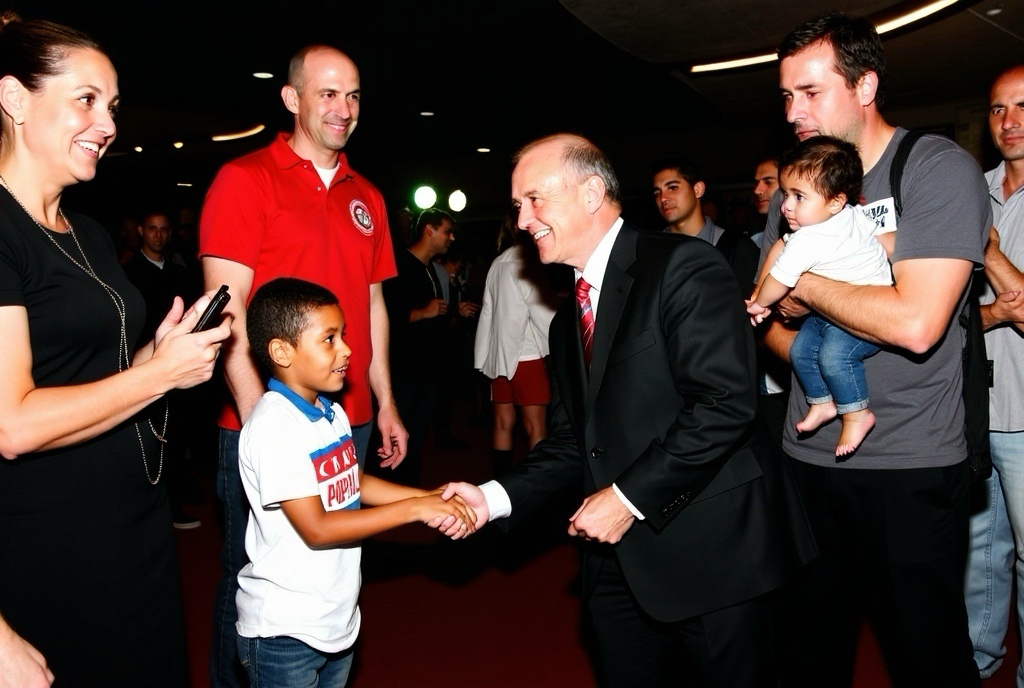}}

\vspace{0.3em}
\noindent{\footnotesize The primary interaction---the handshake between the suited man and the boy---is correctly grounded, and all specified people are present with the correct roles (red-shirted man behind, woman with phone on the left, man with child on the right). However, not all bystanders direct their gaze toward the handshake as described. Gaze direction is a subtle, globally coordinated attribute that the model does not explicitly control at the person level, making it a recurring source of error in scenes where multiple people are expected to attend to a shared focal point.}

\captionof{figure}{\revadded{\textbf{Representative failure cases.} (a)~Missing fine-grained held objects despite correct overall scene structure. (b)~Incorrect bystander gaze direction despite correct interaction grounding and role assignment. Text highlighted in \failhl{red} indicates the unsatisfied prompt conditions.}}
\end{minipage}
\label{fig:failure_cases}
\par
\endgroup
\smallskip


\providecommand{\blacksquare}{\mathord{\hbox{\rule{0.9ex}{0.9ex}}}}
\providecommand{\square}{\mathord{\hbox{\fbox{\rule{0pt}{0.9ex}\hspace{0.9ex}}}}}

\newcommand{\mmA}[1]{\counttext{$#1$}}
\newcommand{\mmB}[1]{\attrtext{$#1$}}
\newcommand{\mmC}[1]{\spatialtext{$#1$}}
\newcommand{\mmABC}[3]{\mmA{#1}\,\mmB{#2}\,\mmC{#3}}

\newcommand{\MMRoot}{supplement/figures/comparisons_mmtest/images}

%
\newcommand{\MMincludegraphics}[3][]{%
  \def\MMfname{\MMRoot/#2_#3.jpg}%
  \IfFileExists{\MMfname}{}{%
    \def\MMfname{\MMRoot/#2_#3.jpeg}%
    \IfFileExists{\MMfname}{}{%
      \def\MMfname{\MMRoot/#2_#3.png}%
    }%
  }%
  \IfFileExists{\MMfname}{%
    \includegraphics[#1]{\MMfname}%
  }{%
    \fbox{\scriptsize Missing: \texttt{#2\_#3}}%
  }%
}

\newcommand{\MMcell}[4][0.12]{%
  \begin{minipage}[t]{#1\linewidth}%
    \centering \small #4\\
    \MMincludegraphics[width=\linewidth]{#2}{#3}%
  \end{minipage}%
}

\newcommand{\MMHeader}[2]{%
  \noindent
  \begin{minipage}[t]{0.10\linewidth}\raggedright\textbf{Prompt:}\end{minipage}%
  \begin{minipage}[t]{0.88\linewidth}\footnotesize\emph{``#1''}\end{minipage}\\[0.2em]
  \begin{minipage}[t]{0.10\linewidth}\raggedright\textbf{Question:}\end{minipage}%
  \begin{minipage}[t]{0.88\linewidth}\footnotesize\emph{#2}\end{minipage}%
}

\begin{figure*}[t]
\centering
\captionsetup{font=footnotesize,skip=5pt}
\begin{adjustbox}{max width=\textwidth,max totalheight=0.90\textheight,center}
\begin{minipage}{\textwidth}
\centering

\MMHeader{A colorful full-body portrait of a total number of two people painting a mural on a building: Among them, one person is applying paint, and another person is mixing colors. Ultra-Realistic. 8K.
}%
{\counttext{a.} Is one person in this image applying paint? \\ \attrtext{b.} Is one person in this image mixing colors? \\ \spatialtext{c.} Is there a mural on a building in this image?}

\vspace{1mm}

\MMcell{515}{ours}{\pass{} \quad \mmABC{\blacksquare}{\blacksquare}{\blacksquare}}\hfill%
\MMcell{515}{flux}{\fail{} \quad \mmABC{\blacksquare}{\square}{\blacksquare}}\hfill%
\MMcell{515}{sd35}{\fail{} \quad \mmABC{\blacksquare}{\square}{\blacksquare}}\hfill%
\MMcell{515}{sdxl}{\fail{} \quad \mmABC{\square}{\square}{\blacksquare}}\hfill%
\MMcell{515}{kontext}{\fail{} \quad \mmABC{\blacksquare}{\square}{\blacksquare}}\hfill%
\MMcell{515}{fill}{\fail{} \quad \mmABC{\blacksquare}{\square}{\blacksquare}}\hfill%
\MMcell{515}{creati}{\fail{} \quad \mmABC{\blacksquare}{\square}{\blacksquare}}\hfill%
\MMcell{515}{realcompo}{\fail{} \quad \mmABC{\blacksquare}{\square}{\blacksquare}}%

\vspace{2mm}

\MMHeader{An urban portrait of a total number of two people waiting at a bus stop: among them, one person is looking at their phone, and another person is talking. Focus on the diverse expressions and the urban environment. Ultra-Realistic. 8K.}%
{\counttext{a.} Is one person in this image looking at their phone? \\ \attrtext{b.} Is one person in this image talking to another person? \\ \spatialtext{c.} Is there a bus stop in this image?}

\vspace{1mm}

\MMcell{517}{ours}{\pass{} \quad \mmABC{\blacksquare}{\blacksquare}{\blacksquare}}\hfill%
\MMcell{517}{flux}{\fail{} \quad \mmABC{\blacksquare}{\square}{\blacksquare}}\hfill%
\MMcell{517}{sd35}{\fail{} \quad \mmABC{\blacksquare}{\square}{\blacksquare}}\hfill%
\MMcell{517}{sdxl}{\fail{} \quad \mmABC{\blacksquare}{\square}{\blacksquare}}\hfill%
\MMcell{517}{kontext}{\fail{} \quad \mmABC{\blacksquare}{\square}{\blacksquare}}\hfill%
\MMcell{517}{fill}{\fail{} \quad \mmABC{\blacksquare}{\square}{\blacksquare}}\hfill%
\MMcell{517}{creati}{\fail{} \quad \mmABC{\blacksquare}{\square}{\blacksquare}}\hfill%
\MMcell{517}{realcompo}{\fail{} \quad \mmABC{\blacksquare}{\square}{\blacksquare}}%

\vspace{2mm}

\MMHeader{A sun-drenched full-body portrait of a total number of three people with clearly visible faces working in a garden: among them, one person is planting flowers, and two people are watering plants. Ultra-Realistic. 8K.}%
{\counttext{a.} Is one person in this image planting flowers? \\ \attrtext{b.} Are two people in this image watering plants? \\ \spatialtext{c.} Are the people in a garden?}

\vspace{1mm}

\MMcell{533}{ours}{\pass{} \quad \mmABC{\blacksquare}{\blacksquare}{\blacksquare}}\hfill%
\MMcell{533}{flux}{\fail{} \quad \mmABC{\blacksquare}{\square}{\blacksquare}}\hfill%
\MMcell{533}{sd35}{\fail{} \quad \mmABC{\blacksquare}{\square}{\blacksquare}}\hfill%
\MMcell{533}{sdxl}{\fail{} \quad \mmABC{\blacksquare}{\square}{\blacksquare}}\hfill%
\MMcell{533}{kontext}{\fail{} \quad \mmABC{\square}{\square}{\blacksquare}}\hfill%
\MMcell{533}{fill}{\fail{} \quad \mmABC{\blacksquare}{\square}{\blacksquare}}\hfill%
\MMcell{533}{creati}{\fail{} \quad \mmABC{\square}{\blacksquare}{\blacksquare}}\hfill%
\MMcell{533}{realcompo}{\fail{} \quad \mmABC{\square}{\square}{\blacksquare}}%

\vspace{2mm}

\MMHeader{A DSLR photograph of a total number of five people gathered around a campfire: among them, two people are roasting marshmallows, and three people are watching. Focus on the campfire glow and the expressions of the people. Ultra-Realistic. 8K.}%
{\counttext{a.} Are two people in this image roasting marshmallows? \\ \attrtext{b.} Are three people watching? \\ \spatialtext{c.} Is there a campfire in this image?}

\vspace{1mm}

\MMcell{574}{ours}{\pass{} \quad \mmABC{\blacksquare}{\blacksquare}{\blacksquare}}\hfill%
\MMcell{574}{flux}{\fail{} \quad \mmABC{\square}{\blacksquare}{\blacksquare}}\hfill%
\MMcell{574}{sd35}{\pass{} \quad \mmABC{\blacksquare}{\blacksquare}{\blacksquare}}\hfill%
\MMcell{574}{sdxl}{\fail{} \quad \mmABC{\square}{\blacksquare}{\blacksquare}}\hfill%
\MMcell{574}{kontext}{\fail{} \quad \mmABC{\square}{\blacksquare}{\blacksquare}}\hfill%
\MMcell{574}{fill}{\fail{} \quad \mmABC{\square}{\blacksquare}{\blacksquare}}\hfill%
\MMcell{574}{creati}{\fail{} \quad \mmABC{\square}{\blacksquare}{\blacksquare}}\hfill%
\MMcell{574}{realcompo}{\fail{} \quad \mmABC{\square}{\blacksquare}{\square}}%

\vspace{2mm}

\MMHeader{A full-body portrait of a total number of five co-workers at an office: among them, three people are standing and two people are sitting. Ultra-Realistic. 8K.}%
{\counttext{a.} Are three people in this image standing? \\ \attrtext{b.} Are two people in this image sitting? \\ \spatialtext{c.} Is there an office in this image?}

\vspace{1mm}

\MMcell{586}{ours}{\pass{} \quad \mmABC{\blacksquare}{\blacksquare}{\blacksquare}}\hfill%
\MMcell{586}{flux}{\fail{} \quad \mmABC{\blacksquare}{\square}{\blacksquare}}\hfill%
\MMcell{586}{sd35}{\fail{} \quad \mmABC{\blacksquare}{\square}{\blacksquare}}\hfill%
\MMcell{586}{sdxl}{\fail{} \quad \mmABC{\square}{\blacksquare}{\blacksquare}}\hfill%
\MMcell{586}{kontext}{\pass{} \quad \mmABC{\blacksquare}{\blacksquare}{\blacksquare}}\hfill%
\MMcell{586}{fill}{\fail{} \quad \mmABC{\square}{\blacksquare}{\blacksquare}}\hfill%
\MMcell{586}{creati}{\fail{} \quad \mmABC{\blacksquare}{\square}{\blacksquare}}\hfill%
\MMcell{586}{realcompo}{\fail{} \quad \mmABC{\blacksquare}{\square}{\blacksquare}}%

\vspace{1mm}

\input{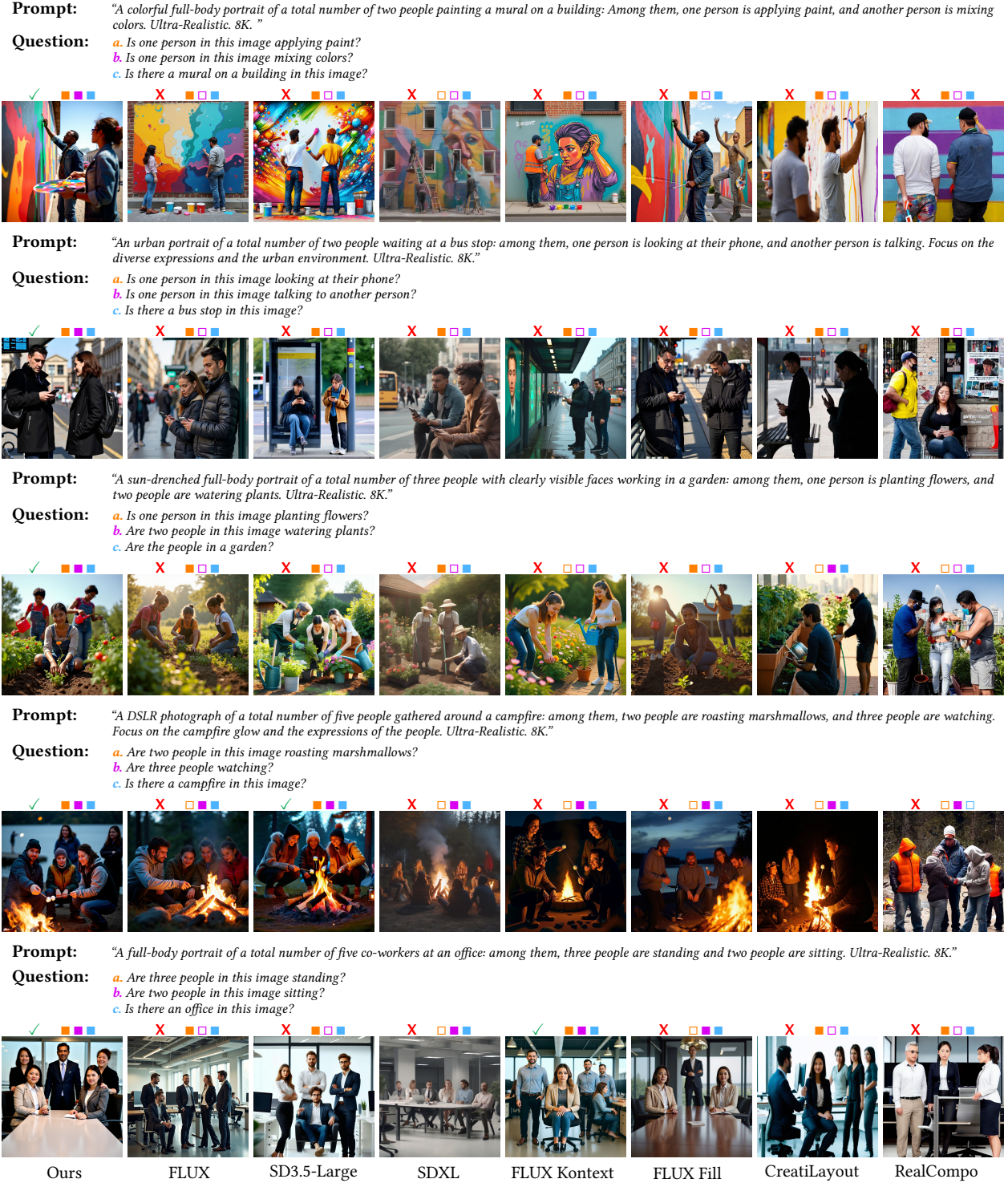}

\end{minipage}
\end{adjustbox}

\caption{\textbf{Qualitative comparison on MultiHuman-Testbench.}
Additional qualitative results on prompts sampled from MultiHuman-Testbench (Multi-Person \emph{complex} set), together with per-sample VQA Accuracy evaluation. For each result, an image is considered \textit{correct} (\pass{}) if a MLLM (GPT-5.2) returns positive answers to \emph{all} evaluation questions associated with that tier. We compare our method against baselines from three model families:
T2I models (FLUX~[dev]~\cite{flux2024}, SD3.5-Large~\cite{esser2024scaling}, SDXL~\cite{podell2023sdxl}), editing / inpainting models (FLUX Kontext~[dev]~\cite{labs2025flux1kontextflowmatching}, FLUX Fill~[dev]~\cite{flux2024}), and layout-controlled models (CreatiLayout~\cite{zhang2025creatilayout}, RealCompo~\cite{zhang2024realcompo}).}
\label{fig:mmtest_comparison}
\end{figure*}

\section{Training Data Collection Pipeline}
\label{supp:data}

Training our dual pose--image \revadded{iterative} generator requires
fine-grained supervision that links per-person textual descriptions,
spatial layout, and 2D pose visualizations. This section provides
additional details of the annotation pipeline used to construct these
aligned multimodal annotations.

\subsection{Source images}
We build our training set from the \emph{Who’s Waldo} training set~\cite{cui2021whoswaldo},
a large-scale, person-centric vision--language dataset originally introduced for
human--human interaction reasoning and visual grounding. The dataset contains a
diverse collection of real-world images depicting rich interactions among
multiple people.

We apply a multi-stage automatic filtering pipeline to remove low-quality images
and samples that do not contain meaningful human--human interactions
(e.g., people merely co-present without interaction).

\textbf{(1) Resolution filtering.}
Starting from the full training pool ($\sim$120k images), we first discard all images
whose resolution is lower than $1024{\times}1024$.

\textbf{(2) Visual quality filtering.}
We then apply an automated visual quality screening stage using a MLLM (Gemini 3 Flash) as a verifier to remove images with obvious
artifacts (e.g., blur, heavy noise, low sharpness, or degraded details).
This step focuses purely on image quality, independent of interaction semantics.

\textbf{(3) Interaction meaningfulness filtering.}
Finally, we remove samples that do not depict meaningful human--human interactions.
For each remaining image, we use a MLLM (Gemini 3 Flash) to (i) assign an interaction label,
(ii) produce a short interaction-focused caption, and
(iii) identify whether the scene contains a \emph{meaningful interaction} or is
\emph{non-meaningful} (e.g., multiple people present but no clear interaction).
We keep only the meaningful subset and discard the rest.

After these stages, we obtain a final set of approximately 30k
high-quality interaction images used for training.

\subsection{Pose extraction}
For each image, we apply an off-the-shelf multi-human pose detector to obtain
articulated 2D poses together with their associated bounding boxes. In our
implementation, we use a closed-source pose detection API~\cite{baidu_body_pose_api},
which provides robust multi-person keypoint estimates in diverse real-world
scenes.
The resulting bounding boxes identify candidate individuals in the scene and
are used both for the caption--person assignment step and as the per-person
layout boxes $\{b_i\}$ supplied to the generator during training. We discard
detections with extremely small spatial extent or unreliable keypoints (e.g.,
missing or low-confidence joints). The remaining detections provide stable
spatial references that allow each per-person description to be paired with its
corresponding pose visualization, forming the structured supervision used by
our model.

\subsection{LLM-based structured captioning and person mapping}
Given the original image, a MLLM (GPT-5.2) is first prompted to
describe the scene at two levels: (i) a \emph{global scene caption} that
summarizes the overall activity and interaction, and (ii) a
\emph{person-wise description list} that provides concise posture,
placement, and interaction cues for each visible individual. The model is
encouraged to write the list in a coherent order, typically beginning
with the main actor(s) and proceeding to nearby participants and
supporting figures.

To associate these descriptions with specific individuals, we perform a
second VLM pass using the detected bounding boxes. For every detected
person, we create a highlighted view by overlaying the bounding box on
the original image. The full set of highlighted views is provided to
GPT-5.2 together with the previously generated description list. The model
is asked to determine which descriptions correspond to which individuals,
select the best-matched caption for each detected person, record the
assigned person’s position within the ordered list, and flag detections
that do not correspond to any description (e.g., background figures or
unreliable pose estimates). Each caption and each detected person is used
at most once, and we discard samples where a consistent one-to-one
mapping cannot be established.

This two-step process yields, for every image, a global scene caption and
a set of aligned per-person annotations that link
text, spatial layout, and pose visualization.

\section{User Study Interface and Protocol}
\begin{figure}[H]
  \centering
  \includegraphics[width=0.66\columnwidth]{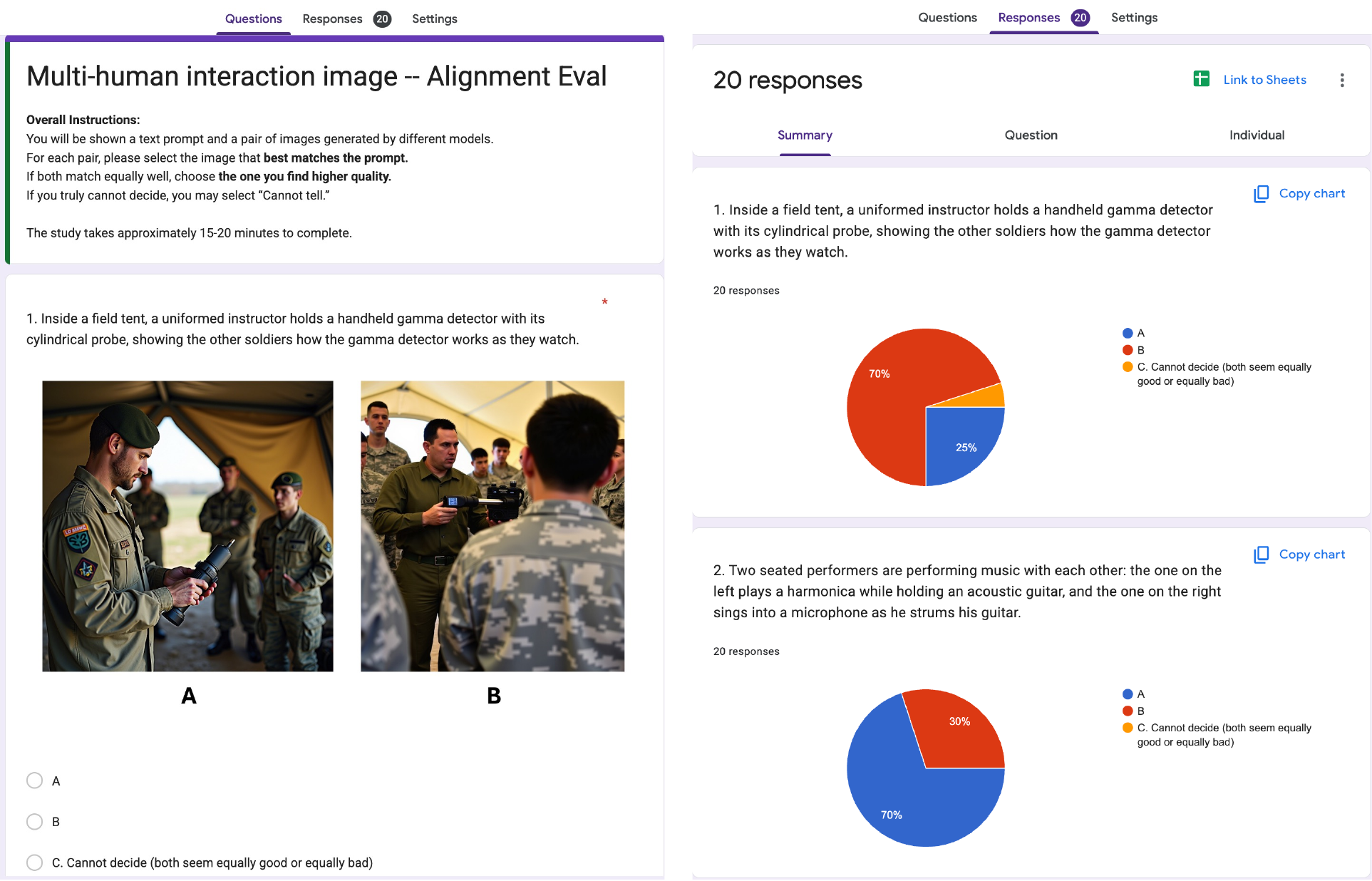}
  \caption{
  Screenshot of our user study interface (left) and collected responses (right). 
  Participants are shown a prompt and a pair of images from different models and asked to select the image that best matches the prompt. 
  If both images are equally good or bad, they may select ``Cannot decide.'' Each comparison is evaluated by 20 participants.
  }
  \label{fig:user_study_ui}
\end{figure}

As described in Section~\ref{sec:user_study} of the main paper, we conduct a user study to evaluate
semantic alignment between generated images and textual prompts.
The quantitative results and analysis are reported in the main paper.

Here, we provide additional supporting material in the form of the user study
interface and instructions, to facilitate transparency and reproducibility. Figure~\ref{fig:user_study_ui} shows a screenshot of the user study interface.
In each trial, participants are presented with a text prompt and two images generated by different models in randomized order, and are asked to select the image that best matches the prompt, or choose ``Cannot decide'' if no clear preference can be made.

\end{document}